\documentclass[a4paper,10pt]{article}

\usepackage{amssymb,amsthm,amsmath,amscd}
\usepackage{epsfig}
\usepackage{subfigure}
\usepackage{algorithmic}
\usepackage{algorithm}

\newtheorem{theorem}{Theorem}[section]
\newtheorem{lemma}[theorem]{Lemma}
\newtheorem{proposition}[theorem]{Proposition}
\newtheorem{corollary}[theorem]{Corollary}

\newtheorem{problem}{Problem}

\theoremstyle{definition}
\newtheorem{definition}{Definition}[section]
\newtheorem{example}[definition]{Example}
\newtheorem{remark}[definition]{Remark}

\newtheorem{assumption}{Assumption}
\DeclareMathOperator{\tv}{\mathrm{TV}(\mathbf{U})}
\DeclareMathOperator{\trace}{trace}
\DeclareMathOperator{\dist}{dist}
\DeclareMathOperator{\supp}{supp}
\DeclareMathOperator{\diam}{diam}
\DeclareMathOperator{\diag}{diag}
\DeclareMathOperator{\ball}{\mathrm{B}}

\newcommand{\reals}{\mathbb R}

\newcommand{\nats}{\mathbb N}
\newcommand{\be}{\begin{equation}}
\newcommand{\ee}{\end{equation}}
\newcommand{\latop}[2]{\substack{{#1}\\{#2}}}
\newcommand{\di}{{\,\mathrm{d}}}

\providecommand{\perm}[2]{\mathrm{P}(#1,#2)}

\providecommand{\abs}[1]{\lvert#1\rvert}
\providecommand{\norm}[1]{\lVert #1\rVert}
\providecommand{\normF}[1]{\left\| #1 \right\|_\mathrm{F}}
\providecommand{\normV}[1]{\left\| #1 \right\|_2}

\newcommand{\qqed}{\nobreak \ifvmode \relax \else
      \ifdim\lastskip<1.5em \hskip-\lastskip
      \hskip1.5em plus0em minus0.5em \fi \nobreak
      \vrule height0.75em width0.5em depth0.25em\fi}

\begin{document}

\title{Foundations of a Multi-way Spectral Clustering Framework
for Hybrid Linear Modeling \thanks{This work was supported by NSF
grant \#0612608 }}
%\footnote{This work was supported by NSF grant \#0612608.}}
%\author{Guangliang Chen \\ Department of Mathematics \\ University of
%Minnesota\\
%        127 Vincent Hall \\ 206 Church St. S.E. \\ Minneapolis, MN 55455
%        USA \\ glchen@math.umn.edu \and Gilad Lerman\thanks{corresponding author: Gilad Lerman, phone: (612)
%        624-5541, fax: (612) 626-2017}\\ Department of Mathematics \\ University of
%Minnesota\\
%        127 Vincent Hall \\ 206 Church St. S.E. \\ Minneapolis, MN 55455
%        USA \\ lerman@umn.edu}
\author{Guangliang Chen, Gilad Lerman \footnote{Corresponding Author: Gilad Lerman,
%Department of Mathematics, University of
%Minnesota, 127 Vincent Hall, 206 Church Street SE, Minneapolis, MN
%55455,
phone: (612) 624-5541,
fax: (612) 626-2017
%, email: lerman@umn.edu
}\\
         {\small Department of Mathematics, University of
Minnesota}\\ {\small 127 Vincent Hall, 206 Church Street SE,
Minneapolis, MN 55455} \\ {\small \{chen0796, lerman\}@umn.edu}}
\date{\today}

%\subjclass[2000]{Primary 52-xx, 60D05, 39Bxx, 46C05; Secondary
%42B99}

\maketitle

\begin{abstract}
The problem of Hybrid Linear Modeling (HLM) is to model and segment
data using a mixture of affine subspaces. Different strategies have
been proposed to solve this problem, however, rigorous analysis
justifying their performance is missing. This paper suggests the
Theoretical Spectral Curvature Clustering (TSCC) algorithm for
solving the HLM problem, and provides careful analysis to justify
it. The TSCC algorithm is practically a combination of Govindu's
multi-way spectral clustering framework (CVPR 2005) and Ng et al.'s
spectral clustering algorithm (NIPS 2001). The main result of this
paper states that if the given data is sampled from a mixture of
distributions concentrated around affine subspaces, then with high sampling
probability the TSCC algorithm segments well the different
underlying clusters. The goodness of clustering depends on the
within-cluster errors, the between-clusters interaction, and
a tuning parameter applied by TSCC.
%flatness of the distributions, their separation,
%and a tuning parameter applied by the algorithm.
The proof also provides new insights for the analysis of Ng et
al.~(NIPS 2001).
%Parallel works show how to incorporate necessary numerical
%techniques so that this framework may perform well on various data
%sets that arise in real world applications.
\end{abstract}

\newline
\newline
 \noindent Keywords: Hybrid linear modeling, clustering
$d$-flats, multi-way clustering, spectral clustering, polar
curvature, perturbation analysis, concentration inequalities
%
%\newline
%\newline
%%
%\noindent Communicated by Albert Cohen

\section{Introduction}
The problem of \textit{Hybrid Linear Modeling} (HLM) is to model
data using a collection of affine subspaces, or equivalently, flats,
and simultaneously segment data into subsets representing those
flats (see also formulations in \cite{Vidal05} and \cite{Ma07}).
%It is formulated by Vidal et al.~\cite{Vidal05} as
%\emph{subspace segmentation}, and also by Ma et al.~\cite{Ma07} as
%\emph{estimation of subspace arrangements}.
This problem has diverse applications in many areas, such as motion
segmentation in computer vision, hybrid linear representation of
images, classification of face images, and temporal segmentation of
video sequences (see \cite{Ma07} and references therein). Also, it
is closely related to sparse representation and manifold
learning~\cite{Szlam08qflats,Mairal08Discriminative}.

Many algorithms and strategies can be applied to this problem. For
example,
RANSAC~\cite{Fischler81RANSAC,Torr98geometricmotion,Yang06Robust},
$K$-Subspaces~\cite{Kambhatla94fastnon-linear,Ho03}/$K$-Planes~\cite{Bradley00kplanes,Tseng00nearest},
Subspace Separation~\cite{Costeira98,Kanatani01,Kanatani02},
Mixtures of Probabilistic PCA~\cite{Tipping99mixtures}, Independent
Component Analysis~\cite{ICA_00}, Tensor Voting~\cite{Medioni00},
Multi-way Clustering~\cite{Agarwal05,Govindu05,Shashua06,Agarwal06},
Generalized Principal Component Analysis~\cite{Vidal05,Ma07},
Manifold Clustering~\cite{Souvenir05}, Local Subspace
Affinity~\cite{Yan06LSA}, Grassmann Clustering~\cite{Gruber06},
Algebraic Multigrid~\cite{Kushnir06multiscale}, Agglomerative Lossy
Compression~\cite{Ma07Compression} and Poisson Mixture
Model~\cite{Haro08TPMM}. However, we are not aware of any
probabilistic analysis of the performance of such algorithms given
data sampled from a corresponding hybrid model (with additive
noise). The goal of this paper is to rigorously justify a particular
solution to the HLM problem.

For simplicity we restrict the discussion to the case where all the
underlying flats have the same dimension $d\geq 0$, although our
theory extends to mixed dimensions by considering only the maximum
dimension. We also assume here that the intrinsic dimension, $d$,
and the number of clusters, $K$, are known, and leave their
estimation to future works.

Our solution to HLM, the Theoretical Spectral Curvature Clustering
(TSCC) algorithm, follows the multi-way spectral clustering
framework of Govindu~\cite{Govindu05}. This framework (when applied
to HLM) starts by computing an affinity measure quantifying
$d$-dimensional flatness for any $d+2$ points of the data. It then
forms pairwise weights by decomposing the corresponding $(d+2)$-way
affinity tensor. At last, it applies spectral clustering
(e.g.,~\cite{Shi00}) with the pairwise weights. However, these steps
are based on heuristic arguments~\cite{Govindu05}, with no formal
justification for them.

The TSCC algorithm combines Govindu's framework~\cite{Govindu05}
with Ng et al.'s spectral clustering algorithm~\cite{Ng02}, while
introducing ``the polar tensor'' (see
Subsection~\ref{subsec:polar_tensor}). We justify the TSCC algorithm
following the strategy of~\cite{Ng02} in two steps. First, we
consider a general affinity tensor instead of the polar tensor, and
control the goodness of clustering of TSCC by the deviation of the
affinity tensor from an ideal tensor
(Section~\ref{sec:perturbation_analysis}). Next, we show that for a
more restricted class of affinity tensors (also including the polar
tensor) and data sampled from a hybrid linear model, the TSCC
algorithm clusters the data well with high sampling probability
(Section~\ref{subsec:prob_analysis_result}). For the polar tensor,
the goodness of clustering can be expressed in terms of the
within-cluster errors (which depend directly on the flatness of the
underlying measures), the between-clusters interaction (which
depends on the separation of the measures), and a tuning parameter
applied by TSCC (Section~\ref{sec:prob_analysis}).

The rest of the paper is organized as follows. In
Section~\ref{sec:theoretical_background} we review some theoretical
background.
%In particular, we formulate more
%precisely the problem of hybrid linear modeling, introduce a class
%of Least Squares Via Integration (LSVI) curvatures as well as
%formulate their relation with least squares approximations, and at
%last review both the spectral clustering algorithm of Ng et
%al.~\cite{Ng02} and Govindu's framework~\cite{Govindu05} of
%multi-way spectral clustering.
In Section~\ref{sec:theoretical_algorithm} we present the TSCC
algorithm as a combination of Govindu's framework~\cite{Govindu05}
and Ng et al.'s algorithm~\cite{Ng02} while using the specific polar
tensor. Both Sections~\ref{sec:perturbation_analysis}
and~\ref{sec:prob_analysis} analyze the performance of the TSCC
algorithm. The former section presents the main technical estimates
for a large class of affinity tensors, while quantifying fundamental
notions, in particular, the goodness of clustering. The latter
section assumes a hybrid linear probabilistic model and the use of the polar tensor, and relates the
estimates of Section~\ref{sec:perturbation_analysis} to the sampling
distribution of the model.
Section~\ref{sec:conclusions} concludes with a brief discussion and
possible avenues for future work. Mathematical proofs are given in
the appendix.

\section{Background}\label{sec:theoretical_background}

\subsection{Notation and Basic Definitions}\label{sec:notation}
Throughout this paper we assume an ambient space ${\reals}^D$ and a
collection of $d$-flats that are embedded in $\reals^D$, where
$0\leq d<D$.

We denote scalars with possibly large values by upper-case plain
letters (e.g., $N,C$), and scalars with relatively small values by
lower-case Greek letters (e.g., $\alpha,\varepsilon$); vectors by
boldface lower-case letters (e.g., $\mathbf{u,v}$); matrices by
boldface upper-case letters (e.g., $\mathbf{A}$); tensors by
calligraphic capital letters (e.g., $\mathcal{A}$); and sets by
upper-case Roman letters (e.g., $\mathrm{X}$).

For any integer $n>0$, we denote the $n$-dimensional vector of ones
by $\mathbf{1}_n$, and the $n\times n$ matrix of ones by
$\mathbf{1}_{n\times n}$. The $n\times n$ identity matrix is written
as $\mathbf{I}_n$.

The $(i,j)$-element of a matrix $\mathbf{A}$ is denoted by $A_{ij}$,
and the $(i_1,\ldots,i_{n})$-element of an $n$-way tensor
$\mathcal{A}$ is denoted by $\mathcal{A}(i_1,\ldots,i_{n})$. We
denote the transpose of a matrix $\mathbf{A}$ by $\mathbf{A}'$ and
that of a vector $\mathbf{v}$ by $\mathbf{v}'$. The Frobenius norm
of a matrix/tensor, denoted by $\normF{\cdot}$, is the $\ell_2$ norm
of the quantity when viewed as a vector.

%The column space of
%a matrix $\mathbf{A}$ is denoted by $\Sp(\mathbf{A})$.
If $k>0$ is an integer and $\mathbf{A}$ is a positive semidefinite
square matrix, we use $E_k(\mathbf{A})$ to denote the subspace
spanned by the top $k$ eigenvectors of $\mathbf{A}$, and
$P^k(\mathbf{A})$ to represent the orthogonal projector onto
$E_k(\mathbf{A})$.

If $\mathbf{x} \in {\reals}^D$ and $F$ is a $d$-flat in
$\mathbb{R}^D$, then we denote the orthogonal distance from
$\mathbf{x}$ to $F$ by $\dist(\mathbf{x},F)$. For any $r>0$, the
ball centered at $\mathbf{x}$ with radius $r$ is written as
$\ball(\mathbf{x},r)$. If $c>0$, then $c\cdot \ball(\mathbf{x},r):=
\ball(\mathbf{x},c \cdot r)$.
%That is,
%$B(\mathbf{x},r) = \{ \mathbf{y} \in {\reals}^D \mid
%\dist(\mathbf{y},\mathbf{x})\leq r\}$, where
%$\dist(\mathbf{y},\mathbf{x})$ represents the Euclidean distance
%between $\mathbf{y}$ and $\mathbf{x}$.
If $\mathrm{S}$ is a subset of $\reals^D$, we denote its diameter by
$\diam(\mathrm{S})$ and its complement by $\mathrm{S}^c$. If
$\mathrm{S}$ is furthermore  discrete, we use $|\mathrm{S}|$ to
denote its number of elements.

%If $B$ is a $d$-dimensional ball in a $d$-flat $L \subset \reals^D$,
%then we denote by $B^\perp$ the ``orthogonal complement'' of $B$ in
%$\mathbf{R}^D$, i.e.,
%%
%%$$B^\perp := \{\mathbf{y} \in
%%\reals^D  : \ \exists \, \mathbf{x} \in B \ \text{s.t.} \ \langle
%%\mathbf{y}-\mathbf{x},\mathbf{z}-\mathbf{x} \rangle = 0 \quad
%%\forall \, \mathbf{z} \in L\}\,.$$
%$$B^\perp := \{\mathbf{y} \in L^c  \colon P_L(\mathbf{y}) \in B\}.$$
%%
%%We define the Tube of distance $r$ around $B$ by the formula
%%$$\rm{T_{ube}}(B,r)=\{y \in B^\perp: \ \dist(y,B) \leq r\}\,.$$

Let $\mu$ be a measure on ${\reals}^D$. We denote the support of
$\mu$ by $\supp(\mu)$, its restriction to a given set $\mathrm{S}$
by $\mu|_\mathrm{S}$, and the product measure of $n$ copies of
$\mu$, where $n \in \nats$, by $\mu^n$. The $d$-dimensional Lebesgue
measure is denoted by ${\cal L}_d$. Also, we use $(\reals^D)^n$ to
denote the Cartesian product of $n$ copies of ${\reals}^D$.
%
%For any $n+1$ distinct points $\mathbf{z}_1,\ldots, \mathbf{z}_{n+1}
%\in \reals^D$, we denote by
%$V_{n}(\mathbf{z}_{1},\ldots,\mathbf{z}_{n+1})$ the $n$-volume of
%the $n$-simplex formed by these points. Then the {\em polar sine} at
%each vertex $\mathbf{z}_i, 1\leq i\leq n+1$, is
%\begin{equation}\label{eq:polar_sine}
%\mathrm{psin}_{\mathbf{z}_i}(\mathbf{z}_{1},\ldots,\mathbf{z}_{n+1})=
%     \frac{n!\cdot V_{n}(\mathbf{z}_{1},\ldots,\mathbf{z}_{n+1})}{\prod_{1\leq j\leq n+1\atop j\ne i}
%    \norm{\mathbf{z}_j-\mathbf{z}_i}}.
%\end{equation}
%In order to study $d$-dimensional flatness, we use the polar sines
%at all vertices of $(d+1)$-simplices, i.e., $n=d+1$.

We use $\perm{n}{r}$ to denote the number of permutations of size
$r$ from a sequence of $n$ available elements. That is,
$$\perm{n}{r}:=n(n-1)\dotsm (n-r+1).$$

%We say that $f$ is \emph{comparable to} $g$
%%and write $f \approx g$,
%if there exists a positive constant $C$ such that ${C}^{-1} f\leq
%g\leq C f$.
%If $f$ and $g$ are both functions, we assume that $C$ is
%independent of their arguments, unless otherwise specified.

\subsection{The Problem of Hybrid Linear Modeling}
We formulate here a version of the problem of HLM. We will introduce
further restrictions on its setting throughout the paper. Before
presenting the problem we need to define the notions of
$d$-dimensional least squares errors and flats.

If $\mu$ is a Borel probability measure, then the {\it least squares
error} of approximating $\mu$ by a $d$-flat is denoted by $e_2(\mu)$
and defined as follows:
\begin{equation} \label{def:ls_error_mu}
e_2(\mu) :=\sqrt{\inf_{d \text{-flats } F} \int
\dist^2(\mathbf{x},F) \di \mu (\mathbf{x})}.
\end{equation}
Any minimizer of the above quantity is referred to as a {\it least
squares $d$-flat}.

We now incorporate the above definitions and present the problem of
hybrid linear modeling below.
\begin{problem}\label{prob:hlm}
Let $\mu_1, \ldots, \mu_K$ be Borel probability measures and assume
that their $d$-dimensional least square errors
$\{e_2(\mu_k)\}_{k=1}^K$ are sufficiently small and that their least
squares d-flats do not coincide.
%We assume that these underlying measures are separated
%in the following sense: $\mu_i(\supp(\mu_j))=0$, for all $1 \leq i
%\neq j \leq K$.
Suppose a data set
$\mathrm{X}=\{\mathbf{x}_1,\ldots,\mathbf{x}_N\}\subset
\mathbb{R}^D$ generated as follows: For each $k=1,\ldots,K$, $N_k$
points are sampled independently and identically from $\mu_k$, so
that $N=N_1+\cdots+N_K$. The goal of hybrid linear modeling is to
segment $\mathrm{X}$ into $K$ subsets representing the underlying
$d$-flats and simultaneously estimate the parameters of the
underlying flats.
\end{problem}

We remark that the above notion of sufficiently small least square
errors combined with non-coinciding least squares $d$-flats is
quantified for our particular solution later in
Subsection~\ref{subsec:prob_analysis_result} (by restricting the
size of the constant $\alpha$ of equation~\eqref{eq:def_alpha}). We
also remark that we restrict the above setting in
Subsection~\ref{subsec:pol_curv} by requiring the measures $\mu_1,
\ldots, \mu_K$ to be regular and possibly $d$-separated (see
Remark~\ref{rem:d-sep}) and later in
Subsection~\ref{subsec:assume_hlm} by imposing the comparability of
sizes of $N_1, \ldots, N_K$ (see
equation~\eqref{eq:comparable_sizes}).

%For simplicity, we have avoided some technical assumptions when
%defining Problem~\ref{prob:hlm}. In
%Section~\ref{subsec:perturbation_analysis} we restrict the sizes of
%$N_1, \ldots, N_K$ (see equations~\eqref{eq:comparable_sizes}
%and~\eqref{ineq:N_norm}). Additional requirement on the separation
%of the measures $\mu_1, \ldots, \mu_K$
%is introduced in Section~\ref{sec:prob_analysis}. %More general
%possibilities for the notion of concentration around $d$-flats
%%$\mu_1,\ldots, \mu_K$, follow
%follow from~\cite{LW-volume}.
%In particular, we assumed that $\mu_B$
%is uniformly distributed, but~\cite{LW-volume} allows concentration
%around lower dimensional subsets of the given $d$-flat, which are
%sufficiently ``$d$-separated''.
%Hybrid linear modeling with mixed dimensions is considered
%in~\cite{spectral_applied}.

\subsection{The Polar Curvature}
\label{subsec:pol_curv}
For any $d+2$ distinct points $\mathbf{z}_1,\ldots, \mathbf{z}_{d+2}
\in \reals^D$, we denote by
$V_{d+1}(\mathbf{z}_{1},\ldots,\mathbf{z}_{d+2})$ the $(d+1)$-volume
of the $(d+1)$-simplex formed by these points. The polar sine at
each vertex $\mathbf{z}_i, 1\leq i\leq d+2$, is
\begin{equation}\label{eq:polar_sine}
\mathrm{psin}_{\mathbf{z}_i}(\mathbf{z}_{1},\ldots,\mathbf{z}_{d+2}):=
     \frac{(d+1)!\cdot V_{d+1}(\mathbf{z}_{1},\ldots,\mathbf{z}_{d+2})}{\prod_{1\leq j\leq d+2,\,j\ne i}
    \normV{\mathbf{z}_j-\mathbf{z}_i}}.
\end{equation}
\begin{definition}
The polar curvature of $\mathbf{z}_1,\ldots, \mathbf{z}_{d+2}$ is
\begin{equation}
\label{eq:pol_curv} c_\mathrm{p}(\mathbf{z}_1,\ldots,
\mathbf{z}_{d+2}):=
    \diam(\{\mathbf{z}_1,\ldots, \mathbf{z}_{d+2}\})\cdot
    \sqrt{\sum_{i=1}^{d+2}\mathrm{psin}^2_{\mathbf{z}_i}(\mathbf{z}_1,\ldots,
\mathbf{z}_{d+2})}.
\end{equation}
\end{definition}
\begin{remark} \label{rem:other_ls_curvatures}
%We remark that, when $d=0$, the polar curvature coincides with the
%Euclidean distance.
The notion of \textit{curvature} designates here a function of $d+2$
variables generalizing the distance function. Indeed, when $d=0$,
the polar curvature coincides with the Euclidean distance. We use
this name (and probably abuse it) due to the comparability when
$d=1$ of the polar curvature with the Menger curvature multiplied by
the square of the corresponding diameter (see~\cite{LW-part1}).
\end{remark}

Let $\mu$ be a Borel probability measure on ${\reals}^D$. We define
the polar curvature of $\mu$ to be
\begin{equation}\label{def:polar_curv_mu}
c_\mathrm{p}(\mu) := \sqrt{\int
c_\mathrm{p}^2(\mathbf{z}_1,\ldots,\mathbf{z}_{d+2}) \,\di
\mu(\mathbf{z}_1) \ldots \di \mu(\mathbf{z}_{d+2})}.
\end{equation}

The polar curvatures of randomly sampled $(d+1)$-simplices can be
used to estimate the least squares errors of approximating certain
probability measures by $d$-flats. We start with two preliminary
definitions and then state the main result, which is proved
in~\cite{LW-volume} (following the methods
of~\cite{LW-semimetric,LW-part1,LW-part2}).

\begin{definition} We say that a Borel probability measure $\mu$ on
$\reals^D$ is {\it $d$-separated} (with parameters $0<\delta,
\omega<1$) if there exist $d+2$ balls $\{B_i\}_{i=1}^{d+2}$ in
$\reals^D$ with $\mu$-measures at least $\delta$ such that
\[V_d(\mathbf{x}_{i_1}, \ldots, \mathbf{x}_{i_{d+1}})>\omega \cdot
\diam(\supp(\mu))^d,\] for any $\mathbf{x}_{i_k} \in 2 B_{i_k},
1\leq k \leq d+1$ and $1\leq i_1< \cdots < i_{d+1}\leq d+2$.
\end{definition}

\begin{definition} \label{def:sep_reg} We say that a Borel
probability measure $\mu$ on $\reals^D$ is {\it regular} (with
parameters $C_\mu$ and $\gamma$) if there exist constants $\gamma
> 2$ and $C_\mu \geq 1$ such that for any $\mathbf{x} \in \supp(\mu)$
and $0 < r \leq \diam(\supp(\mu))$:
$$\mu(B(\mathbf{x},r)) \leq C_\mu r^\gamma\,.$$
If $D=2$ (or $\supp(\mu)$ is contained in a 2-flat), then one can
allow $1<\gamma \leq 2$ while strengthening the above equation as
follows:
$$C_\mu^{-1} r^\gamma \leq \mu(B(\mathbf{x},r)) \leq C_\mu r^\gamma\,.$$
\end{definition}

\begin{theorem}\label{thm:comparable_cp_lsq}
For any regular and $d$-separated Borel probability measure $\mu$
there exists a constant $C$ (depending only on the $d$-separation
parameters, i.e., $\omega$, $\delta$, and the regularity parameters,
i.e., $\gamma$, $C_\mu$) such that
\begin{equation}
\label{eq:comparable_cp_lsq} C^{-1} \cdot e_2(\mu) \leq
c_\mathrm{p}(\mu) \leq C\cdot e_2(\mu)\,.
\end{equation}
\end{theorem}

The following two curvatures also satisfy
Theorem~\ref{thm:comparable_cp_lsq}~\cite{LW-volume}:
\begin{align*}
c_\mathrm{dls} (\mathbf{z}_1,\ldots,\mathbf{z}_{d+2}) &:=
\sqrt{\inf_{d-\textrm{flats }F} \sum_{1\leq i\leq
d+2}\dist^2(\mathbf{z}_i,F)},\\
c_\mathrm{h} (\mathbf{z}_1,\ldots,\mathbf{z}_{d+2}) &:= \min_{1\leq
i\leq d+2}\dist(\mathbf{z}_i, F_{(i)}),
\end{align*}
where $F_{(i)}$ is the $(d-1)$-flat spanned by all the $d+2$ points
except $\mathbf{z}_i$. In this paper we use $c_\mathrm{p}$ as a
representative of the class of curvatures that satisfy
Theorem~\ref{thm:comparable_cp_lsq}, since it seems computationally
faster than the above two (using the numerical framework described
in~\cite{spectral_applied}). However, all the theory developed in
this paper applies to the rest of the class.

\begin{remark}
\label{rem:d-sep} Since we use Theorem~\ref{thm:comparable_cp_lsq}
in Subsection~\ref{subsec:interpreting_alpha} to justify our
proposed solution for HLM, we need to assume that the measures
$\mu_1, \ldots, \mu_K$ of Problem~\ref{prob:hlm} are regular and
$d$-separated. However, those restrictions could be relaxed or
avoided as follows. If either $c_\mathrm{dls}$ or $c_\mathrm{h}$ is
used instead of $c_\mathrm{p}$, then
Theorem~\ref{thm:comparable_cp_lsq} holds for mere $d$-separated
probability measures (no need for regularity). Moreover, in
Subsection~\ref{subsec:interpreting_alpha} we may only use the right
hand side of equation~\eqref{eq:comparable_cp_lsq}, i.e., the bound
of $c_\mathrm{p}(\mu)$ in terms of $e_2(\mu)$ (though it is
preferable to have a tight estimate as suggested by the full
equation). For such a bound it is enough to assume that $\mu$ is
merely a regular probability measure. If we use instead of
$c_\mathrm{p}$ any of the curvatures $c_\mathrm{dls}$,
$c_\mathrm{h}$, then this latter bound holds for any Borel
probability measure. We also comment that the regularity conditions
described in Definition~\ref{def:sep_reg} could be further relaxed
when replacing $\diam(\{\mathbf{z}_1,\ldots, \mathbf{z}_{d+2}\})$ in
equation~\eqref{eq:pol_curv} with e.g., a geometric mean of
corresponding edge lengths. More details appear in~\cite{LW-volume}.
\end{remark}

\subsection{Affinity Tensors and their Matrix Representations}
\label{subsec:polar_tensor}
Throughout the rest of this paper, we consider $(d+2)$-way tensors
of the form
\[\{\mathcal{A}(i_1,\ldots,i_{d+2})\}_{1 \leq i_1,\ldots,i_{d+2} \leq
N}.\]
%
%We often associate their elements with $d+2$ points in the given
%data $\mathbf{x}_1,\ldots, \mathbf{x}_N \in \reals^D$ (specified in
%Problem~\ref{prob:hlm}), so that we can abuse notation and write
%%
%$$
%\mathcal{A}(i_1,\ldots,i_{d+2}) \equiv
%\mathcal{A}(\mathbf{x}_{i_1},\ldots,\mathbf{x}_{i_{d+2}}) , \quad
%\text{for all}\  1 \leq i_1,\ldots,i_{d+2} \leq N.$$
%
We assume that their elements are between zero and one, and
invariant under arbitrary permutations of the indices
$\{i_1,\ldots,i_{d+2}\}$, i.e., these tensors are super-symmetric.

Most commonly,
%((\mathbf{x}_{i_1},\ldots,\mathbf{x}_{i_{d+2}})_{1 \leq i_1 \leq
%\ldots i_{d+2} \leq N}$ with respect to a given data set
%$\{\mathbf{x}_1,\ldots,\mathbf{x}_N\}$
we form the following affinities using the polar curvature:
%
% use the following polar tensor, whose
%elements are formed with respect to a given data set
%$\{\mathbf{x}_1,\ldots,\mathbf{x}_N\}$:
%
\begin{equation}
\label{eq:affinity_tensor}
\mathcal{A}_\mathrm{p}({i_1},\ldots,{i_{d+2}}) :=
\begin{cases}
e^{-{c_\mathrm{p}(\mathbf{x}_{i_1},\ldots,\mathbf{x}_{i_{d+2}})}/\sigma},&
\text{if ${i_1},\ldots,{i_{d+2}}$ are distinct};\\
0, &\text{otherwise}.
\end{cases}
\end{equation}
The corresponding tensor $\mathcal{A}_\mathrm{p}$ is referred to as
\emph{the polar tensor}.

In the special case of underlying linear subspaces (instead of
general affine ones), we may work with the following $(d+1)$-tensor:
%$\mathcal{A}_\mathrm{p;L}$ obtained from the $(d+2)$-tensor
%$\mathcal{A}_\mathrm{p}$, in the following way:
%
\begin{equation}
\label{eq:affinity_tensor_linear}
\mathcal{A}_\mathrm{p,L}(i_1,\ldots,i_{d+1}):=
%\equiv \mathcal{A}_\mathrm{p;L}(\mathbf{x}_1,\ldots,\mathbf{x}_{d+1}):=
%\mathcal{A}_\mathrm{p}(\mathbf{0},\mathbf{x}_1,\ldots,\mathbf{x}_{d+1})\,.
\begin{cases}
e^{-{c_\mathrm{p}(\mathbf{0},\mathbf{x}_{i_1},\ldots,\mathbf{x}_{i_{d+1}})}/\sigma},&
\text{if ${i_1},\ldots,{i_{d+1}}$ are distinct};\\
0, &\text{otherwise}.
\end{cases}
\end{equation}
In most of the paper we use the $(d+2)$-tensor
$\mathcal{A}_\mathrm{p}$, while in a few places we refer to the
$(d+1)$-tensor $\mathcal{A}_\mathrm{p,L}$.

Given a $(d+2)$-way affinity tensor
$\mathcal{A}\in\mathbb{R}^{N\times N\times \cdots \times N}$ we
unfold it into an $N \times N^{d+1}$ matrix $\mathbf{A}$ in a
similar way as in~\cite{Bader04,Lathauwer00hdsvd}. The
$i^{\text{th}}$ row of $\mathbf{A}$ contains all the elements in the
$i^{\text{th}}$ ``slice'' of $\cal A$:
$\{\mathcal{A}(i,i_2,\ldots,i_{d+2}), 1 \leq i_2,\ldots,i_{d+2} \leq
N\}$, according to an arbitrary but fixed ordering of the last $d+1$
indices $(i_2,\ldots,i_{d+2})$, e.g., the lexicographic ordering.
This ordering (when fixed for all rows) is not important to us,
since we are only interested in the uniquely determined matrix
$\mathbf{W} := \mathbf{A}\mathbf{A}'$ (see Algorithm~\ref{alg:TSCC}
below).

\section{Theoretical Spectral Curvature Clustering}\label{sec:theoretical_algorithm}

We combine Govindu's framework of multi-way spectral
clustering~\cite{Govindu05} with Ng et al.'s spectral clustering
algorithm~\cite{Ng02}, while incorporating the polar affinities
(equation~\eqref{eq:affinity_tensor}), to formulate below
(Algorithm~\ref{alg:TSCC}) the Theoretical Spectral Curvature
Clustering (TSCC) algorithm for solving Problem~\ref{prob:hlm}.
\begin{algorithm}
\caption{Theoretical Spectral Curvature Clustering (TSCC)}
\label{alg:TSCC}
\begin{algorithmic}[1]
\REQUIRE $\mathrm{X}=\{\mathbf{x}_1,\mathbf{x}_2,...,\mathbf{x}_N\} \subset \mathbb{R}^{D}$: data set,\\
$d$: common dimension of flats,\\
$K$: number of $d$-flats,\\
$\sigma$: the tuning parameter for computing $\mathcal{A}$\\
%$\mathcal{A}$: $(d+2)$-way affinity tensor
%
\ENSURE $K$ disjoint clusters $\mathrm{C}_1,\ldots,\mathrm{C}_K$.\\
\hspace{-.27in} \textbf{Steps}:
\STATE \label{step:construct_polar_tensor} Construct the polar
tensor $\mathcal{A}_\mathrm{p}$ using
equation~\eqref{eq:affinity_tensor} and the given $\sigma$.

\STATE \label{step:construct_W} Unfold $\mathcal{A}_\mathrm{p}$ to
obtain the affinity matrix $\mathbf{A}$, and form the weight matrix
$\mathbf{W}:=\mathbf{A}\cdot\mathbf{A}'$.
\STATE \label{step:normalize_W} Compute the degree matrix
$\mathbf{D}:=\diag\{\mathbf{W}\cdot\mathbf{1}_N\}$, and use it to
normalize $\mathbf{W}$ to get
$\mathbf{Z}:=\mathbf{D}^{-{1}/{2}}\cdot\mathbf{W}\cdot\mathbf{D}^{-{1}/{2}}$.
\STATE \label{step:form_U} Find the top $K$ eigenvectors
$\mathbf{u}_1,\mathbf{u}_2,\ldots,\mathbf{u}_K$ of $\mathbf{Z}$ and
define $\mathbf{U}:=[\mathbf{u}_1 \mathbf{u}_2 \ldots
\mathbf{u}_K]\in\mathbb{R}^{N\times K}$.
\STATE \label{step:normalize_U} (optional) Normalize the rows of
$\mathbf{U}$ to have unit length or using other methods (see
Subsection~\ref{subsubsec:U_normalization}).
\STATE \label{step:apply_kmeans} Apply $K$-means~\cite{MacQueen67} to the rows of
$\mathbf{U}$ to find $K$ clusters, and partition the original data
into $K$ subsets $\mathrm{C}_1,\ldots,\mathrm{C}_K$ accordingly.
%
%\end{enumerate}
%
\end{algorithmic}
\end{algorithm}

We refer to this algorithm as theoretical because its complexity and
storage requirement can be rather large (even though polynomial).
In~\cite{spectral_applied} we make the algorithm practical by
applying various numerical techniques. In particular, we suggest a
sampling strategy to approximate the matrix $\mathbf{W}$ in an
iterative way, an automatic scheme of tuning the parameter $\sigma$,
and a straightforward procedure to initialize $K$-means for
clustering the rows of $\mathbf{U}$.

The TSCC algorithm can be seen as two steps of embedding data
%(ignoring the normalizations in Steps~\ref{step:normalize_W}
%and~\ref{step:normalize_U} of Algorithm~\ref{alg:TSCC})
followed by $K$-means. First, each data point $\mathbf{x}_i$ is
mapped to $\mathbf{A}(i,:)$, the $i^\textrm{th}$ row of the matrix
$\mathbf{A}$, which contains the interactions between the point
$\mathbf{x}_i$ and all $d$-flats spanned by any $d+1$ points in the
data (indeed, each column corresponds to $d+1$ data points). Second,
$\mathbf{x}_i$ is further mapped to the $i^\textrm{th}$ row of the
matrix $\mathbf{U}$. The rows of $\mathbf{U}$ are treated as points
in $\mathbb{R}^K$, to which $K$-means is applied.

The question of whether or not to normalize the rows of the matrix
$\mathbf{U}$ is an interesting one. For ease of the subsequent
theoretical development in this paper we do not normalize the rows
of $\mathbf{U}$. Such a choice was also adopted
in~\cite{spectral_applied} where the TSCC algorithm yielded good
numerical results. In Subsection~\ref{subsubsec:U_normalization} we
discuss more carefully the normalization of the matrix $\mathbf{U}$
and show the advantage of such practice. %which also slightly
%improves the numerical results of~\cite{spectral_applied}.

%\begin{remark}
%When $d=0$, The TSCC algorithm is not identical to the algorithm of
%Ng et al.~\cite{Ng02} in two aspects. First, in \cite{Ng02} the
%leading eigenvectors of the affinity matrix are taken to produce the
%matrix $\mathbf{U}$, while our algorithm uses its top left singular
%vectors, which are equivalently the top eigenvectors of the square
%of the affinity matrix. Second, we do not mandate normalization of
%the rows of the matrix $\mathbf{U}$ to have unit length when
%applying $K$-means. The normalization along the rows only keeps the
%angular information, and thus loses all radial information.
%\end{remark}
%The differences between this algorithm and the spectral clustering
%algorithm of Ng et al.~\cite{Ng02}, we do not normalize the rows of
%the matrix $\mathbf{U}$ to have unit length when applying $K$-means.
%Justification of this choice appears in {\bf Subsection....}
%%The
%%normalization along the rows only keeps the angular information, and
%%thus loses all radial information.
%{\bf recall that comparison with
%spectral clustering is in a later section, so other details will be
%put then}

We remark that one can replace the polar tensor (applied in
Step~\ref{step:construct_polar_tensor} of Algorithm~\ref{alg:TSCC})
with other affinity tensors, based on the polar curvature or other ones that satisfy Theorem~\ref{thm:comparable_cp_lsq},
to form different
versions of TSCC. For example, when the underlying subspaces are
known to be linear, one may use the $(d+1)$-tensor
$\mathcal{A}_\mathrm{p,L}$ of
equation~\eqref{eq:affinity_tensor_linear}, forming the Theoretical
Linear Spectral Curvature Clustering (TLSCC) algorithm. Another
example is the following class of affinity tensors that are based on
the powers of the polar curvature:
\begin{equation} \label{eq:affinity_tensor_Lq}
 \mathcal{A}_{\textrm{p},q}(i_1,\ldots,i_{d+2}):=
\begin{cases}
e^{-\frac{c^q_\mathrm{p}\left(\mathbf{x}_{i_1},\ldots,\mathbf{x}_{i_{d+2}}\right)}{\sigma}},
& \textrm{if $i_1,\ldots,i_{d+2}$ are distinct;}\\
0, &\textrm{otherwise},
\end{cases}
\end{equation}
where $q \geq 1$ (see Remark~\ref{rem:interpret_alpha_Lq} for
interpretation). While Algorithm~\ref{alg:TSCC} uses $q=1$, its
formulation in~\cite{spectral_applied} uses $q=2$, as the latter
version of TSCC, when applied in an iterative way, converges faster.

We justify the TSCC algorithm in two steps. In
Section~\ref{sec:perturbation_analysis} we analyze the TSCC
algorithm with a very general tensor (replacing the polar tensor),
and develop conditions under which TSCC is expected to work well. In
particular, the corresponding analysis applies to the polar tensor.
In Section~\ref{sec:prob_analysis} we relate this analysis with the
sampling of Problem~\ref{prob:hlm}, and correspondingly formulate a
probabilistic statement for TSCC. The use of the polar curvature
yields a clear explanation for the statement.

\section{Analysis of TSCC with a General Affinity Tensor}
\label{sec:perturbation_analysis}
Following a strategy of Ng et al.~\cite{Ng02}, we analyze the
performance of the TSCC algorithm with a general affinity tensor
(replacing the polar tensor in
Step~\ref{step:construct_polar_tensor} of Algorithm~\ref{alg:TSCC})
in two steps. First, we define a ``perfect'' tensor representing the
ideal affinities, and show that in such a hypothetical situation,
the $K$ underlying clusters are correctly separated by the TSCC
algorithm. Next, we assume that TSCC is applied with a general
affinity tensor, and control the goodness of clustering of TSCC by
the deviation of the given tensor from the perfect tensor. Finally,
we discuss the effect of the two normalizations in the TSCC
algorithm (Steps~\ref{step:normalize_W} and~\ref{step:normalize_U}
of Algorithm~\ref{alg:TSCC}).
%discuss the effect of normalization of the matrix $\mathbf{W}$,
%i.e.,
%$\mathbf{Z}=\mathbf{D}^{-\frac{1}{2}}\cdot\mathbf{W}\cdot\mathbf{D}^{-\frac{1}{2}}$.

\subsection*{Notational Convenience}
\label{subsec:perturb_setting}
We maintain the common setting of Problem~\ref{prob:hlm} and
all the notation used in the TSCC algorithm.% a data set
%$\mathrm{X}=\{\mathbf{x}_1,\mathbf{x}_2,\ldots,\mathbf{x}_N\}$
%consisting of

We denote the $K$ underlying clusters by
$\widetilde{\mathrm{C}}_1,\ldots,\widetilde{\mathrm{C}}_K$. Each
$\widetilde{\mathrm{C}}_k$ has $N_k$ points, so that $N=\sum_{1\leq
k\leq K}N_k$. For ease of presentation we suppose that $N_1 \leq N_2
\leq \cdots \leq N_K$, and that the points in $\mathrm{X}$ are
ordered according to their membership. That is, the first $N_1$
points of $\mathrm{X}$ are in $\widetilde{\mathrm{C}}_1$, the next
$N_2$ points in $\widetilde{\mathrm{C}}_2$, etc..

We define $K$ index sets $\mathrm{I}_1,\ldots,\mathrm{I}_K$ having
the indices of the points in
$\widetilde{\mathrm{C}}_1,\ldots,\widetilde{\mathrm{C}}_K$
respectively, that is,
\begin{equation}
\label{eq:I_k} \mathrm{I}_k :=\{n\in\mathbb{N}\mid \sum_{1\leq j\leq
k-1} N_j <n\leq \sum_{1\leq j \leq k} N_j\}, \quad \text{for each
}1\leq k \leq K.
\end{equation}

We let $\mathbf{u}^{(i)}$, $1 \leq i \leq N$, denote the
$i^\textrm{th}$ row of $\mathbf{U}$ and $\mathbf{c}^{(k)}$, $1 \leq
k \leq K$, denote the center of the $k^\textrm{th}$ cluster, i.e.,
\begin{equation}
\label{eq:center_k} \mathbf{c}^{(k)} :=
\frac{1}{N_k}\sum_{j\in\mathrm{I}_k}\mathbf{u}^{(j)}.
\end{equation}
%We also recall that the columns of
%$\mathbf{U}$ are denoted by $\mathbf{u}_1,\ldots,\mathbf{u}_K$.
%These notation will be used in the rest of the paper.

\subsection{Analysis of TSCC with the Perfect Tensor}\label{subsec:perfect_tensor}
%The perfect tensor is defined as follows.
We define here the notion of a perfect tensor and show that TSCC
obtains a perfect segmentation with such a tensor.
\begin{definition}
The {\em perfect tensor} associated with Problem~\ref{prob:hlm} is
defined as follows. For any $1\leq i_1,\ldots,i_{d+2}\leq N$,
\begin{equation}
\label{eq:perfect_tensor}
\widetilde{\mathcal{A}}(i_1,\ldots,i_{d+2}):=
    \begin{cases}
    1,& \text{if $\mathbf{x}_{i_1},\ldots,\mathbf{x}_{i_{d+2}}$ are distinct and in the same $\widetilde{\mathrm{C}}_k$};\\
    0,& \text{otherwise}.
    \end{cases}
\end{equation}
\end{definition}

We designate quantities derived from the perfect tensor
$\widetilde{\mathcal{A}}$ (by following the TSCC algorithm) with the
tilde notation, e.g., $\widetilde{\mathbf{A}},
\widetilde{\mathbf{W}}, \widetilde{\mathbf{D}},
\widetilde{\mathbf{Z}}, \widetilde{\mathbf{U}}$.

\begin{remark}
\label{rmrk:ng_ideal} When $d=0$, the perfect tensor
$\widetilde{\mathcal{A}}$ reduces to a block diagonal matrix, with
the blocks corresponding to the underlying clusters. Ng et
al.~\cite{Ng02} also considered an ideal affinity matrix with a
block diagonal structure. However, they maintained the diagonal
blocks computed from the data, while we assume a more extreme case
in which the elements of these blocks are identically one (except at
the diagonal entries). With our assumption it is possible to follow
the steps of TSCC and exactly compute each quantity.
\end{remark}

\begin{figure}[htbp]
     \centering
     \subfigure[data points]{\label{fig:three_clean_lines_data}
          \includegraphics[width=.4\textwidth]{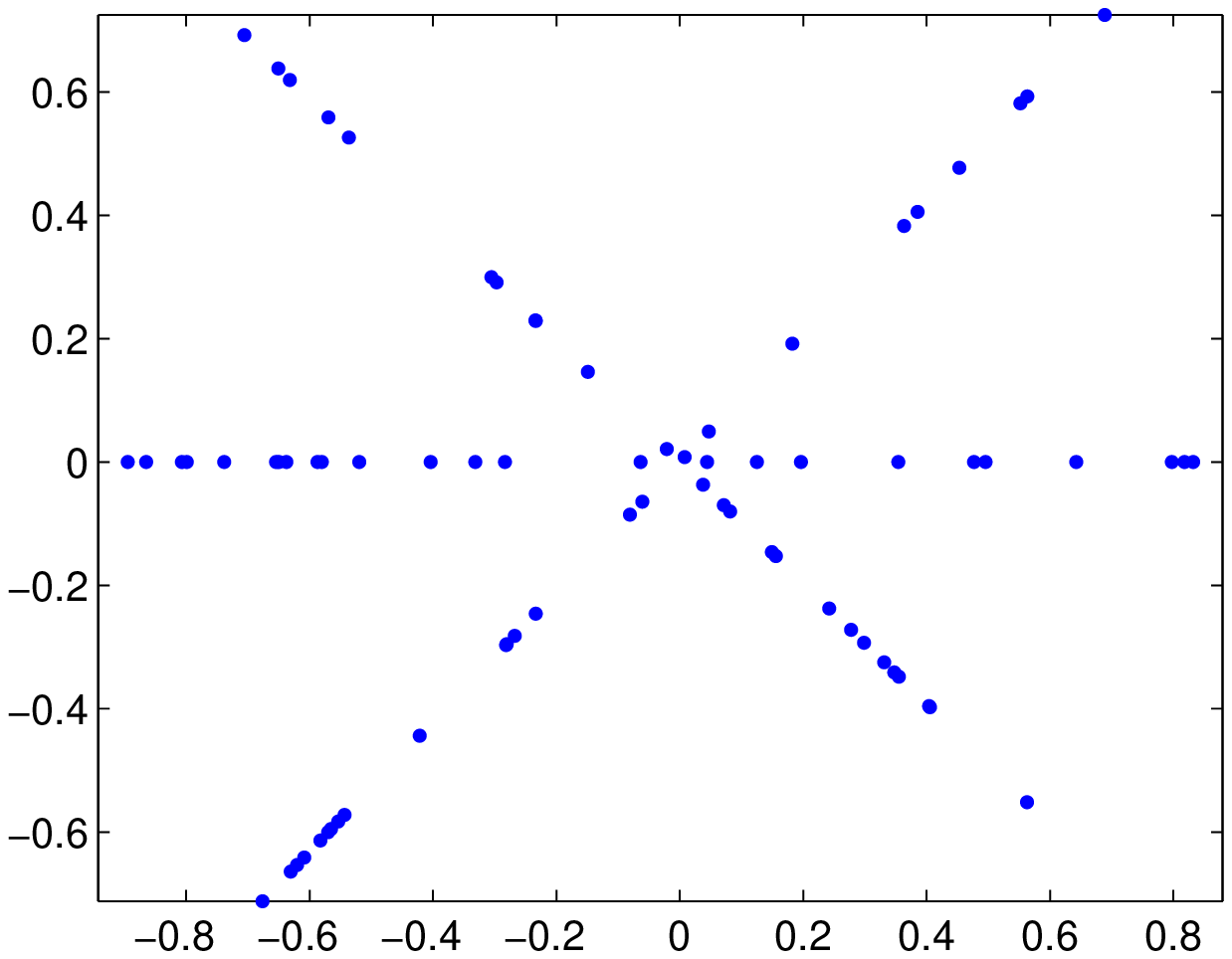}}
     \hspace{.3in}
     \subfigure[eigenvalues of $\mathbf{Z}$]{\label{fig:three_clean_lines_eigvals}
           \includegraphics[width=.4\textwidth]{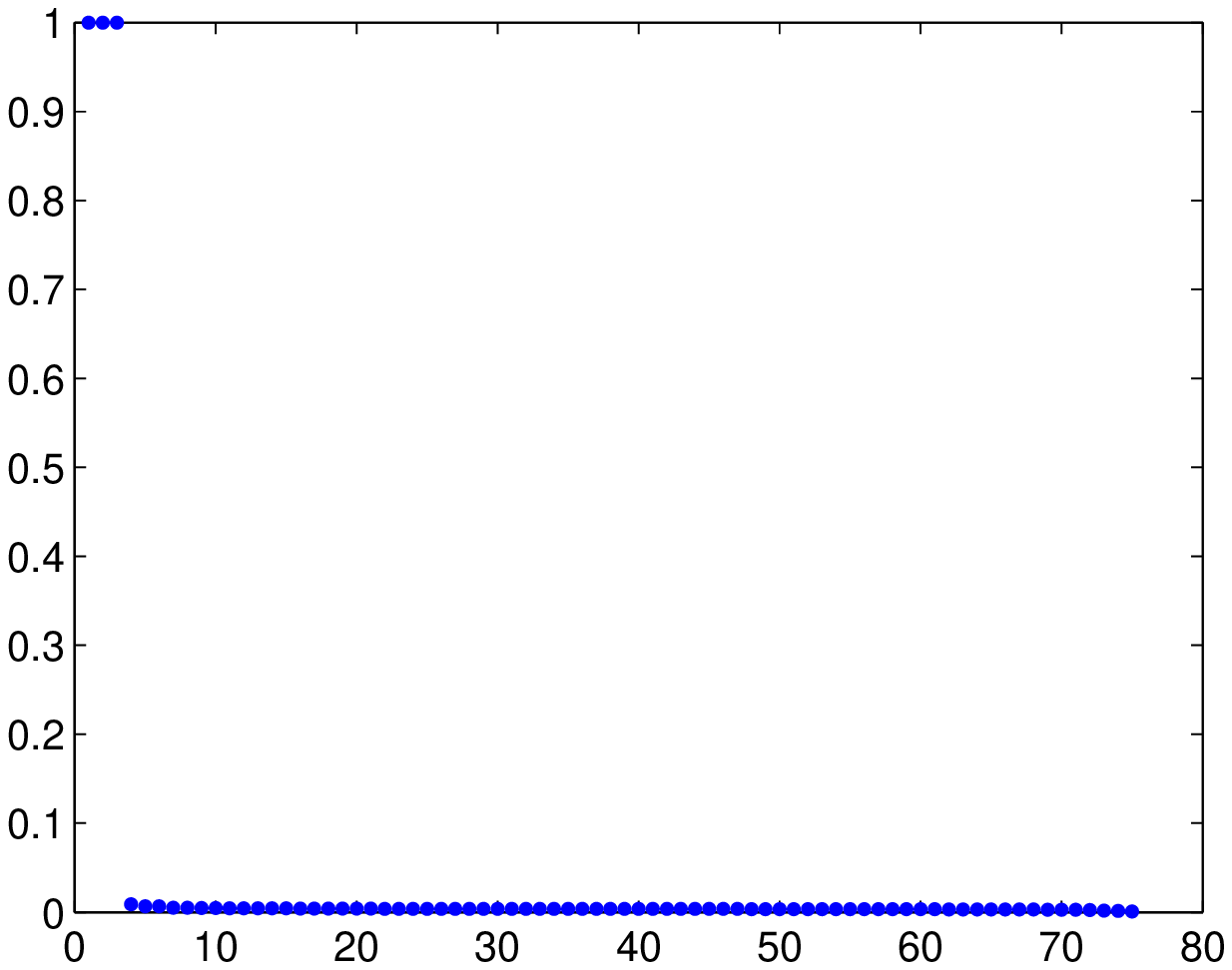}}\\
     \vspace{.3in}
     \subfigure[rows of $\mathbf{U}$]{\label{fig:three_clean_lines_proj}
           \includegraphics[width=.4\textwidth]{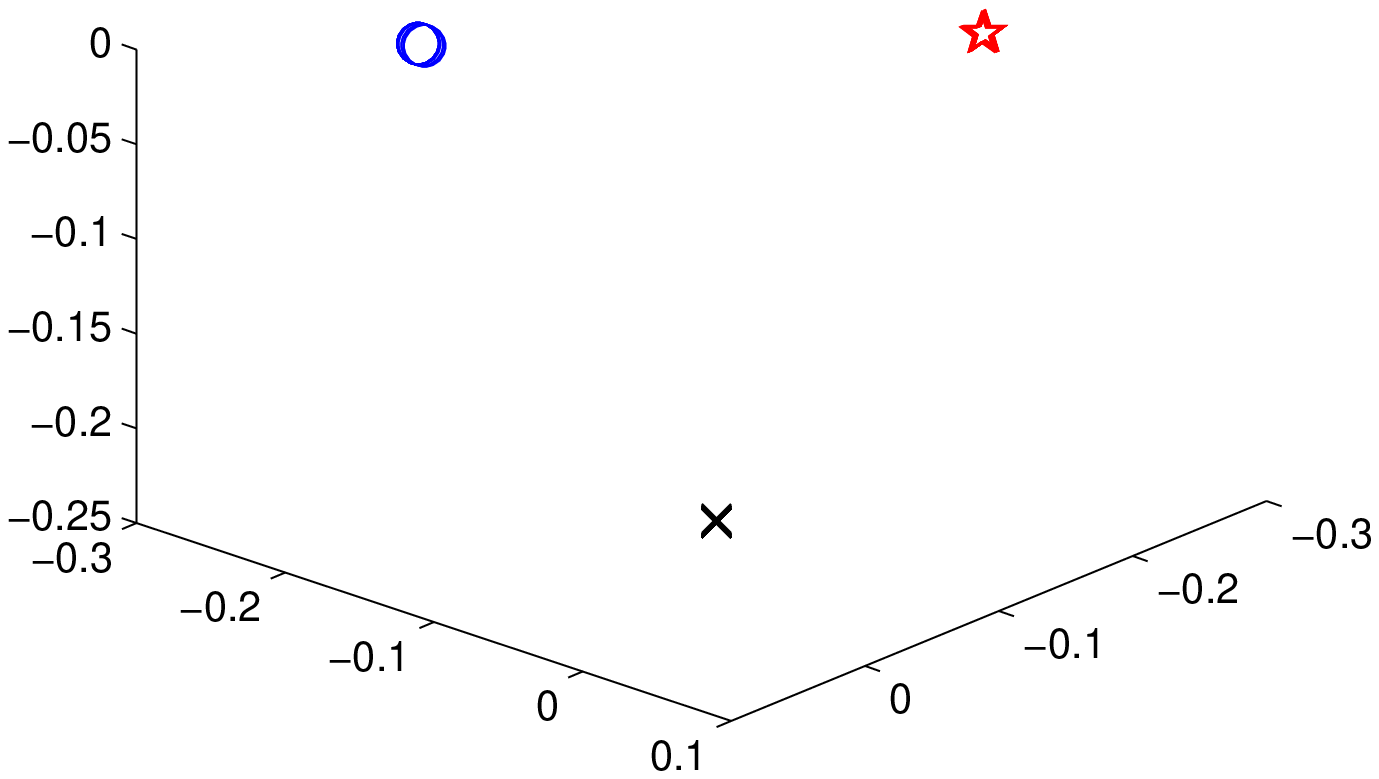}}
     \hspace{.3in}
     \subfigure[detected clusters]{\label{fig:three_clean_lines_clusters}
           \includegraphics[width=.4\textwidth]{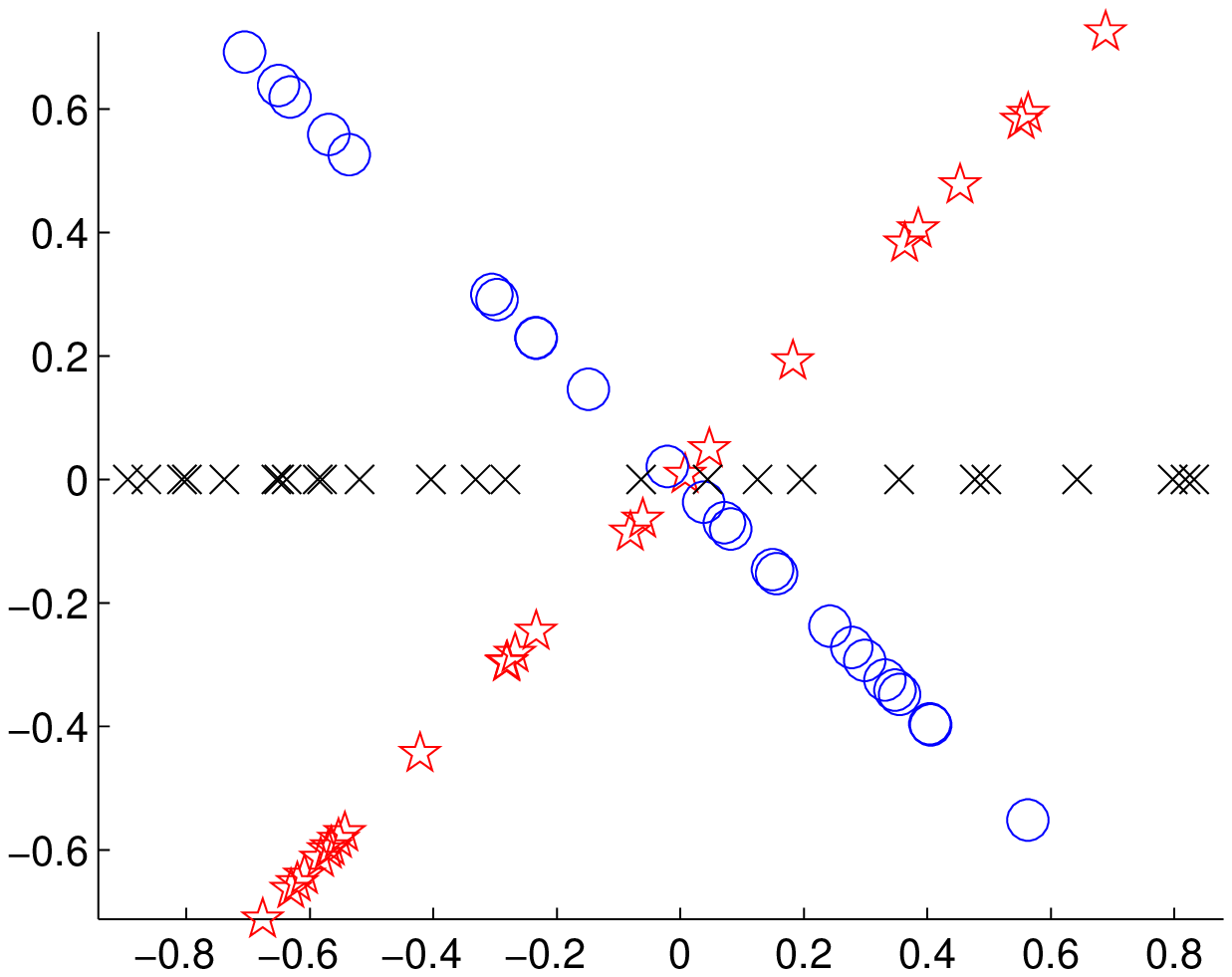}}\\
     \caption{Illustration of the perfect tensor analysis}
     \label{fig:three_clean_lines}
\end{figure}

Our result for TSCC with the perfect tensor
$\widetilde{\mathcal{A}}$ is formulated as follows (see proof in
Appendix~\ref{prf:prop_perfect_tensor}).
\begin{proposition}\label{prop:perfect_tensor}
If $N_k>d+2$ for all $k=1,\ldots,K$, then
\begin{enumerate}
\item $\widetilde{\mathbf{Z}}$ has exactly $K$ eigenvalues of
one; the rest are $\frac{d+1}{(N_k-1)(N_k-d-1)}, 1\leq k \leq K$,
each replicated $N_k-1$ times.
\item The rows of $\mathbf{\widetilde{U}}$ are $K$ mutually
orthogonal vectors in $\mathbb{R}^K$. Moreover, each vector
corresponds to a distinct underlying cluster. %The rows of
%$\mathbf{\widetilde{V}}$ are $K$ mutually orthogonal unit vectors in
%$\mathbb{R}^K$ having the same property.
\end{enumerate}
\end{proposition}
\begin{remark}
For the TLSCC algorithm, the corresponding perfect tensor
$\widetilde{\mathcal{A}}_\mathrm{L}$ is a $(d+1)$-dimensional
equivalent of the $(d+2)$-way tensor $\widetilde{\mathcal{A}}$ of
equation~\eqref{eq:perfect_tensor}.
Proposition~\ref{prop:perfect_tensor} still holds for
$\widetilde{\mathcal{A}}_\mathrm{L}$ but with $d$ replaced by $d-1$.
%The analysis of the TLSCC
%algorithm shows very similar results in the ideal case. The
%eigenvalues of $\widetilde{\mathbf{Z}}_\mathrm{L}$, the equivalent
%of $\widetilde{\mathbf{Z}}$ in TLSCC, are 1 (of multiplicity $K$),
%and $\frac{d}{(N_k-1)(N_k-d)}$ (of multiplicity $N_k-1$), $1\leq k
%\leq K$; the matrix $\widetilde{\mathbf{U}}_\mathrm{L}$ satisfies the same property of
%Proposition~\ref{prop:perfect_tensor}.
\end{remark}
\begin{example} \textbf{Illustration of the perfect tensor analysis}:
We randomly generate three clean lines in $\mathbb{R}^2$ and then
sample 25 points from each line (see
Figure~\ref{fig:three_clean_lines_data}). We then apply TSCC with
the polar tensor of equation~\eqref{eq:affinity_tensor} and $\sigma
= .00001$. The corresponding tensor is a close approximation to the
perfect tensor, because taking the limit of
equation~\eqref{eq:affinity_tensor} as $\sigma\to 0+$ essentially
yields the perfect tensor. Intermediate and final clustering results
are reported in
Figures~\ref{fig:three_clean_lines_eigvals}-\ref{fig:three_clean_lines_clusters}.

In this case, the top three eigenvalues are hardly distinguished
from 1, and the rest are close to zero (see
Figure~\ref{fig:three_clean_lines_eigvals}). The rows of $\mathbf{U}$ accumulate at three
orthogonal vectors (see Figure~\ref{fig:three_clean_lines_proj}), and
thus form three tight clusters, each representing an underlying line
(see Figure~\ref{fig:three_clean_lines_clusters}).
\end{example}

\subsection{Perturbation Analysis of TSCC with a General Affinity Tensor}
\label{subsec:perturbation_analysis}
\subsubsection{Assumptions}
\label{subsec:assume_hlm}

We assume that the underlying clusters have comparable and adequate sizes, more
precisely, there exists a constant $0<\varepsilon_1\leq 1$ such that
\begin{equation}
\label{eq:comparable_sizes} N_k\geq \max\left(\varepsilon_1 \cdot N/K, 2d+3\right), \quad k=1,\ldots, K.
\end{equation}

%We also assume that %$N$ is sufficiently large in the following way
%\begin{equation} \label{ineq:N_norm}
%N > 2(d+1)K / \varepsilon_1,
%\end{equation}
%
%and consequently obtain that
%%
%\be \label{eq:bound_N_k} N_k
%> 2(d+1), \quad \text{ for all } k=1,\ldots,K. \ee

We also assume that all the affinity tensors $\mathcal{A}$ considered in this section are
super-symmetric, and with elements between 0 and 1. Moreover, they satisfy the following condition.
\begin{assumption}\label{assmp:large_D_ii}
There exists a constant $\varepsilon_2>0$ such that
\[
\mathbf{D}\geq \varepsilon_2 \cdot \widetilde{\mathbf{D}}.
%\frac{1}{\widetilde{d}_{k}}\cdot
%\sum_{j\in\mathrm{I}_k}\sum_{k_1,\ldots,k_{d+2}\in\mathrm{I}_k}\mathcal{A}(i,k_1,\ldots,k_{d+2})\mathcal{A}(j,k_1,\ldots,k_{d+2})
%\geq \varepsilon_2,
\]
%for all $i \in \mathrm{I}_k$ and $1\leq k\leq K$.
%Equivalently, we can write $\mathbf{D}\geq \varepsilon_2 \cdot
%\widetilde{\mathbf{D}}$.
\end{assumption}
\begin{remark}
We feel the need to have some lower bound on $\mathbf{D}$, possibly
even weaker than that of Assumption~\ref{assmp:large_D_ii}, to
ensure that the TSCC algorithm would work well. Indeed, for each
$i\in \mathrm{I}_k, 1\leq k\leq K$, the sum $\sum_{j\in
\mathrm{I}_k} W_{ij}$ measures the ``connectedness'' between the
point $\mathbf{x}_i$ and the other points in
$\widetilde{\mathrm{C}}_k$, and thus should be sufficiently large.
Accordingly, since $D_{ii}\geq \sum_{j\in \mathrm{I}_k} W_{ij}, i\in
\mathrm{I}_k, 1\leq k\leq K$, these diagonal entries of the matrix
$\mathbf{D}$ should be correspondingly large as well. In
Subsection~\ref{subsubsec:existence_assmp1} we discuss the existence
of this condition for the polar tensor while taking into account the
restrictions on the tuning parameter $\sigma$ implied by
Theorem~\ref{thm:good_with_high_prob}.
\end{remark}

\subsubsection{Measuring Goodness of Clustering of the TSCC Algorithm}
We use two equivalent ways to quantify the goodness of clustering of
the TSCC algorithm when applied with a general affinity tensor
$\mathcal{A}$. In Subsection~\ref{subsubsec:U_normalization} we
relate them to the more absolute notion of clustering identification
error.

We first investigate each of the $K$ underlying clusters in the
$\mathbf{U}$ space, i.e., $\{\mathbf{u}^{(i)}\}_{i\in \mathrm{I}_k},
1\leq k \leq K$, and estimate the sum of their variances. We refer
to this sum as the total variation of the matrix $\mathbf{U}$.
\begin{definition}
The total variation of $\mathbf{U}$ (with respect to the $K$
underlying clusters) is
\begin{align}\label{eq:total_variance}
\mathrm{TV}(\mathbf{U}) := \sum_{1\leq k\leq K}
\sum_{i\in\mathrm{I}_k}\normV{\mathbf{u}^{(i)}-\mathbf{c}^{(k)}}^2,
\end{align}
where $\mathbf{c}^{(1)},\ldots,\mathbf{c}^{(K)}$ are the centers of
the underlying clusters in the $\mathbf{U}$ space (see
equation~\eqref{eq:center_k}).
\end{definition}

The smaller the total variation $\mathrm{TV}(\mathbf{U})$ is, the
more concentrated the underlying clusters in the $\mathbf{U}$ space
are. In fact, the following lemma (proved in
Appendix~\ref{prf:lem_separation_leq_tv}) implies that the smaller
$\mathrm{TV}(\mathbf{U})$ is, the more separated the centers are
from the origin and from each other.
\begin{lemma}\label{lem:separation_leq_tv}
\begin{gather}
\sum_{1\leq k\leq K} N_k \cdot
\normV{\mathbf{c}^{(k)}}^2=K-\tv, \label{eq:norm_centers_tv}\\
\sum_{1\leq k<\ell \leq K} N_k N_\ell \cdot \langle
\mathbf{c}^{(k)},\mathbf{c}^{(\ell)} \rangle ^2 \leq \tv.
\label{ineq:separation_between_centers}
%\dist^2(\mathrm{S}(\mathbf{U}),\mathrm{S}(\widetilde{\mathbf{U}})).
\end{gather}
\end{lemma}

The other measurement of the goodness of clustering of TSCC is
motivated by the fact that, in the ideal case, the subspace spanned
by the top $K$ eigenvectors of $\widetilde{\mathbf{Z}}$,
$E_K(\widetilde{\mathbf{Z}})$, leads to a perfect segmentation (see
Proposition~\ref{prop:perfect_tensor}). When given a general
affinity tensor $\mathcal{A}$, the eigenspace $E_K(\mathbf{Z})$
determines the clustering result of TSCC. We thus suggest to measure
the discrepancy between these two eigenspaces, $E_K(\mathbf{Z})$ and
$E_K(\widetilde{\mathbf{Z}})$, by comparing the orthogonal
projectors onto them, $P^K(\mathbf{Z})$ and
$P^K(\widetilde{\mathbf{Z}})$, in the following way.
%Due to the randomness of an
%orthonormal transformation associated to the eigenvectors, neither
%of the columns of $\widetilde{\mathbf{U}}$ nor its rows can be
%uniquely determined. However, the column space of
%$\widetilde{\mathbf{U}}$ is always fixed.
%Let $\mathbf{U}$ be
%obtained from a given affinity tensor $\mathcal{A}$ by following the
%TSCC algorithm. Similarly, only the column space of the matrix
%${\mathbf{U}}$ is unique.
%We will thus compare
%$\mathrm{S}(\mathbf{U})$ with $\mathrm{S}(\widetilde{\mathbf{U}})$
%to analyze the goodness of separation of the TSCC algorithm with the
%given tensor $\mathcal{A}$.
%
%We actually notice that the columns subspaces of
%$\widetilde{\mathbf{U}}$ and $\mathbf{U}$ are equal to the subspaces
%spanned by the top $K$ eigenvectors of $\widetilde{\mathbf{Z}}$ and
%$\mathbf{Z}$ respectively. We thus formulate the goodness of
%clustering using the latter subspaces in the following way.
%
\begin{definition}
%\label{def_good} If
%$\{\mathbf{x}_i\}_{i=1}^N$ is a data sampled
%according to the formulation of Problem...,
%$\mathcal{A}$ is an
%affinity tensor, %defined for any $d+2$ of this data ,
%$\tilde{\mathcal{A}}$ is the corresponding perfect tensor, and
%$P^K(\mathbf{Z})$, $P^K(\widetilde{\mathbf{Z}})$ are the .... . Then
%the goodness of clustering of TSCC, when applied with the tensor
%$\mathcal{A}$, is defined by
%
The distance between the two subspaces $E_K(\widetilde{\mathbf{Z}})$
and $E_K(\mathbf{Z})$ is
\begin{equation}\label{eq:dist_projectors}
\dist(E_K(\mathbf{Z}),E_K(\widetilde{\mathbf{Z}})) :=
\normF{P^K(\mathbf{Z})-P^K(\widetilde{\mathbf{Z}})}.
\end{equation}
\end{definition}
A geometric interpretation of the above distance is provided in the
following lemma using the notion of \emph{principal angles}~\cite{Golub96}.
%Roughly speaking, for two subspaces of the
%same dimension $K$, the principal angles are the $K$ smallest
%possible angles between all pairs of vectors in the two subspaces,
%and the $K$ vectors in each subspace are mutually orthogonal.
We review the definition of principal angles and also prove
Lemma~\ref{lem:dist_principal_angles} in
Appendix~\ref{subsec:review_principal_angles}.
\begin{lemma}\label{lem:dist_principal_angles}
%Following the setting of Definition~\ref{def_good}, let
Let $0\leq\theta_1\leq \theta_2\leq \cdots \leq \theta_K\leq \pi/2$
be the $K$ principal angles between the two subspaces
$E_K(\mathbf{Z})$ and $E_K(\widetilde{\mathbf{Z}})$. Then
\begin{equation}\label{eq:dist_principal angles}
\dist^2(E_K(\mathbf{Z}),E_K(\widetilde{\mathbf{Z}}))= 2\cdot
\sum_{k=1}^{K}\sin^2\theta_k.
\end{equation}
\end{lemma}

At last, we claim that the above two ways of measuring the goodness
of clustering of TSCC are equivalent in the following sense (see
proof in Appendix~\ref{prf:lem_dist_tv}).
\begin{lemma}\label{lem:dist_tv}
%If TSCC is applied with a general tensor $\mathcal{A}$, then
% Let
%$\mathbf{U}$ be derived from an arbitrary affinity tensor
%$\mathcal{A}$ by following the TSCC algorithm. Then
\begin{equation}\label{eq:total_variance_of_rows_U}
\dist^2(E_K(\mathbf{Z}),E_K(\widetilde{\mathbf{Z}})) = 2\cdot \tv.
\end{equation}
%and
\end{lemma}

\subsubsection{The Perturbation Result}
Given a general affinity tensor $\mathcal{A}$ we quantify its
deviation from the perfect tensor $\widetilde{\mathcal{A}}$ by the
difference
$$\mathcal{E}:=\mathcal{A}-\widetilde{\mathcal{A}}.$$
Our main result shows that the magnitude of this perturbation
controls the goodness of clustering of the TSCC algorithm.
\begin{theorem}
\label{thm:perturbation_main} Let $\mathcal{A}$ be any affinity
tensor satisfying Assumption~\ref{assmp:large_D_ii} and
$\mathcal{E}$ its deviation from the perfect tensor. There exists a
constant $C_1=C_1(K,d,\varepsilon_1,\varepsilon_2)$ (estimated in
equation~\eqref{eq:constant_C1} of
Appendix~\ref{prf:thm_perturbation_main}) such that if
\[N^{{-(d+2)}}\normF{\mathcal{E}}^2\leq \frac{1}{8 C_1},\]
then
\begin{align}\label{eq:pert_result}
\tv &\leq C_1\cdot N^{{-(d+2)}}\normF{\mathcal{E}}^2.
\end{align}
\end{theorem}

%The following corollary is a direct consequence of
%Lemma~\ref{lem:var_proj} and Theorem~\ref{thm:perturbation_main}.
%%
%\begin{theorem}
%\label{thm:perturbation_rows_1} If either
%Assumption~\ref{assmp:large_D_ii} or
%Assumption~\ref{assmp:weaker_than_d_ii} holds, then
%\begin{align}\label{eq:pert_result_rows_1}
%%\norm{P^K(\mathbf{Z})-P^K(\widetilde{\mathbf{Z}})}_F&\leq
%\mathrm{TV}(\mathbf{U}) &\leq 1/2 \cdot
%C(K,d,\varepsilon_1,\varepsilon_2)\cdot
%N^{-(d+2)}\norm{\mathcal{E}}^2_F.
%\end{align}
%That is, if $N$ is large and $\norm{\mathcal{E}}_F =
%o\big(N^{(d+2)/2}\big)$, then the underlying clusters are tight
%point clouds concentrating around orthogonal vectors in
%$\mathbb{R}^K$. Therefore, we can still expect good clustering
%result.
%\end{theorem}

\begin{remark}
For the TLSCC algorithm, Theorem~\ref{thm:perturbation_main} holds
with $d$ replaced by $d-1$.
\end{remark}

\begin{figure}[htbp]
     \centering
     \subfigure[data points]{\label{fig:three_noisy_lines_data}
          \includegraphics[width=.4\textwidth]{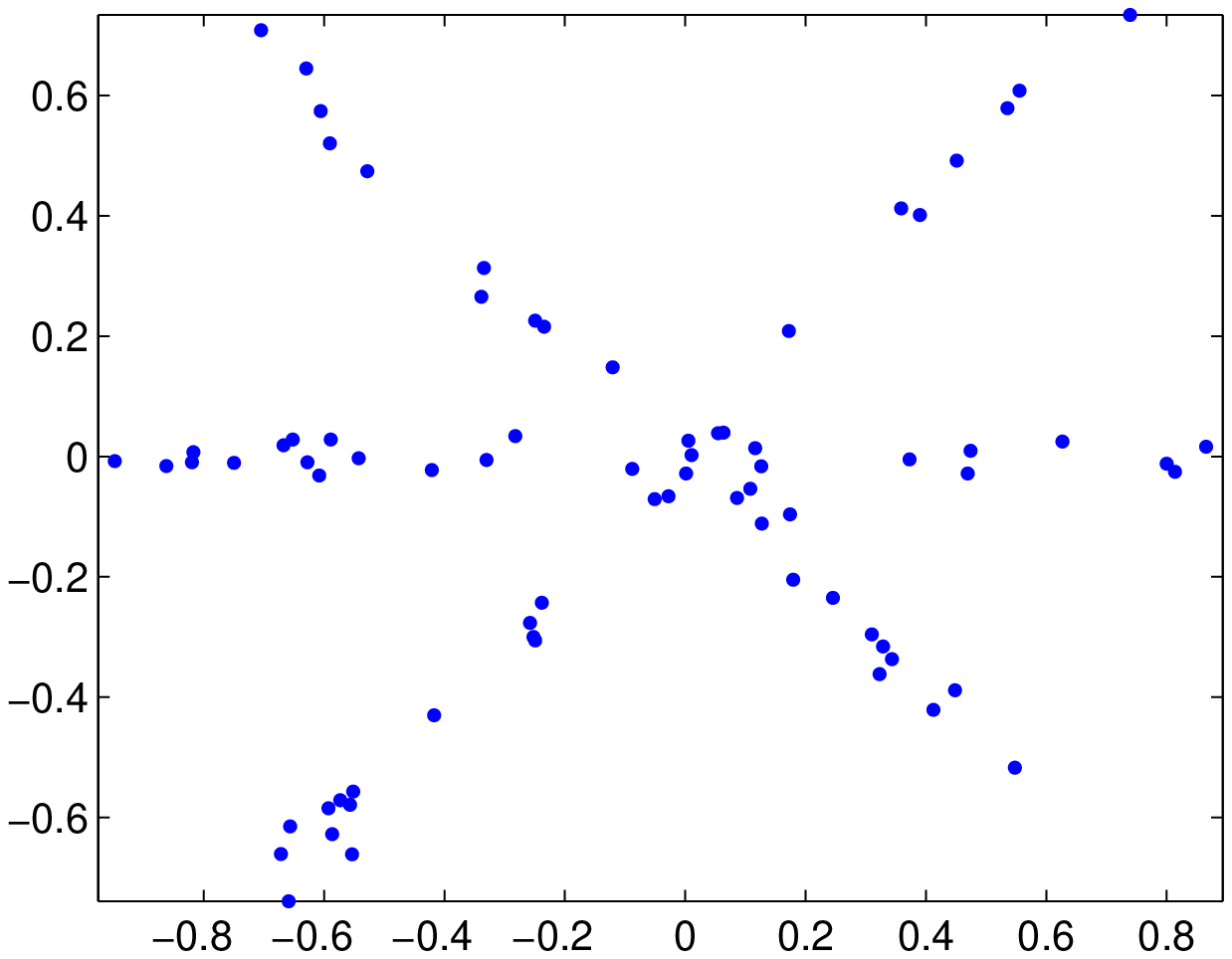}}
     \hspace{.3in}
     \subfigure[eigenvalues of $\mathbf{Z}$]{\label{fig:three_noisy_lines_eigvals}
           \includegraphics[width=.4\textwidth]{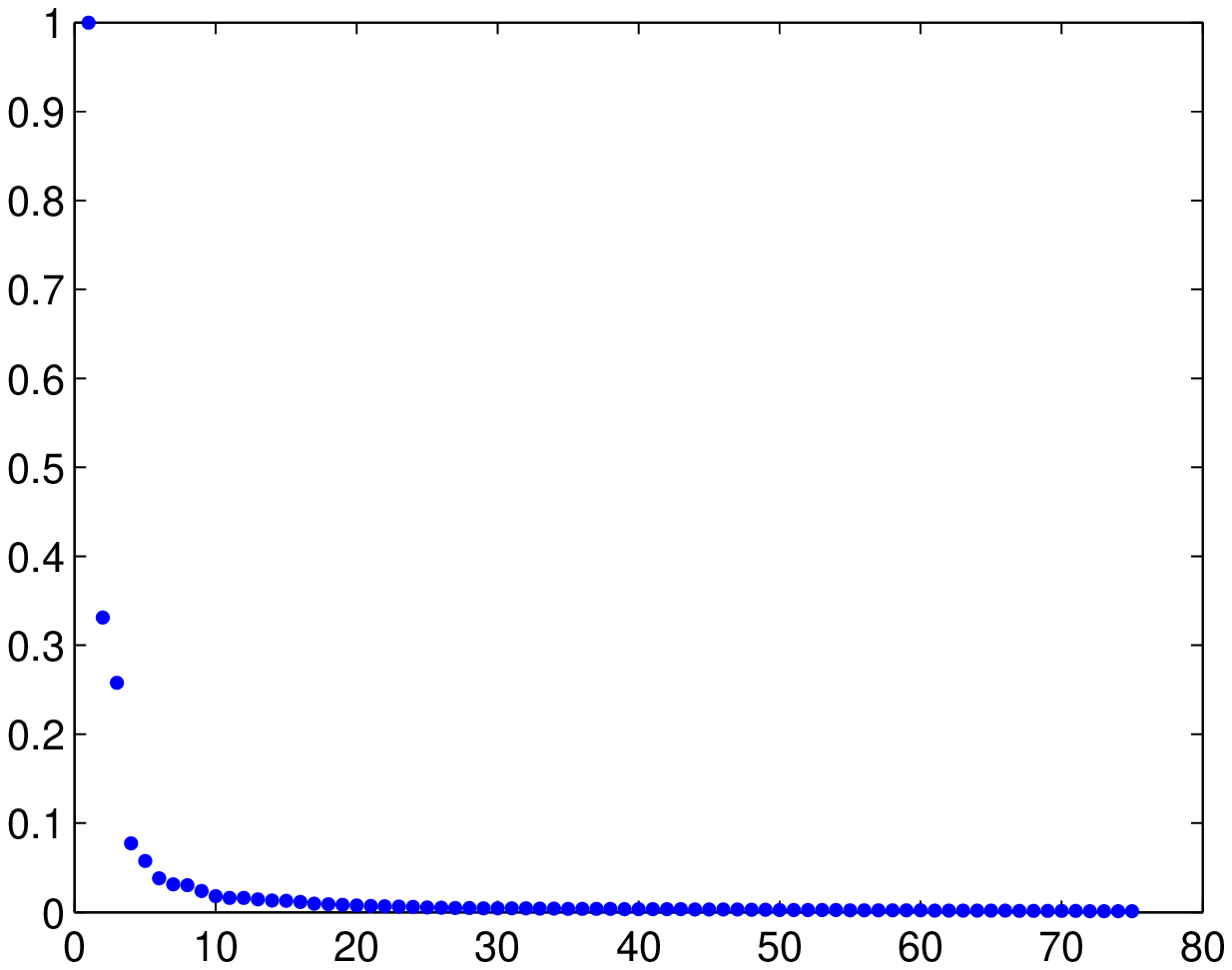}}
     \vspace{.3in}
     \subfigure[rows of $\mathbf{U}$]{\label{fig:three_noisy_lines_proj}
           \includegraphics[width=.4\textwidth]{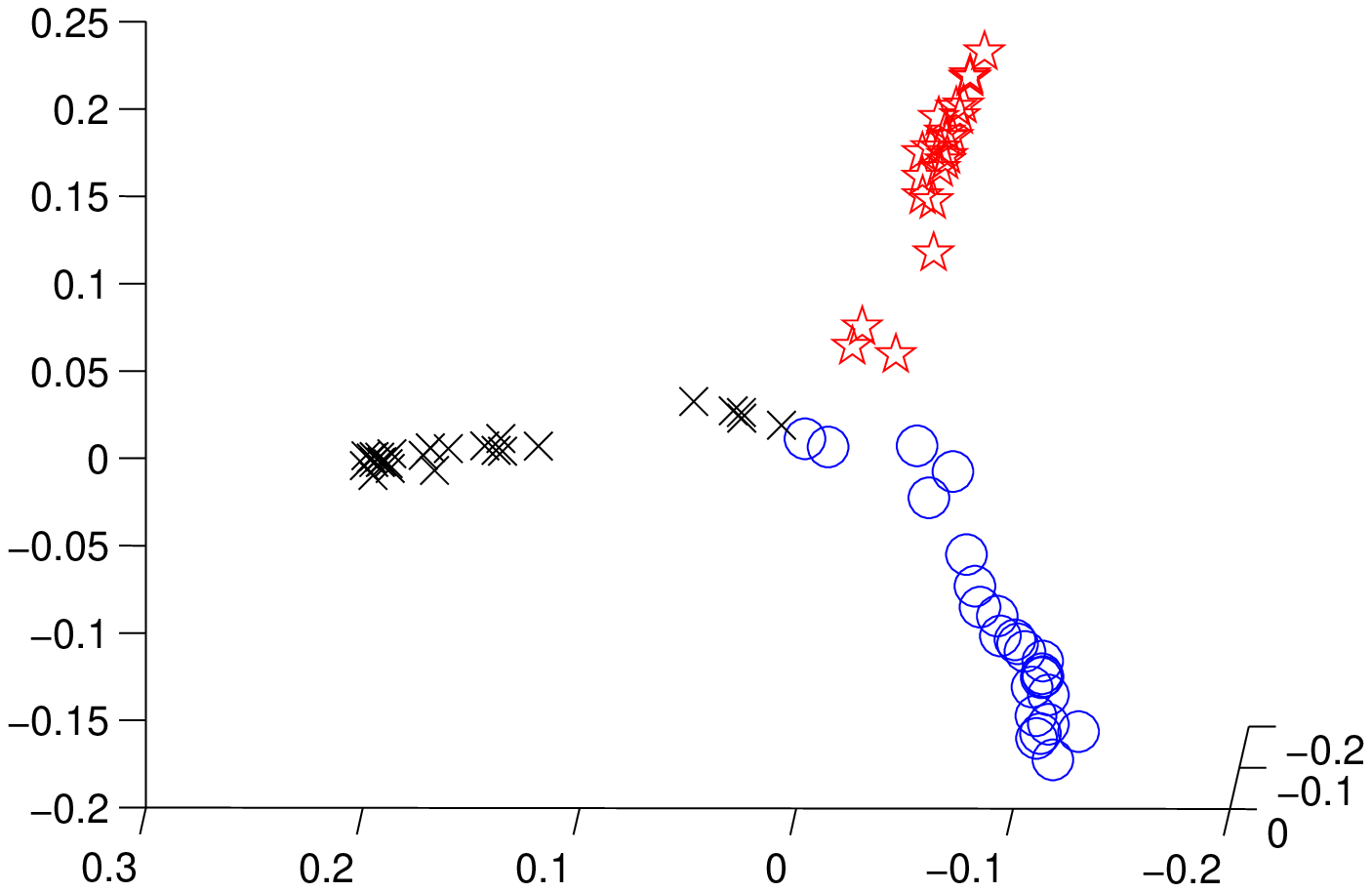}}
     \hspace{.3in}
     \subfigure[detected clusters]{\label{fig:three_noisy_lines_clusters}
           \includegraphics[width=.4\textwidth]{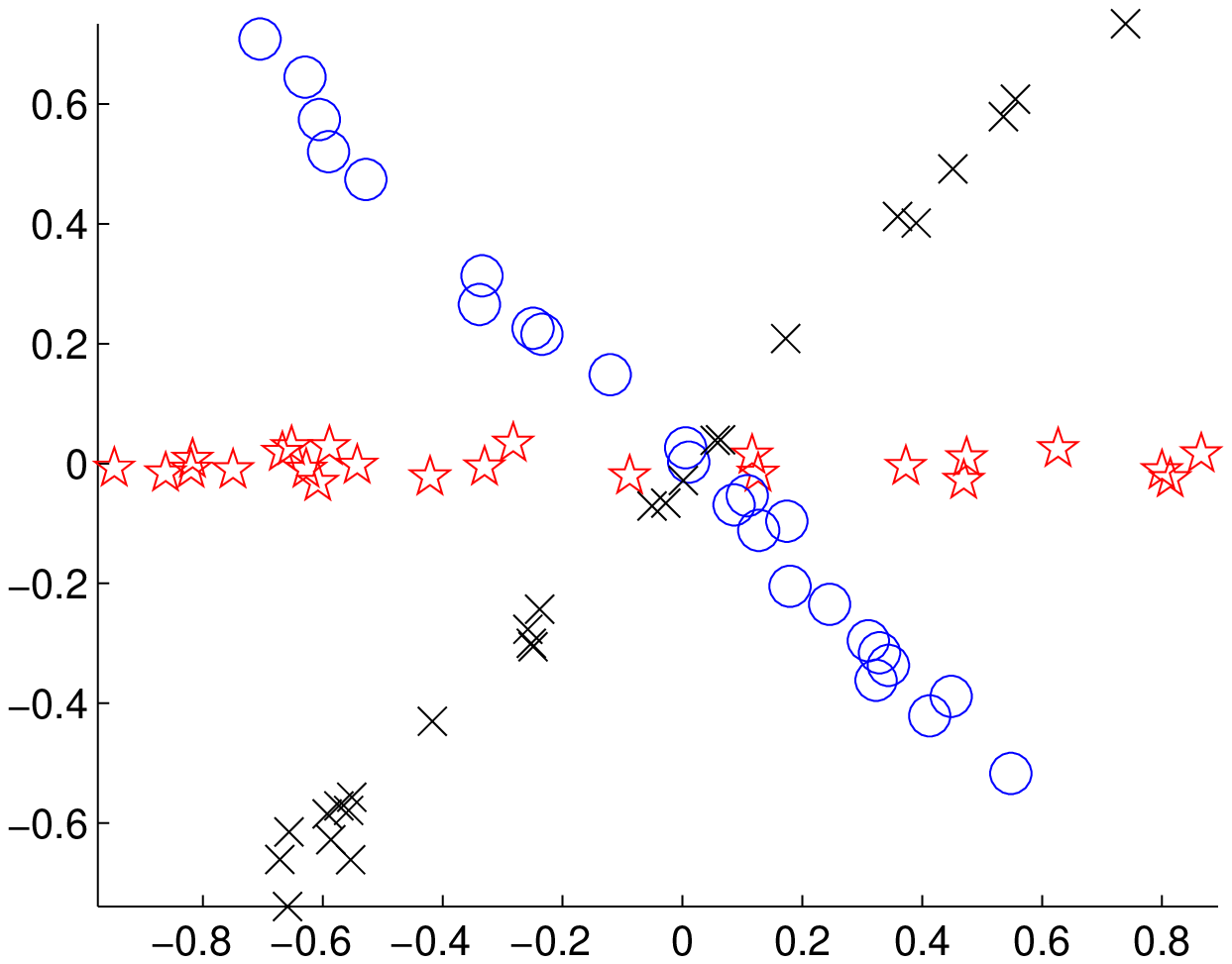}}
     \caption{Illustration of the perturbation analysis}
     \label{fig:three_noisy_lines}
\end{figure}

\begin{example}
\textbf{Illustration of the perturbation analysis}: We corrupt the
data in Figure~\ref{fig:three_clean_lines} with $2.5\%$ additive
Gaussian noise (see Figure~\ref{fig:three_noisy_lines_data}), and
apply TSCC with the polar tensor of
equation~\eqref{eq:affinity_tensor} and $\sigma = 0.1840$. In this
case of moderate noise, the top three eigenvalues are still nicely
separated from the rest, even though two of them deviate from 1 (see
Figure~\ref{fig:three_noisy_lines_eigvals}). The rows of
$\mathbf{U}$ still form three clear clusters, but they deviate from
concentrating at exactly three orthogonal vectors (see
Figure~\ref{fig:three_noisy_lines_proj}). The underlying clusters
are detected correctly, except possibly for a few points at their
intersection (see Figure~\ref{fig:three_noisy_lines_clusters}).
\end{example}

\subsection{The Effects of the Normalizations in TSCC}
\subsubsection{Possible Normalizations of $\mathbf{U}$ and Their Effects on Clustering}
\label{subsubsec:U_normalization} The analysis of the previous
subsections uses the embedding represented by the rows of $\mathbf{U}$.
It is possible to normalize these rows (e.g., by their lengths as
in~\cite{Ng02}) before applying $K$-means. In the following we
consider two normalized versions of the rows of $\mathbf{U}$, and
analyze their effects on the TSCC algorithm (in comparison with the rows of $\mathbf{U}$).

\begin{figure}\label{fig:UTV}
\centering
    \subfigure[The underlying clusters in the $\mathbf{U, T, V}$ spaces
    respectively]{
    \label{fig:UTV_truth}
        \includegraphics[width=.33\textwidth]{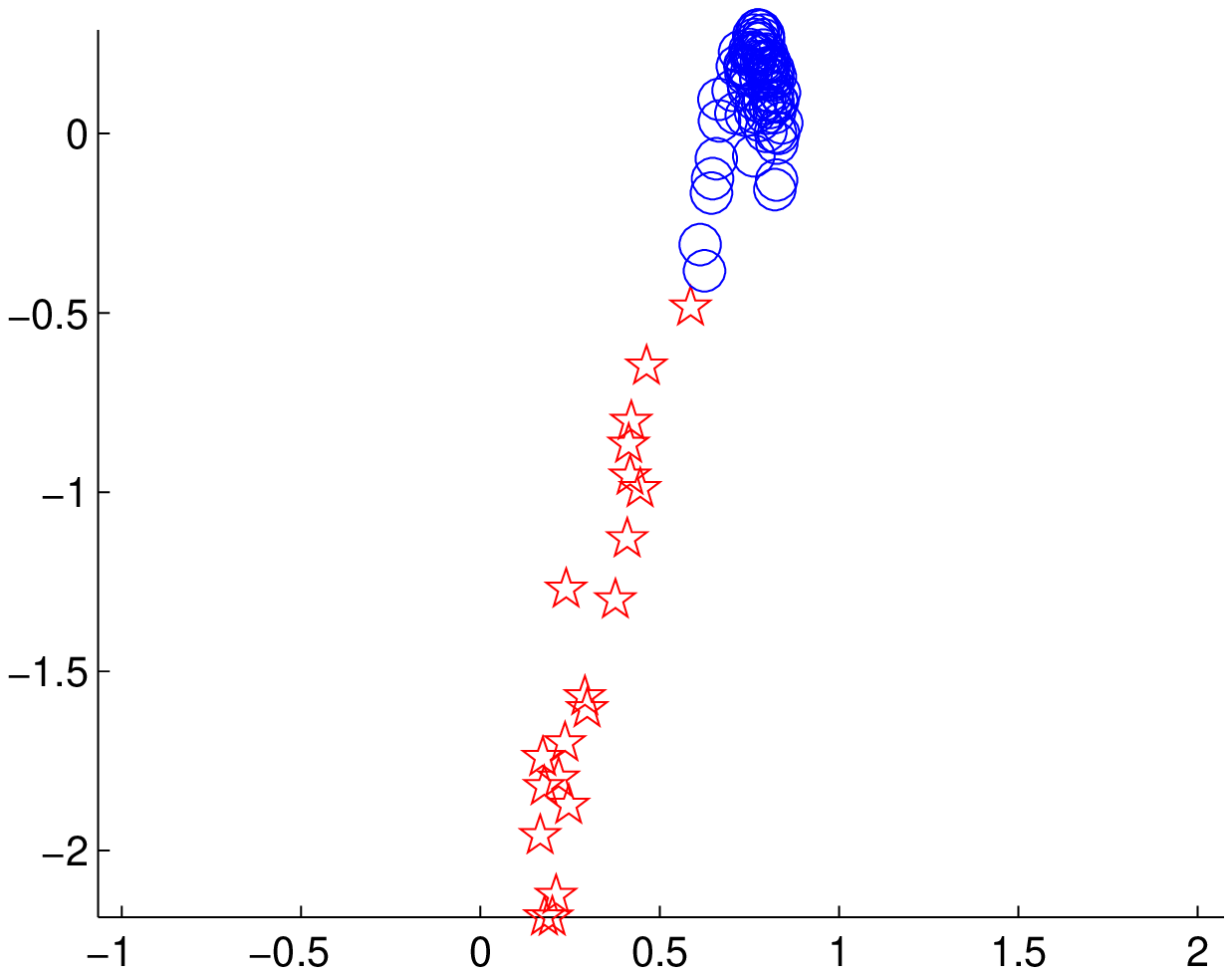}
        \includegraphics[width=.33\textwidth]{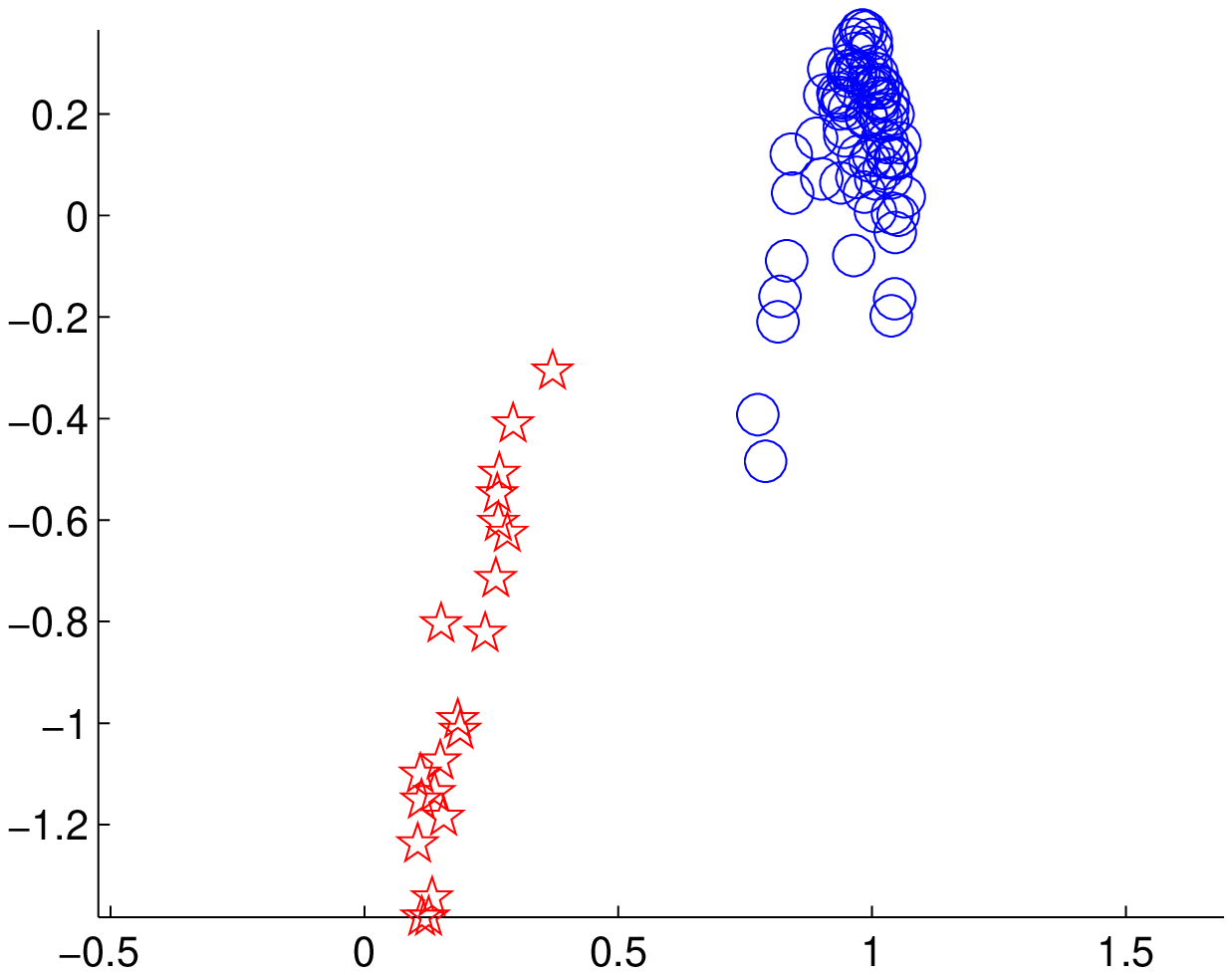}
        \includegraphics[width=.33\textwidth]{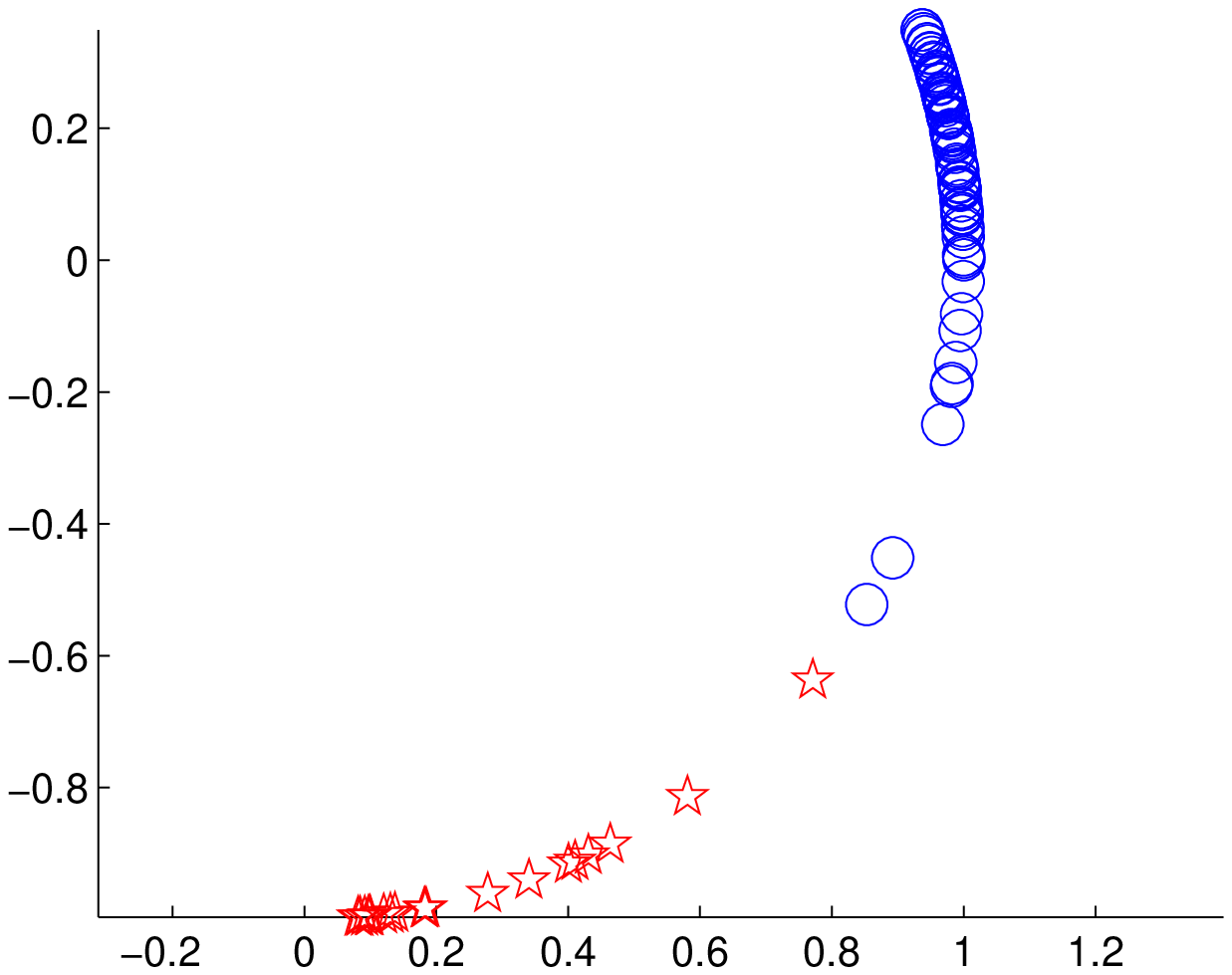}
    }
    \subfigure[The clusters found by $K$-means in the $\mathbf{U, T, V}$ spaces]{
    \label{fig:UTV_kmeans}
        \includegraphics[width=.33\textwidth]{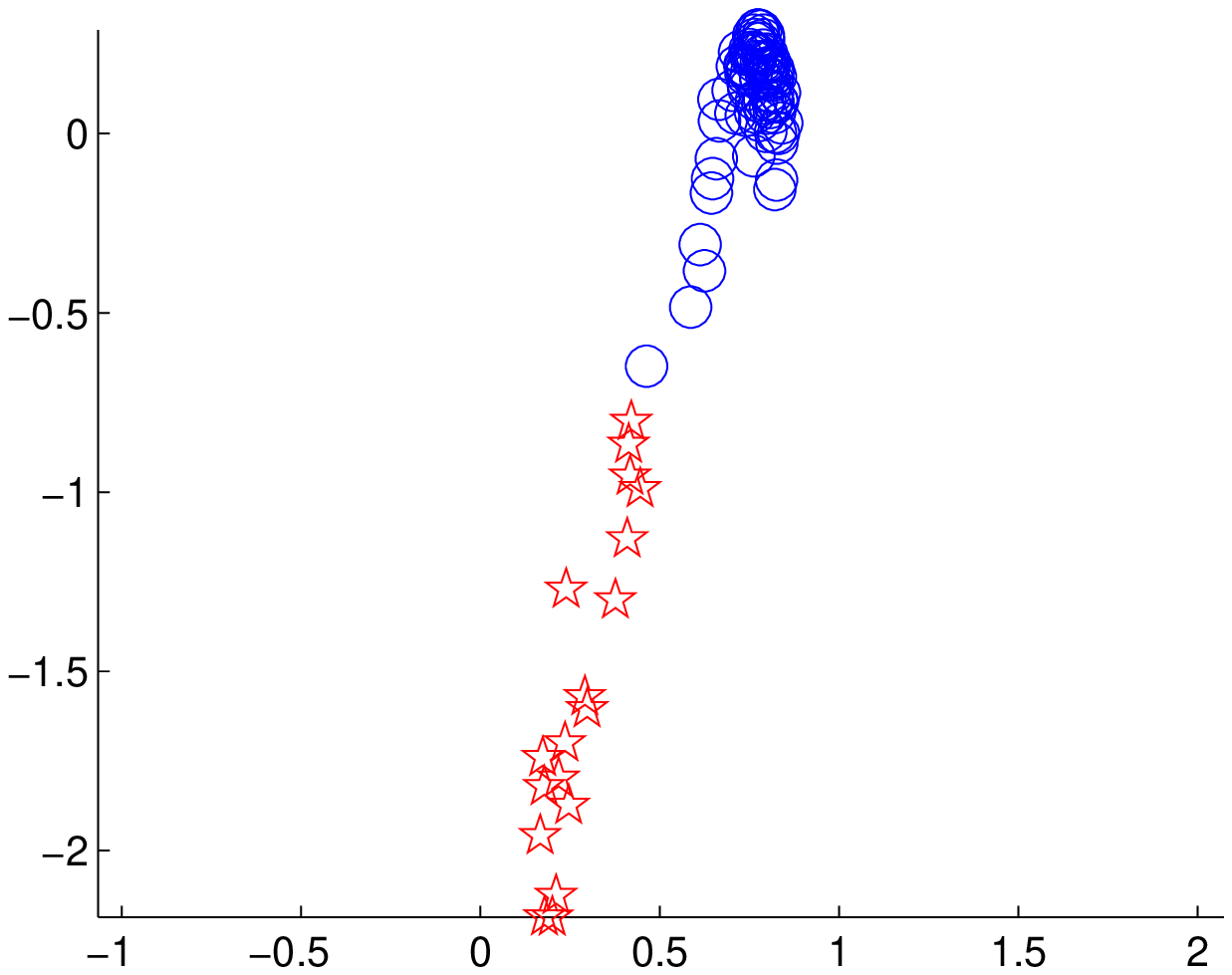}
        \includegraphics[width=.33\textwidth]{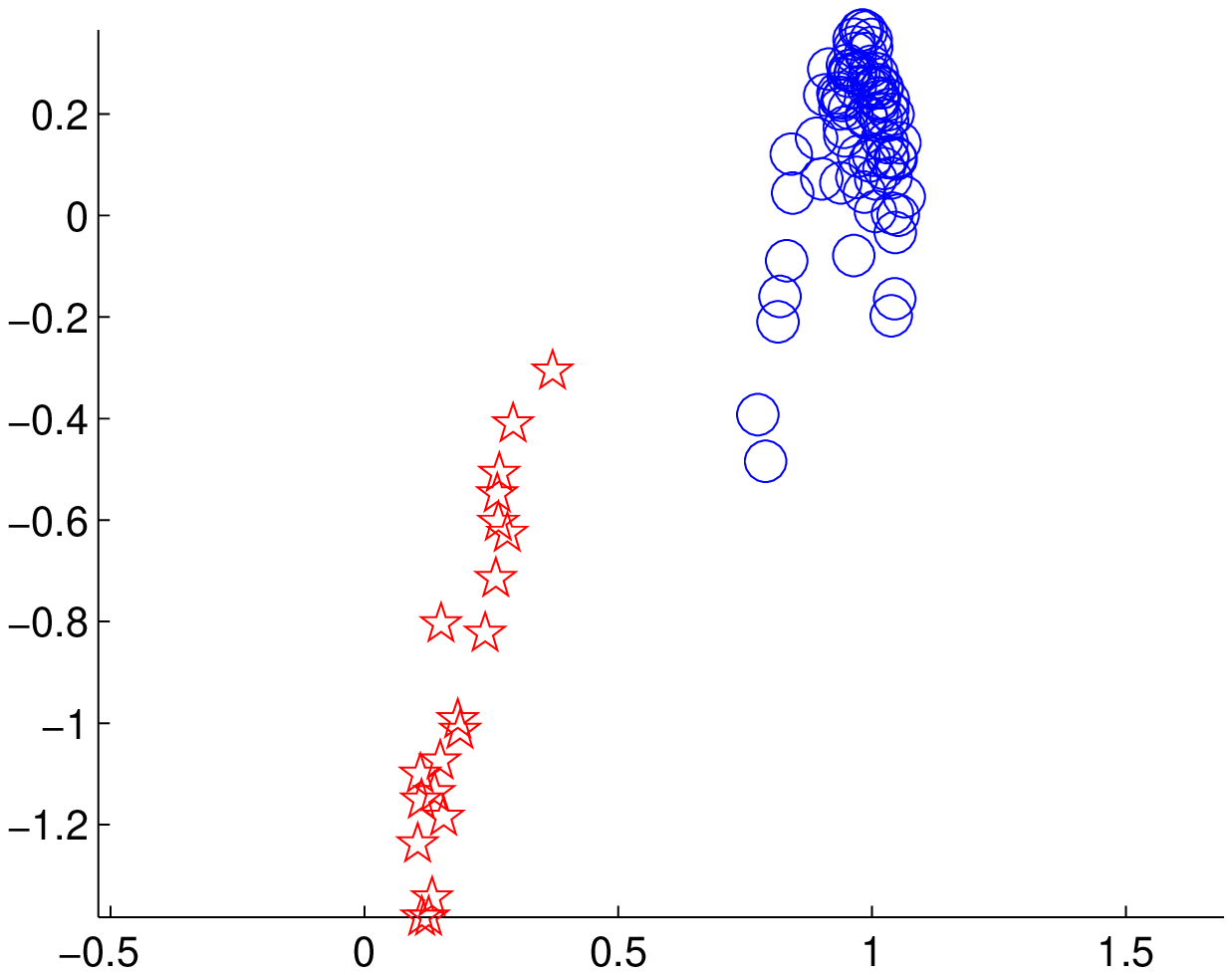}
        \includegraphics[width=.33\textwidth]{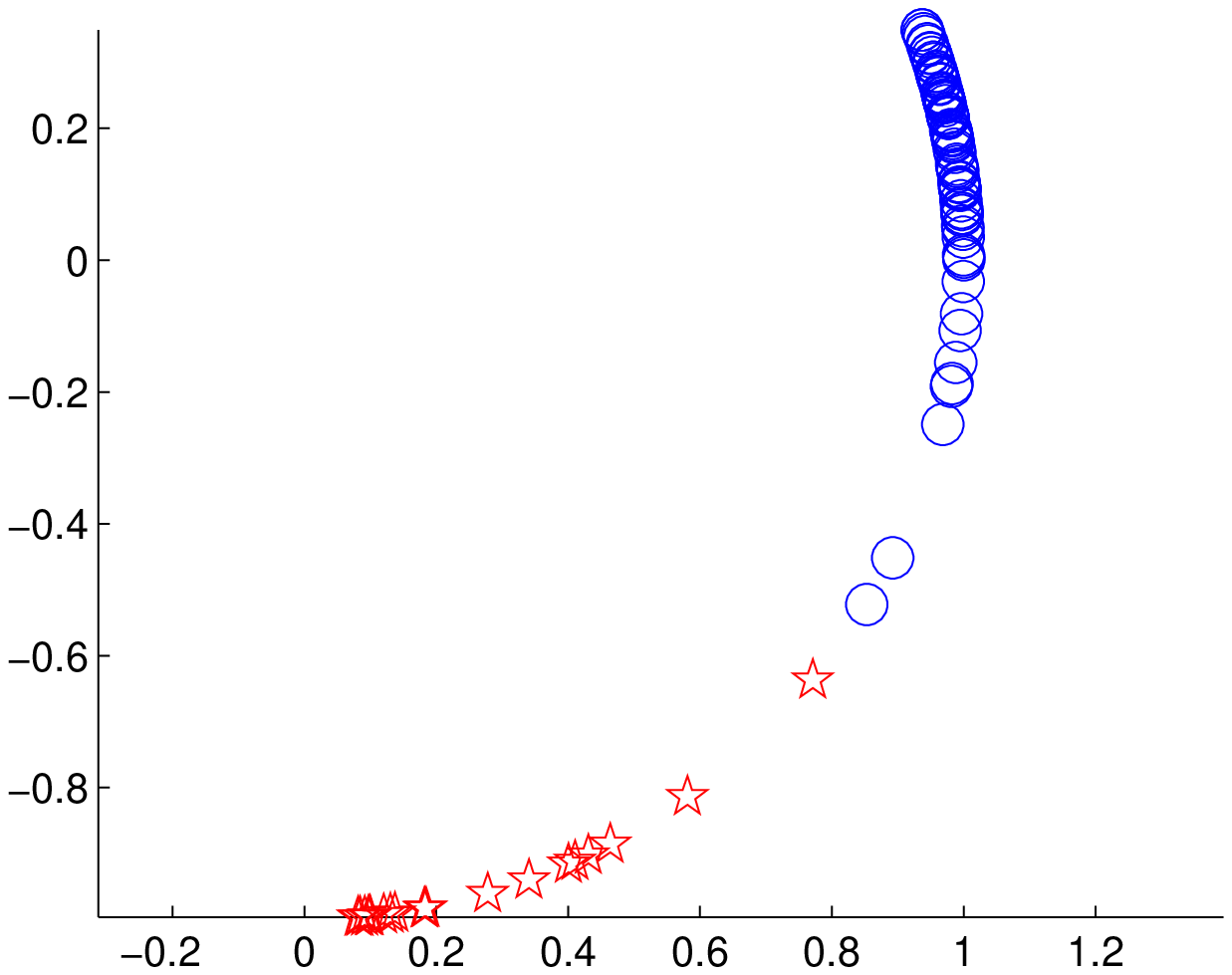}
    }
\caption{The underlying clusters and those found by $K$-means in the
$\mathbf{U}$, $\mathbf{T}$, $\mathbf{V}$ spaces. The given data
consists of 80 and 20 points on two lines in $\mathbb{R}^2$. We note
that, in order for the rows of $\mathbf{U}$ to have similar
magnitudes to those of $\mathbf{T}$ and $\mathbf{V}$, we have scaled
each row of $\mathbf{U}$ with the square root of the average cluster
size, i.e., $\sqrt{N/K}$.}
\end{figure}

Using the cluster sizes, or the row lengths, one could
normalize the matrix $\mathbf{U}$ and obtain two matrices
$\mathbf{T,V}$ whose rows are defined as follows:
\begin{align}
%\mathbf{u}^{(i)}&=\sqrt{\frac{N}{K}} \cdot \mathbf{u}^{(i)}, \quad 1\leq i\leq N; \\
\mathbf{t}^{(i)}&=\sqrt{N_k} \cdot \mathbf{u}^{(i)}, \quad  i\in\mathrm{I}_k, 1\leq k\leq K\\
\mathbf{v}^{(i)}&=\frac{1}{\normV{\mathbf{u}^{(i)}}}\cdot\mathbf{u}^{(i)},
\quad 1\leq i\leq N.
\end{align}
These two normalizations are explained as follows. The
$\mathbf{V}$ normalization discards all the magnitude information of
the rows of $\mathbf{U}$ to contain only the angular information
between them. The $\mathbf{T}$ normalization, containing the same
angular information, reduces to $\mathbf{U}$ when
$N_1=\cdots=N_K=N/K$, and otherwise tries to further separate the
underlying clusters by scaling the rows using the cluster sizes. See
Figure~\ref{fig:UTV_truth} for an illustration of the $\mathbf{U}$,
$\mathbf{T}$, $\mathbf{V}$ spaces.

\begin{remark} \label{rmrk:prac_normalize}
The normalization $\mathbf{T}$ assumes knowledge of the underlying cluster sizes, but can be effectively approximated
without this knowledge when using our practical version of TSCC, i.e., SCC~\cite{spectral_applied}.
The SCC algorithm employs an iterative sampling procedure which converges quickly,
thus it can estimate $\mathbf{T}$ in the current iteration by using the clusters
obtained in the previous iteration. %The
%normalization suggested in Remark~\ref{rmrk:ng_D_normalize} can be
%implemented similarly.
\end{remark}

%\begin{remark}
%\label{rmrk:ng_D_normalize} A third normalization arises when considering the ideal case of Ng et
%al.~\cite{Ng02} (see Remark~\ref{rmrk:ng_ideal}). It uses the degrees of the data points,
%i.e., the diagonal elements of the matrix $\mathbf{D}$ in the following way:
%\[\mathbf{r}^{(i)}:=\sqrt{\frac{\sum_{j\in\mathrm{I}_k} D_{jj}}{D_{ii}}}\cdot \mathbf{u}^{(i)}, \quad i\in\mathrm{I}_k, 1\leq k\leq K.\]
%%The resulting matrix $\mathbf{R}$ can be approximated more tightly by the matrix $\mathbf{V}$.
%\end{remark}

We view the matrix $\mathbf{V}$ as a weak approximation to
$\mathbf{T}$. Indeed, in the ideal case they coincide, since for all
$1 \leq k \leq K$,
$$\normV{\widetilde{\mathbf{u}}^{(i)}} =  \frac{1}{\sqrt{N_k}},\quad i\in\mathrm{I}_k$$
(see equation~\eqref{eq:U_tilde}). In the general case, the above
equality only holds on average. More precisely, the orthonormality
of $\mathbf{U}$ implies that
\[\sum_{k=1}^K \sum_{i\in\mathrm{I}_k} \normV{\mathbf{u}^{(i)}}^2
=\normF{\mathbf{U}}^2 = \sum_{j=1}^K \normV{\mathbf{u}_j}^2 = K\,.\]

We next define two criterions for analyzing the performance of
$\mathbf{U}$, $\mathbf{T}$ and $\mathbf{V}$ when directly applying
$K$-means to them.

First, we define a notion of \emph{the separation factor} for the
centers of the underlying clusters in each of the $\mathbf{U}$,
$\mathbf{T}$ and $\mathbf{V}$ spaces. The separation factor of the
centers in the $\mathbf{U}$ space is defined as follows:
\begin{align}
\beta(\mathbf{U}):=\frac{\sum_{1\leq i<j\leq
K}\langle\mathbf{c}^{(i)},\mathbf{c}^{(j)}\rangle^2}{\left(\sum_{1\leq
k\leq K}\normV{\mathbf{c}^{(k)}}^2\right)^2}.
\end{align}
The separation factors $\beta(\mathbf{T}), \beta(\mathbf{V})$ are
defined similarly. The smaller $\beta$ is, the more separated in
$\mathbb{R}^K$ the centers of the underlying clusters are.
Lemma~\ref{lem:separation_leq_tv} directly implies that
$\beta(\mathbf{T})$ is controlled by $\tv$ as follows.
\begin{lemma}
\[\beta(\mathbf{T}) \leq \frac{\tv}{(K-\tv)^2}.\]
\end{lemma}
We note that $\beta(\mathbf{U})=\beta(\mathbf{T})$ when $N_k = N/K,
k=1,\ldots, K$. In general, we observe that
$\beta(\mathbf{U})\leq\beta(\mathbf{T})\leq \beta(\mathbf{V})$, with
the former two being fairly close. For example,
$\beta(\mathbf{U})=.0004,\beta(\mathbf{T})=.0006,\beta(\mathbf{V})=.0032$
in Figure~\ref{fig:UTV_truth}. In practice, however, we have found
that the underlying clusters in the $\mathbf{U,T,V}$ spaces are usually not
closely concentrated around their centers, thus this criterion may not be
sufficient.

Second, we define a notion of \emph{the clustering identification
error} in the $\mathbf{U}$, $\mathbf{T}$ and $\mathbf{V}$ spaces
respectively. For ease of discussion, we suppose that $K=2$. In the
$\mathbf{U}$ space, the corresponding error has the form:
\begin{align}
e_\mathrm{id}(\mathbf{U}):=\frac{1}{N}\cdot\sum_{k=1}^2
\#\left\{i\in\mathrm{I}_k \mid
\normV{\mathbf{u}^{(i)}-\mathbf{c}^{(k)}}\geq 1/2 \cdot
\normV{\mathbf{c}^{(1)}-\mathbf{c}^{(2)}}\right\}
\end{align}
The errors in the $\mathbf{T,V}$ spaces are defined similarly. The
following lemma (proved in Appendix~\ref{prf:lem_id_error}) shows
that both $e_\mathrm{id}(\mathbf{T})$ and
$e_\mathrm{id}(\mathbf{U})$ can be controlled by $\tv$, with the
former having a smaller upper bound.
\begin{lemma}\label{lem:id_error}
Suppose that $K=2$. If
$$\tv<\left(\sqrt{3}-1\right)^2,$$ then the identification error in the $\mathbf{T}$ space is
bounded above as follows:
\begin{align}
e_\mathrm{id}(\mathbf{T})&\leq \frac{4 \cdot
\tv}{2-\tv-2\sqrt{\tv}}. \end{align}
If
\[\tv<\left(\sqrt{2+\frac{4}{\varepsilon_1^2}}-\frac{2}{\varepsilon_1}\right)^2,\] then
the identification error in the $\mathbf{U}$ space is bounded above
as follows:
\begin{align}
e_\mathrm{id}(\mathbf{U})&\leq
\frac{4\cdot\tv}{2-\tv-4/\varepsilon_1\cdot \sqrt{\tv}},
\end{align}
where the constant $\varepsilon_1$ is defined in
equation~\eqref{eq:comparable_sizes}.
\end{lemma}

We remark that the clustering identification errors
$e_\mathrm{id}(\mathbf{U}),e_\mathrm{id}(\mathbf{T}),e_\mathrm{id}(\mathbf{V})$
have only theoretical meanings. However, they can be used to
estimate the clustering errors of $K$-means when applied in the
$\mathbf{U,T,V}$ spaces respectively. We observed in practice that
$e_\mathrm{id}(\mathbf{T})$ and $e_\mathrm{id}(\mathbf{V})$ are
often very close.
% (see e.g., Figure~\ref{fig:UTV_kmeans}).

Following the above discussion we think that $\mathbf{T}$ %(or possibly the normalization $\mathbf{R}$ suggested in Remark~\ref{rmrk:ng_D_normalize})
is probably
the right normalization to be used in TSCC. Its practical implementation
should follow Remark~\ref{rmrk:prac_normalize}. We note that the
application of this normalization in
Lemma~\ref{lem:separation_leq_tv} results in analogous estimates for
the $\mathbf{T}$ space which are independent of the sizes of
clusters. Indeed, this normalization seems to outperform
$\mathbf{U}$ when $N_1,\ldots, N_K$ vary widely (this claim is
supported in practice by numerical experiments and in theory by
Lemma~\ref{lem:id_error}). Another reason for our preference of
$\mathbf{T}$ is that performing $K$-means in the $\mathbf{T}$ space
is equivalent to performing weighted $K$-means (with weights $N_k/N,
1\leq k\leq K$) in the $\mathbf{U}$ space, which allows small
clusters to have relatively larger variance (see e.g.,
Figure~\ref{fig:UTV_truth}).

The $\mathbf{V}$ normalization is another possibility to use in
TSCC. On one hand, it is a weak approximation to $\mathbf{T}$; on
the other hand, it contains only the angular information of the rows
of $\mathbf{U}$. The use of only angular information for $K$-means
clustering, partly supported by the polarization theorem in
\cite{Brand03unifying}, seems to also separate the underlying
clusters further. However, we need to understand this normalization
more thoroughly, i.e., in terms of theoretical analysis.

In~\cite{spectral_applied} we have used $\mathbf{U}$ to demonstrate
our numerical strategies, which also apply to $\mathbf{T}$ and
$\mathbf{V}$, and obtained good numerical results.

\subsubsection{TSCC Without Normalizing $\mathbf{W}$}
\label{subsec:unnorm} We analyze here the TSCC algorithm when the
matrix $\mathbf{W}$ is not normalized, i.e., skipping
Step~\ref{step:normalize_W} of Algorithm~\ref{alg:TSCC} and letting
$\mathbf{Z}\equiv \mathbf{W}$. We refer to the corresponding variant
of TSCC as TSCC-UN, and formulate below analogous results of
Proposition~\ref{prop:perfect_tensor} and
Theorem~\ref{thm:perturbation_main}. The proof of
Proposition~\ref{prop:perfect_tensor_unnorm} directly follows that
of Proposition~\ref{prop:perfect_tensor} in
Appendix~\ref{prf:prop_perfect_tensor} (in particular,
equations~\eqref{mat:W_tilde} and~\eqref{mat:W_tilde_j}).
Theorem~\ref{thm:perturbation_main_unnorm} is proved in
Appendix~\ref{prf:thm_perturbation_main_unnorm}.
\begin{proposition}
\label{prop:perfect_tensor_unnorm} Suppose that the TSCC-UN
algorithm is applied with the perfect tensor
$\widetilde{\mathbf{A}}$.
%\[N>\sqrt{(d+1)\left(1-\frac{K-1}{K}\varepsilon_1\right)^d\left(\frac{2K}{\varepsilon_1}\right)^{d+2}},\]
%\[\widetilde{d}_1 > (d+1)\cdot \perm{N_K-2}{d},\]
Then
\begin{enumerate}
\item The eigenvalues of $\widetilde{\mathbf{W}}$ are
$\widetilde{d}_K\geq \cdots \geq \widetilde{d}_2\geq
\widetilde{d}_1$ (each of multiplicity 1), and
$\widetilde{\nu}_K\geq \cdots \geq \widetilde{\nu}_2\geq
\widetilde{\nu}_1$ (of multiplicity $N_K,\ldots, N_2,N_1$
respectively), where
\begin{align}
\widetilde{d}_k   &:= (N_k-d-1)\cdot\perm{N_k-1}{d+1}, \label{eq:d_j} \\
\widetilde{\nu}_k &:= (d+1)\cdot\perm{N_k-2}{d}.
\end{align}
\item If $\widetilde{d}_1>\widetilde{\nu}_K$, the rows of $\widetilde{\mathbf{U}}$ are exactly $K$ mutually orthogonal
vectors, each representing a distinct underlying cluster.
\end{enumerate}
\end{proposition}
\begin{theorem} Suppose that TSCC-UN is applied with a general affinity tensor
$\mathcal{A}$, and that
\begin{equation}\label{ineq:N_unnorm}
N\geq
%
%\max\left(\frac{2(d+1)K}{\varepsilon_1},
               \sqrt{2(d+1)\left(1-\frac{K-1}{K}\varepsilon_1\right)^d\left(\frac{2K}{\varepsilon_1}\right)^{d+2}},
               %\right),
\end{equation}
Let
\[C_2(K,d,\varepsilon_1,\varepsilon_2):=32\left(\frac{2K}{\varepsilon_1}\right)^{2(d+2)}.\]
If \[N^{-(d+2)}\normF{\mathcal{E}}^2 \leq \frac{1}{8C_2},\] then
\begin{equation}
\tv \leq C_2 \cdot N^{-(d+2)}\normF{\mathcal{E}}^2.
\end{equation}
\label{thm:perturbation_main_unnorm}
\end{theorem}

In view of equation~\eqref{ineq:N_unnorm}, the TSCC-UN algorithm
seems to require large data size in order to work well. Numerical
experiments also indicate that this approach is very sensitive to
the variation of cluster sizes, and works consistently worse than
the normalized approach, i.e., TSCC. Our current analysis, however, does not
manifest the significant advantage of the normalized approach. We
thus leave the related exploration to later research.

Von Luxburg et al.~\cite{vonLuxburg08} have shown that in the
framework of kernel spectral clustering, the normalized method is
consistent under very general conditions. On the other hand, the
unnormalized method is only consistent under very specific
conditions that are rarely met in practice. Since $\mathbf{W}$ can
be seen as a kernel matrix, \cite{vonLuxburg08} provides another
evidence for our preference of the normalized approach.

\section{Probabilistic Analysis of TSCC}
\label{sec:prob_analysis} In this section we analyze the performance
of the TSCC algorithm with its own affinity tensor, i.e., the polar
tensor of equation~\eqref{eq:affinity_tensor}. We control with high
probability (with respect to the sampling in Problem~\ref{prob:hlm})
the goodness of clustering of TSCC when applied to the data
generated in Problem~\ref{prob:hlm}.
%
%Our goal in this section is to estimate this quantity for special
%types of data sets representing the hybrid linear modeling problem
%and also using properties of our particular polar affinities of
%equation~\eqref{eq:affinity_tensor}. Consequently, we obtain a
%probabilistic statement about the performance of the algorithm in
%this setting.

%In Subsection~\ref{subsec:5.1} we introduce necessary notation and
%the basic context of this section. We then present in
%Subsection~\ref{subsec:5.3} the main result of this paper, while
%interpret it in Subsection...

\subsection{Basic Setting and Definitions}
\label{subsec:prob_analysis_setting} We follow the setting of hybrid
linear modeling described in Problem~\ref{prob:hlm} together with
the assumptions of regularity and possibly $d$-separation of $\{
\mu_i \}_{i=1}^K$ (see Remark~\ref{rem:d-sep}) as well as the
restriction imposed by equation~\eqref{eq:comparable_sizes}. We
denote the corresponding $N$ random variables by
$\mathfrak{X}_1,\ldots,\mathfrak{X}_N\in \mathbb{R}^D$ and maintain
the previous notation for their sampled values
$\mathbf{x}_1,\dots,\mathbf{x}_N$. The joint sample space is
$(\reals^D)^N$, and the corresponding joint probability measure is
\begin{equation}\label{eq:mu_p}
\mu_\mathrm{p}:=\mu_1^{N_1}\times\cdots\times\mu_K^{N_K}.
\end{equation}
%We will
%Theorem~\ref{thm:ls_by_curvature} states that the flatness of these
%measures can be quantified by their polar curvatures
%$c^2_\mathrm{p}(\mu_k; \lambda, B_k), k=1,\ldots,K$.
%In the extreme
%case where $c^2_\mathrm{p}(\mu_k; \lambda, B_k)=0$, the support of
%$\mu_k$ will be a $d$-dimensional ball contained in some $d$-flat.

%We will initially assume an arbitrary nonnegative curvature
%function, $c$, defined for any $d+2$ vectors in the given data, and
%later restrict it to be a LS curvature.

We introduce an incidence constant reflecting the separation between
the measures $\mu_1,\ldots,\mu_K$ in regard to the polar curvature
$c_\mathrm{p}$ and the tuning parameter $\sigma$. We first define
the following sets
\begin{equation*}
%\mathrm{S} = \bigcup^K_{k=1} \mathrm{S}_k, \qquad \text{where}\quad
\mathrm{S}_k := \left(\supp(\mu_k)\right)^{d+2},\ 1\leq k \leq K.
\end{equation*}
%and a measure
%%
%\be \label{eq:def_mu_s} \mu_\mathrm{s} =
%\frac{1}{K}\cdot\sum^K_{k=1} \mu_k. \ee
%%
Then, given a constant $\sigma>0$, the incidence constant has the
form:
\begin{align} \label{def_incidence}
&C_{\mathrm{in}}(\mu_1,\ldots,\mu_K;\sigma) := \nonumber \\
&\quad \max_{\latop{1\leq k_1,\ldots,k_{d+2}\leq K}{\text{not all
equal}}} \int_{\mathrm{S}_{k_1}}\cdots\int_{\mathrm{S}_{k_{d+2}}}
e^{\frac{-c_\mathrm{p}(\mathbf{z}_1,\ldots,\mathbf{z}_{d+2})}{\sigma}}
\di\mu_{k_1}(\mathbf{z}_1)\ldots \di\mu_{k_{d+2}}(\mathbf{z}_{d+2}),
\end{align}
where the maximum is taken over all $1\leq k_1,\ldots,k_{d+2}\leq K$
except $k_1=k_2=\cdots=k_{d+2}$.

\begin{remark}
For TLSCC, the incidence constant is defined as follows:
\begin{align} \label{def_incidence_linear}
&C_{\mathrm{in,L}}(\mu_1,\ldots,\mu_K;\sigma) := \nonumber \\
& \,\max_{\latop{1\leq k_1,\ldots,k_{d+1}\leq K}{\text{not all
equal}}} \int_{\mathrm{S}_{k_1}}\cdots\int_{\mathrm{S}_{k_{d+1}}}
e^{\frac{-c_\mathrm{p}(\mathbf{0},\mathbf{z}_1,\ldots,\mathbf{z}_{d+1})}{\sigma}}
\di\mu_{k_1}(\mathbf{z}_1)\ldots \di\mu_{k_{d+1}}(\mathbf{z}_{d+1}).
\end{align}
\end{remark}

We note that for both TSCC and TLSCC, the incidence constant is
between 0 and 1. The smaller the incidence constant is, the more
separated (in terms of the polar curvature and the tuning parameter) the measures are.
%In general we expect that
%the following equation holds for our underlying measures
%$\mu_1, \ldots, \mu_K$:
%\begin{equation}
%\lim_{\sigma \to 0+} C_{\mathrm{in}}(\mu_1,\ldots,\mu_K;\sigma) = 0.
%\label{eq:incd}
%\end{equation}
%That is, as $\sigma$ approaches zero, the separation between the
%measures improves.
In Subsection~\ref{subsec:interpreting_alpha} we estimate the
incidence constant in a few special instances of hybrid
linear modeling.% when applying TLSCC. %and using the polar curvature,
%and show that the above equation holds in each instance.

\subsection{The Main Result}
\label{subsec:prob_analysis_result} The following theorem (proved in
Appendix~\ref{prf:thm_good_with_high_prob}) shows that, when the
underlying measures are sufficiently flat and well separated from
each other, with high probability (with respect to the sampling of
Problem~\ref{prob:hlm}) the TSCC algorithm segments the $K$
underlying clusters well.
\begin{theorem}
\label{thm:good_with_high_prob} Suppose that the TSCC algorithm is
applied to the data generated in Problem~\ref{prob:hlm} with a
tuning parameter $\sigma> 0$. Let
\be \label{eq:def_alpha} \alpha := \frac{1}{\sigma^2}\sum_{k=1}^K
c^2_\mathrm{p}(\mu_k)
 +
C_{\mathrm{in}}(\mu_1,\ldots,\mu_K;\sigma/2), \ee
and $C_1=C_1(K,d,\varepsilon_1,\varepsilon_2)$ be the constant
defined in Theorem~\ref{thm:perturbation_main}.
If \[\alpha < \frac{1}{16C_1},\] then
%Then \[E_{\mu_\mathrm{p}}(\tv | \text{\rm \
%Assumption~\ref{assmp:large_D_ii} holds}) \leq \alpha \cdot
%C(K,d,\varepsilon_1,\varepsilon_2),\] and moreover,
\begin{align}
&\mu_\mathrm{p}\left(\tv \leq 2\alpha \cdot C_1 \mid
\mathrm{Assumption~\ref{assmp:large_D_ii}\ holds} \right) \geq
%1-\exp\left(-\frac{2N\alpha^2}{(d+2)^2}\right).
1-e^{-2N\alpha^2/(d+2)^2}.
\end{align}
%
%In particular, if $\sigma$ can be chosen so that $\alpha$ is
%sufficiently small, then with high probability, the TSCC algorithm
%separates the different clusters well.
\end{theorem}
%
%\begin{remark}
%\label{remark:sample} The theory described above extends to the case
%of sampling the data points independently and identically from a
%probability measure obtained as a mixture of $\mu_1, \ldots, \mu_K$,
%that is, a measure of the form $\sum_{k=1}^K \gamma_k \, \mu_k$,
%where $\gamma_k$, $k=1,\ldots, K$, are positive constants summing to
%$1$. The precise result in this case is formulated in Appendix.
%\end{remark}
%
\begin{remark}
Theorem~\ref{thm:good_with_high_prob} also holds for the TLSCC
algorithm, but with $d$ replaced by $d-1$, and the constant $\alpha$
by \be \label{eq:def_alpha_L} \alpha_\mathrm{L} :=
\frac{1}{\sigma^2}\sum_{k=1}^K
c^2_\mathrm{p,L}(\mu_k)
 +
C_{\mathrm{in,L}}(\mu_1,\ldots,\mu_K;\sigma/2), \ee where for any
Borel probability measure $\mu$,
\[c_\mathrm{p,L}(\mu):=\sqrt{\int c^2_\mathrm{p}(\mathbf{0},\mathbf{z}_1,\ldots,\mathbf{z}_{d+1})\di\mu(\mathbf{z}_1)\ldots\di\mu(\mathbf{z}_{d+1})}.\]
\end{remark}
\begin{remark}
A similar version of Theorem~\ref{thm:good_with_high_prob} holds for
general affinity tensors of the form
$\{e^{-c(\mathbf{x}_{i_1},\ldots,\mathbf{x}_{i_{d+2}})/\sigma}\}_{1\leq
i_1,\ldots,i_{d+2}\leq N}$, where $c$ is a nonnegative, symmetric
function defined on $\mathbb{R}^{d+2}$. The significance of using
the polar curvature, or any other curvature satisfying
Theorem~\ref{thm:comparable_cp_lsq}, is explained in
Subsection~\ref{subsec:interpreting_alpha}.
\end{remark}

We showed in Lemma~\ref{lem:id_error} that the clustering
identification errors $e_\mathrm{id}(\mathbf{U})$ and
$e_\mathrm{id}(\mathbf{T})$ can be controlled by $\tv$ when $K=2$.
Combining Lemma~\ref{lem:id_error} and
Theorem~\ref{thm:good_with_high_prob} yields the following
probabilistic statement.
\begin{corollary}\label{lem:id_error_alpha}
Suppose that $K=2$, and that $\alpha, C_1$ are the constants defined
in Theorem~\ref{thm:good_with_high_prob}. If \[\alpha <
\frac{1}{16C_1},\] then
\begin{align*}
&\mu_\mathrm{p}\left(e_\mathrm{id}(\mathbf{T}) \leq \frac{4\,
\alpha\, C_1}{1-\alpha\, C_1-\sqrt{2\,\alpha\, C_1}} \mid
\mathrm{Assumption~\ref{assmp:large_D_ii}\ holds} \right) \\& \qquad
\qquad \geq
%1-\exp\left(-\frac{2N\alpha^2}{(d+2)^2}\right).
1-e^{-2N\alpha^2/(d+2)^2}.
\end{align*}
If
\[\alpha< \frac{1}{2C_1}\cdot \min\left(\frac{1}{8},\left(\sqrt{2+\frac{4}{\varepsilon_1^2}}-\frac{2}{\varepsilon_1}\right)^2\right),\] then
\begin{align*}
&\mu_\mathrm{p}\left(e_\mathrm{id}(\mathbf{U}) \leq \frac{4\,
\alpha\, C_1}{1-\alpha\, C_1-2/\varepsilon_1\cdot\sqrt{2\,\alpha\,
C_1}} \mid \mathrm{Assumption~\ref{assmp:large_D_ii}\ holds} \right)
\\ &\qquad \qquad \geq
%1-\exp\left(-\frac{2N\alpha^2}{(d+2)^2}\right).
1-e^{-2N\alpha^2/(d+2)^2}.
\end{align*}
\end{corollary}

\subsection{Interpretation of the Constant $\alpha$}
\label{subsec:interpreting_alpha}
Theorem~\ref{thm:good_with_high_prob} shows the strong effect of the
constant $\alpha$ on the goodness of clustering of the TSCC
algorithm. This constant has two parts, which are explained
respectively as follows.

Theorem~\ref{thm:comparable_cp_lsq} implies that the first part of $\alpha$ is comparable to %~\cite{LW-volume}
$$
\frac{1}{\sigma^2} \cdot  \sum_{k=1}^K e^2_2(\mu_k).
$$
We thus view the first part as the sum of the within-cluster errors
of the model scaled by $\sigma^2$.

\begin{remark}\label{rem:interpret_alpha_Lq}
A similar interpretation applies to the tensors defined in
equation~\eqref{eq:affinity_tensor_Lq}. In this case, for any $q
\geq 1$, the first term of $\alpha$ is replaced by
\[\frac{1}{\sigma^2}\sum_{k=1}^K c_\textrm{p}^{(2q)}(\mu_k),\]
where for any Borel probability measure $\mu$,
\[c^{(2q)}_\mathrm{p}(\mu):=\int c^{2q}_\mathrm{p}(\mathbf{z}_1,\ldots,\mathbf{z}_{d+2})
\,\di\mu(\mathbf{z}_1)\ldots\di\mu(\mathbf{z}_{d+2}).\] The above
sum is then comparable to
\[\frac{1}{\sigma^2}\cdot\sum_{k=1}^K e_{2q}^{2q}(\mu_k),\]
where $e_{2q}(\mu_k)$ is the error of approximating $\mu_k$ by a
$d$-flat while minimizing the $\textrm{L}_{2q}$ norm~\cite{LW-volume}.
\end{remark}

We interpret the second part of $\alpha$, i.e., the incidence
constant, as the between-clusters interaction of the model. Unlike the
first part, we do not have a theoretical result that fully
establishes this interpretation. We show in a few special cases
(with underlying linear subspaces) how to control this constant.

In the first example (Example~\ref{ex:lines_tscc}) we estimate the
incidence constant for two orthogonal line segments when using TSCC.
The next three examples assume the use of the TLSCC algorithm. In
Example~\ref{ex:lines} the model includes distributions along two
clean line segments with an arbitrary angle $\theta$ between them.
We establish the dependence of the incidence constant on $\theta$
and $\sigma$. In Example~\ref{ex:rectangles} we consider two
orthogonal lines with uniform noise around them, and demonstrate the
dependence of the incidence constant on the level of the noise and
$\sigma$. Example~\ref{ex:disks} considers two clean orthogonal
planes in $\mathbb{R}^3$.

%We remark that the combination of the above two terms, having
%opposite dependence on $\sigma$, imply lower and upper bounds on
%$\sigma$ for obtaining near optimal values of $\alpha$. It is only
%in the case of clean $d$-flats where there is no lower bound on
%$\sigma$, i.e., the performance is optimal when $\sigma$ approaches
%zero.

\begin{example}\label{ex:lines_tscc}
(\textbf{TSCC: two orthogonal clean lines}). We consider the
following two orthogonal line segments in $\mathbb{R}^2$:
\[\text{L1}:\ y=0,\quad 0 \leq x\leq L,\] and
\[\text{L2}:\ x=0,\quad 0 \leq y\leq L,\]
in which $L>0$ is a fixed constant. We assume arclength measures
$\mu_1=\frac{\di x}{L}, \mu_2=\frac{\di y}{L}$ supported on L1 and
L2 respectively. For any $\sigma>0$, the incidence constant for TSCC
is bounded as follows (see Appendix~\ref{prf:ex_lines_tscc}):
\begin{equation}
\label{incd:lines_tscc} C_{\mathrm{in}}(\mu_1,\mu_2;\sigma)\leq
\frac{\sigma}{\sqrt{2}L}\left(1-e^{-\sqrt{2}L/\sigma}\right).
\end{equation}
%In particular, $C_{\mathrm{in}}(\mu_1,\mu_2;\sigma)\to 0$ as
%$\sigma\to 0+$.
\end{example}

\begin{example}\label{ex:lines}
(\textbf{TLSCC: two intersecting clean lines}). We consider the
following two lines in $\mathbb{R}^2$:
\[\text{L1}:\ y=0,\quad 0 \leq x\leq L,\] and
\[\text{L2}:\ y=r\sin\theta,\, x=r\cos\theta, \quad 0\leq r\leq L,\]
in which $L>0$ and $0<\theta \leq \pi/2$ are fixed constants. We
assume arclength measures $\mu_1=\frac{\di x}{L}, \mu_2=\frac{\di
r}{L}$ supported on L1 and L2 respectively. For any $\sigma>0$, the
incidence constant for TLSCC is bounded as follows (see
Appendix~\ref{prf:ex_lines}):
\begin{equation}
\label{incd:lines} C_{\mathrm{in,L}}(\mu_1,\mu_2;\sigma)\leq
2\left(\frac{\sigma}{L\sin\theta}\right)^2 \cdot
\left(1-e^{-\frac{L\sin\theta}{\sigma}}\left(1+\frac{L\sin\theta}{\sigma}\right)\right).
\end{equation}
%In particular, $C_{\mathrm{in,L}}(\mu_1,\mu_2;\sigma)\to 0$ as
%$\sigma\to 0+$.
We note that when $\theta=\pi/2$, $C_\textrm{in,L}$
has a faster decay rate than $C_\textrm{in}$ (see
Example~\ref{ex:lines_tscc}).
\end{example}
\begin{example}\label{ex:rectangles}
(\textbf{TLSCC: two orthogonal rectangles}). We consider two
rectangular strips in $\mathbb{R}^2$ determined by the following
vertices respectively:
\[\text{R1}:\ (\epsilon,0),(L+\epsilon,0),(\epsilon,\epsilon),(L+\epsilon,\epsilon),\] and
\[\text{R2}:\ (0,\epsilon),(0,L+\epsilon),(\epsilon,\epsilon),(\epsilon,L+\epsilon),\]
in which $0 < \epsilon \ll L$. We assume uniform measures
$\mu_i=\frac{1}{L\epsilon}\mathcal{L}_2$ restricted to R$i$,
$i=1,2$. We view R1 and R2 as two lines surrounded by uniform noise.
Let $\omega :=L/\epsilon$. For any $\sigma>0$, the incidence
constant for TLSCC has the following upper bound (see
Appendix~\ref{prf:ex_rectangles})
\begin{equation}
\label{incd:rectangles} C_{\mathrm{in,L}}(\mu_1,\mu_2;\sigma) \leq
\frac{\sqrt{\sigma}}{\omega^2} +
\frac{2\sqrt[4]{\sigma}}{\omega}\cdot
e^{-1/\left(2\sigma^{3/4}\right)} + e^{-1/\sigma^{3/4}}.
\end{equation}
%The incidence constant goes to zero as $\sigma\to 0+$.
In the limiting case of $\epsilon \to 0+$, i.e., when having two orthogonal lines with
practically no noise, the above estimate decays faster to zero than the one in Example~\ref{ex:lines} with
$\theta = \pi/2$. This is due to the fact that in the current
example we exclude the intersection of the two lines for any
$\epsilon>0$. As it turned out, the limit of the corresponding integral (as
$\epsilon \to 0+$) is not the same as the full integral of this
limit.
\end{example}
\begin{example}\label{ex:disks}
(\textbf{TLSCC: two perpendicular clean half-disks}). We consider
the following portions of two unit disks (in polar coordinates) in
$\mathbb{R}^3$:
\[\text{D1}:\ x=0, \,y=\rho \cos\varphi,\, z =\rho\sin\varphi, \quad 0\leq \rho\leq 1, 0\leq \varphi\leq \pi,\] and
\[\text{D2}:\ x=r \cos\theta,\, y =r\sin\theta,\, z=0, \quad 0\leq r\leq 1, -\pi/2\leq \theta\leq \pi/2.\]
We also assume uniform measures $\mu_i=\frac{2}{\pi}\mathcal{L}_2$
restricted on D$i$, $i=1,2$. In this case, the incidence constant
for TLSCC is bounded above by the following quantity (see
Appendix~\ref{prf:ex_disks})
\begin{equation}
\label{incd:disks} C_{\mathrm{in,L}}(\mu_1,\mu_2;\sigma) \leq
\frac{8\sqrt{\sigma}}{\pi^2} + \frac{8\sqrt[4]{\sigma}}{\pi} +
\frac{4\sigma^2}{(\sin\sqrt[4]\sigma)^4}.
\end{equation}
%Similarly, it goes to zero as $\sigma\to 0+$.
\end{example}

\subsection{On the Existence of Assumption~\ref{assmp:large_D_ii}}
\label{subsubsec:existence_assmp1} The theory developed in this
paper assumes that all affinity tensors used with TSCC, in
particular the polar tensor, satisfy
Assumption~\ref{assmp:large_D_ii}. We present some partial results
regarding the existence of this assumption for the polar tensor
while taking into account the restrictions on the size of $\sigma$
imposed by Theorem~\ref{thm:good_with_high_prob}. We remark that
those results also extend to some other tensors.

We first show in the following lemma (proved in Appendix~\ref{prf:lem_large_degrees1}) that if a data is sampled from a hybrid linear
model without noise, then Assumption~\ref{assmp:large_D_ii} is
always satisfied with the constant $\varepsilon_2 = 1$.
\begin{lemma}\label{lem:large_degrees1} If the
TSCC is applied to data sampled from a mixture of clean $d$-flats,
then
\[\mathbf{D}\geq \widetilde{\mathbf{D}}.\]
\end{lemma}

For more general data sampled from a hybrid linear model with
respect to a ball $B$ (according to Problem~\ref{prob:hlm}), one can
easily obtain that Assumption~\ref{assmp:large_D_ii} is satisfied
with
\begin{equation} \label{eq:naive_eps_2}
\varepsilon_2 = e^{-\frac{2}{\sigma}\cdot \sqrt{d+2}\cdot \diam(B)}
\end{equation}
(see proof in Appendix~\ref{prf:lem_large_degrees2}).
However, since our main
estimates depend inversely on $\varepsilon_2$ we would need to have
the constant $\varepsilon_2$ sufficiently close to $1$, and thus the
above equation implies a lower bound on $\sigma$ of the order of
$\diam(B)$. On the other hand,  the first term of the constant
$\alpha$ stated in Theorem~\ref{thm:good_with_high_prob} implies an
upper bound for $\sigma$ of the order of $\sum_{k=1}^K
c^2_\mathrm{p}(\mu_k)$. These two bounds are rather contradictory
(it is easy to see this in view of the interpretation of the sum $\sum_{k=1}^K
c^2_\mathrm{p}(\mu_k)$ in Subsection~\ref{subsec:interpreting_alpha}).

To resolve the above issue we can replace
$\diam(B)$ in equation~\eqref{eq:naive_eps_2} with the term
$\sum_{k=1}^K c^2_\mathrm{p}(\mu_k)$ and obtain the following estimate in expectation (see proof in Appendix~\ref{prf:lem_large_degrees3}).
\begin{lemma}\label{lem:large_degrees3} If the
TSCC is applied to data sampled according to Problem~\ref{prob:hlm},
then Assumption~\ref{assmp:large_D_ii} holds in expectation in the following sense:
\[
E_{\mu_\mathrm{p}}(\mathbf{D})\geq \varepsilon_2 \cdot
\widetilde{\mathbf{D}},
\]
where \[\varepsilon_2 = e^{-\frac{2}{\sigma}\cdot \max_{1\leq k\leq
K} c_{\mathrm{p}}(\mu_k)}.\]
\end{lemma}

\begin{remark}\label{rmrk:ofer}
We do not expect Assumption~\ref{assmp:large_D_ii} to hold with high
probability (i.e., having the measure $\mu_\mathrm{p}$ close to one)
while maintaining the constant $\varepsilon_2$ formulated in
Lemma~\ref{lem:large_degrees3}. However, it seems reasonable to have
a statement in high probability when replacing the polar curvature $c_\textrm{p}(\mu_k)$
used in defining this constant with their following upper bounds:
$$
\widehat{c}^{\, 2}_{\mathrm{p}}(\mu_k) = \max_{\mathbf{z}_1 \in
\supp(\mu_k)} \int {c}_\mathrm{p}^2(\mathbf{z_1},
\mathbf{z}_2,\ldots,\mathbf{z}_{d+2}) \,\di \mu_k(\mathbf{z}_2) \ldots
\di \mu_k(\mathbf{z}_{d+2})\,.
$$
We leave the investigation of such a statement and the effect of
using $\widehat{c}^{\, 2}_{\mathrm{p}}(\mu)$ instead of ${c}^{\,
2}_{\mathrm{p}}(\mu)$ to future research.
\end{remark}

\section{Conclusion and Future Work}\label{sec:conclusions}
We have analyzed the performance of TSCC in the setting of hybrid
linear modeling. We first showed that we could precisely cluster the
underlying components knowing the perfect tensor, and then
established good performance in the case of reasonable deviation
from the perfect case. Using this result, we proved that if a data
set is sampled independently and identically according to the
setting of Problem~\ref{prob:hlm}, then with high sampling probability the
TSCC algorithm will perform well as long as the underlying
distributions are sufficiently flat and separated.
In~\cite{spectral_applied} we develop a practical version of the
TSCC algorithm by incorporating different numerical techniques and
exemplify its successful performance using a number of artificial
data sets and several real-world applications.

We conclude this paper by discussing both the open directions and
the possible extensions of this work.

\vskip .3cm \noindent {\bf Further understanding of the two
normalizations discussed in
Subsection~\ref{subsubsec:U_normalization}:} We first explored in
Subsection~\ref{subsubsec:U_normalization} possible normalizations
of the matrix $\mathbf{U}$, and analyzed (to some extent) the
performance of TSCC with and without them. We concluded that the
normalization suggested by the matrix $\mathbf{T}$ %(or possibly $\mathbf{R}$ in Remark~\ref{rmrk:ng_D_normalize})
is probably the
right one to apply in TSCC. It will be interesting to test our
practical strategy when applying such a normalization (see
Remark~\ref{rmrk:prac_normalize}) on both artificial and practical
data sets with varying numbers of points within each cluster. Also, we wish to study more carefully the possible
advantages of the normalization suggested by the matrix
$\mathbf{V}$.

At last, Subsection~\ref{subsubsec:U_normalization} analyzed the
TSCC algorithm when applied without the unnormalized matrix
$\mathbf{Z}$. The perturbation results there were practically
comparable to those obtained when applying TSCC with the normalized
matrix $\mathbf{Z}$. It thus did not reveal the significant advantage of using $\mathbf{Z}$. In future investigations
we would like to improve the current estimates so that they
emphasize this significant advantage.

%\subsection*{Refinement of the Current Theory} The following
%directions of extending the theory discussed here are interesting
%for us.
%
\vskip .3cm \noindent {\bf Further interpretation of the incidence
constant:} Currently we have described the behavior of the incidence
constant in a few typical examples of two intersecting linear
subspaces. We ask about characterization of this constant for
general mixtures of flats, and its dependence on the separation
between the subspaces, the magnitude of noise as well as the tuning
parameter.
\vskip .3cm \noindent {\bf Estimation of the clustering
identification error:} We showed in
Subsection~\ref{subsubsec:U_normalization} that when $K =2$ and
$\tv$ is sufficiently small, then a large percentage of the points
can be clustered correctly. We would like to extend the
corresponding analysis to the case where $K>2$.

\vskip .3cm \noindent {\bf Further investigation of Assumption~\ref{assmp:large_D_ii}:}
Assumption~\ref{assmp:large_D_ii} is a crucial condition for Algorithm~\ref{alg:TSCC} to work well.
Our partial results (i.e., Lemmas~\ref{lem:large_degrees1} and \ref{lem:large_degrees3}) showed that
this assumption holds at least in expectation.
%We would like to develop a similar statement which holds with high probability, possibly in the direction suggested in Remark~\ref{rmrk:ofer}.
We would like to explore the existence in high probability of Assumption~\ref{assmp:large_D_ii} with a constant $\varepsilon_2>0$
that does not contradict the bounds imposed by Theorem~\ref{thm:good_with_high_prob} (see discussion in Section~\ref{subsubsec:existence_assmp1}, in particular, Remark~\ref{rmrk:ofer}).

%\vskip .3cm \noindent {\bf Analysis of the classification error:}
%Our notion of goodness of clustering depends on the underlying
%model, in particular, the amount of noise and geometric separation.
%It will be interesting to also analyze the true classification error
%(based on percentages of truly classified points). In particular, we
%are curious about a proof showing that perfect clustering (in terms
%of the latter measure) is obtained when the number of sampled points
%(from the hybrid linear model) goes to infinity. We remark that
%since in practice only small samples of the whole data can be used
%to implement the TSCC, the goodness of clustering measure described
%here is still satisfying for us for the quality of clustering.
%%
%\vskip .3cm \noindent {\bf Phase transition phenomenon}
%Theorem~\ref{thm:good_with_high_prob1} show that if $\alpha$ is
%sufficiently small, then TSCC clusters the data well. It is
%reasonable to expect that if $\alpha$ is sufficiently large (e.g.,
%due to large within-clusters error) then there will be too large
%overlap between the clusters. We inquire about further
%quantification of this idea and careful analysis of the transition
%between the two phases.
%%
%\vskip .3cm \noindent {\bf Asymptotic consistency} To all this
%regards, something that is still missing is a theoretical study of
%which conditions are sufficient to imply some form of asymptotic
%consistency.
%
\vskip .3cm \noindent {\bf Analysis of other frameworks for
multi-way clustering:} Agarwal et al.~\cite{Agarwal05} and Shashua
et al.~\cite{Shashua06} suggested different frameworks for multi-way
spectral clustering. It will be interesting to analyze the
performance of their algorithms when applied to data sampled from a hybrid
linear model.

%
%\vskip .3cm \noindent {\bf Making the algorithm practical:}
%In~\cite{spectral_applied} we incorporate various techniques of
%modern numerical analysis and statistics in order to significantly
%speed up the TSCC algorithm and reduce its storage. The theory
%developed here guides us in such choices. In particular, we randomly
%sample a subset of the columns of the matrix $\mathbf{A}$ in an
%iterative way to produce $\mathbf{W}$, and suggest an automatic way
%of calibrating the tuning parameter $\sigma$. The resulting
%algorithm is probabilistic (Monte Carlo type) and linear in the size
%of the data. We also discuss in~\cite{spectral_applied} the
%application of the algorithm to mixed dimensions.
%Extensive numerical experiments and
%comparisons with other strategies have shown very good performance
%of our algorithm.
%
%\vskip .3cm \noindent {\bf Applications:} We are currently testing
%the algorithm on various real data sets, following examples studied
%by Ma et al.~\cite{Ma07}. We will report our results
%in~\cite{spectral_applied} as well as subsequent work.
%%
\vskip .3cm \noindent {\bf Clustering flats in non-flat spaces, and
even more general shapes:} We are interested in generalizing the
problem of clustering $d$-flats in Euclidean spaces to more general
metric spaces where $d$-flats are replaced by $d$-dimensional
geodesic surfaces. Also, we would like to modify our curvatures to
cluster other shapes, e.g., circles, parabolas, spheres.
\vskip .3cm \noindent {\bf Detecting $d$-flats:} We believe that it
is possible to modify the methods described in this paper to detect
an unknown $d$-flat in uniformly distributed background noise. If we
are able to develop good curvatures for other geometric shapes, then
we can generalize the detection problem to including such shapes
(see~\cite{Arias-Castro05geometric} and references therein for other solutions to this problem).

\section*{Acknowledgement}
We thank the reviewers and the action editor for valuable comments
that have helped improve this paper; Ery Arias-Castro and Tyler
Whitehouse for commenting on an earlier version of this manuscript;
Ofer Zeitouni for clarifying to us Remark~\ref{rmrk:ofer}; Stefan
Atev, Effrosyni Kokiopoulou, Yi Ma, Guillermo Sapiro, Arthur Szlam
and Rene Vidal for valuable references; Dennis Cook, Peter Olver and
Fadil Santosa for serving on the oral preliminary exam committee of
GC, where the initial version of this work was discussed. We thank
Mark Green, Kevin Vixie and IPAM for inviting us to participate in
the 2005 Graduate Summer School on intelligent extraction of
information from graphs and high dimensional data. GL also thanks
Emmanuel Candes, Mark Green and IPAM for inviting him to participate
in parts of the 2004 fall program on multiscale geometry and
analysis in high dimensions. Both programs had a strong effect on
this research. GL thanks Ingrid Daubechies who encouraged him to
think about ``hybrid linear modeling'' already in 2002 (motivated by
a brain imaging problem). The research presented in this paper is
supported by NSF grant \#0612608.

\appendix

\section{Proofs}
\subsection{Proof of Proposition~\ref{prop:perfect_tensor}}
\label{prf:prop_perfect_tensor}
The affinity matrix $\widetilde{\mathbf{A}}$, the matricized version
of $\widetilde{\mathcal{A}}$, is a 0/1 matrix of size $N\times
N^{d+1}$. We identify the unit entries in each row as follows. For any fixed
$1\leq i\leq N_1$, the entries of the $i^{\text{th}}$ row of
$\widetilde{\mathbf{A}}$ are of the form
$\widetilde{\mathcal{A}}(i,i_2,\ldots,i_{d+2}),1\leq
i_2,\ldots,i_{d+2} \leq N$. These entries will be 1 if they
represent affinities of distinct $d+2$ points in
$\widetilde{\mathrm{C}}_1$, that is, the indices
$i,i_2,\ldots,i_{d+2}$ are distinct and between 1 and $N_1$.
Therefore, the $i^\text{th}$ row has exactly ${\perm{N_1-1}{d+1}}$
entries filled by a 1, which is exactly the number of permutations
of size $d+1$ out of the first $N_1$ points excluding $i$.
%(the corresponding notation was specified in
%Section~\ref{sec:notation}).
Similarly, each of the subsequent $N_2$ rows has
${{\perm{N_2-1}{d+1}}}$ ones, and each of the next $N_3$ rows has
${\perm{N_3-1}{d+1}}$ ones, etc..

The weight matrix
$\widetilde{\mathbf{W}}=\widetilde{\mathbf{A}}\widetilde{\mathbf{A}}'$
can be expressed in terms of the tensor $\widetilde{\mathcal{A}}$ in
the following way:
\begin{equation}
\widetilde{{W}}_{ij} = \sum_{1\leq i_2,\ldots,i_{d+2}\leq N}
\widetilde{\mathcal{A}}(i,i_2,\ldots,i_{d+2})\widetilde{\mathcal{A}}(j,i_2,\ldots,i_{d+2}),
\quad 1\leq i,j\leq N.
\end{equation}
If $\mathbf{x}_i$ and $\mathbf{x}_j$ are not in the same underlying
cluster, then all the products are zero. Therefore,
$\widetilde{\mathbf{W}}$ is block diagonal:
\begin{equation}\label{mat:W_tilde}
\widetilde{\mathbf{W}} = \diag
\{\widetilde{\mathbf{W}}^{(1)},\widetilde{\mathbf{W}}^{(2)},\ldots,\widetilde{\mathbf{W}}^{(K)}\},
\end{equation}
where $\widetilde{\mathbf{W}}^{(k)} \in \mathbb{R}^{N_k\times N_k}$,
corresponding to the underlying cluster $\widetilde{\mathrm{C}}_k$,
has the following form:
%Moreover, each block matrix $\widetilde{\mathbf{W}}^{(j)}$ has the following form
\begin{equation}
\widetilde{{W}}^{(k)}_{ij} =
  \begin{cases}
     \perm{N_k-1}{d+1}, &  \text{if}\ i=j;\\
     \perm{N_k-2}{d+1}, &  \text{otherwise}.
  \end{cases}
%\left\{
%%
%\begin{array}{ll}
%%
%\perm{N_j-1}{d+1}, & \quad \text{if}\ k=l;\\
%%
%\perm{N_j-2}{d+1}, & \quad \text{otherwise}.
%%
%\end{array}
%%
%\right.
%
\label{mat:W_tilde_j}
\end{equation}
Indeed, the diagonal elements of $\widetilde{\mathbf{W}}^{(k)}$ are
simply the number of ones in the corresponding rows of
$\widetilde{\mathbf{A}}$, and the off-diagonal elements are the
number of ones that appear at the intersection of the corresponding
pair of rows.

%\[ \mathbf{W}^{(j)}=\begin{bmatrix}
%{(N_j-1)(N_j-2)} & (N_j-2)(N_j-3) & \dots & (N_j-2)(N_j-3) & (N_j-2)(N_j-3)
%(N_j-2)(N_j-3) & {(N_j-1)(N_j-2)} & \dots & (N_j-2)(N_j-3) & (N_j-2)(N_j-3)
%\hdotsfor{5}
%(N_j-2)(N_j-3) & (N_j-2)(N_j-3) & \dots & {(N_j-1)(N_j-2)} & (N_j-2)(N_j-3)
%(N_j-2)(N_j-3) & (N_j-2)(N_j-3) & \dots & (N_j-2)(N_j-3) & {(N_j-1)(N_j-2)}
%\end{bmatrix} \in \mathbb{R}^{N_j\times N_j}\]
%for each $j=1,2,\text{or\ }3$.
It then follows that
%\[\mathbf{D}=\diag(\underset{N_1 \text{times}}{\underbrace{d_{1},\dots,d_{1}}},\underset{N_2
%\text{times}}{\underbrace{d_{2},\dots,d_{2}}},\underset{N_3
%\text{times}}{\underbrace{d_3,\dots,d_3}}),\]
\begin{equation}
\label{mat:D_tilde_diag}
\widetilde{\mathbf{D}}=\diag\{\widetilde{\mathbf{W}}\cdot\mathbf{1}\}=\diag\{\widetilde{d}_{1}\mathbf{I}_{N_1},\widetilde{d}_{2}\mathbf{I}_{N_2},\ldots,
\widetilde{d}_K\mathbf{I}_{N_K}\},
\end{equation}
where
\begin{align*}
\widetilde{d}_k&=\perm{N_k-1}{d+1}+(N_k-1)\cdot
\perm{N_k-2}{d+1}\nonumber \\ &= (N_k-d-1)\cdot \perm{N_k-1}{d+1}.
\end{align*}

The normalized matrix
$\widetilde{\mathbf{Z}}=\widetilde{\mathbf{D}}^{-{1}/{2}}\widetilde{\mathbf{W}}\widetilde{\mathbf{D}}^{-{1}/{2}}$
is also block diagonal:
\begin{equation}\label{mat:Z_tilde}
\widetilde{\mathbf{Z}} =
\diag\{\widetilde{\mathbf{Z}}^{(1)},\widetilde{\mathbf{Z}}^{(2)},\ldots,\widetilde{\mathbf{Z}}^{(K)}\},
\end{equation}
where each block has the form
$\widetilde{\mathbf{Z}}^{(k)}=\widetilde{\mathbf{W}}^{(k)}/\widetilde{d}_k,1\leq
k\leq K$. The $(i,j)$-element of $\widetilde{\mathbf{Z}}^{(k)}$, for
all $1\leq i,j\leq N_k$, is
\begin{equation}
\widetilde{{Z}}^{(k)}_{ij} =
  \begin{cases}
     \frac{1}{N_k-d-1}, & \text{if}\ i=j;\\
     \frac{N_k-d-2}{(N_k-1)(N_k-d-1)}, & \textrm{otherwise}.
  \end{cases}
\label{mat:Z_tilde_j}
\end{equation}
%\[ \mathbf{Z}^{(j)}=\begin{bmatrix}
%\frac{1}{N_j-d-1} & \frac{N_j-d-2}{(N_j-1)(N_j-d-1)} & \dots & \frac{N_j-d-2}{(N_j-1)(N_j-d-1)} & \frac{N_j-d-2}{(N_j-1)(N_j-d-1)}
%\frac{N_j-3}{(N_j-1)(N_j-d-1)} & \frac{1}{N_j-2} &\dots & \frac{N_j-d-2}{(N_j-1)(N_j-d-1)} & \frac{N_j-d-2}{(N_j-1)(N_j-d-1)}
%\hdotsfor{5}
%\frac{N_j-d-2}{(N_j-1)(N_j-d-1)} & \frac{N_j-d-2}{(N_j-1)(N_j-d-1)} & \dots & \frac{1}{N_j-d-1} & \frac{N_j-d-2}{(N_j-1)(N_j-d-1)}
%\frac{N_j-d-2}{(N_j-1)(N_j-d-1)} & \frac{N_j-d-2}{(N_j-1)(N_j-d-1)} & \dots & \frac{N_j-d-2}{(N_j-1)(N_j-d-1)} & \frac{1}{N_j-d-1}
%\end{bmatrix}.\]

Straightforward calculation shows that each block matrix
$\widetilde{\mathbf{Z}}^{(k)}$ has two distinct eigenvalues:
\begin{equation}\label{evals:Z_tilde_j}
\widetilde{\lambda}^{(k)}_n = \begin{cases} 1, &\text{if}\ n=1;\\
\frac{d+1}{(N_k-1)(N_k-d-1)}, & \text{if}\ 2\leq n\leq
N_k.\end{cases}
\end{equation}
%We note that when $N_k\geq 2(d+1), 1\leq k\leq K$, the rest of the
%eigenvalues have the following upper bound
%\[\widetilde{\lambda}^{(k)}_n \leq
%\frac{1}{2d+1},\qquad \text{for}\quad 2\leq n\leq N_k \text{ and }
%1\leq k\leq K.\]
%$\mathbf{u}_1^{(j)} = (1,1,\ldots,1)'\in\textbb{R}^{N_j}$.
%Hence the normalized matrix $\mathbf{Z}$ has the largest
%eigenvalue 1 with 3 replications and the corresponding eigenvectors
%\[ (\underset{N_1 \text{
%times}}{\underbrace{1,\ldots,1}},\underset{N_2+N_3 \text{
%times}}{\underbrace{0,\ldots,0}})', (\underset{N_1 \text{
%times}}{\underbrace{0,\ldots,0}},\underset{N_2 \text{
%times}}{\underbrace{1,\ldots,1}},\underset{N_3 \text{
%times}}{\underbrace{0,\ldots,0}})', \text{ and } (\underset{N_1+N_2
%\text{ times}}{\underbrace{0,\ldots,0}},\underset{N_3 \text{
%times}}{\underbrace{1,\ldots,1}})'.\]

The eigenspace associated with the single eigenvalue 1 for
$\widetilde{\mathbf{Z}}^{(k)}$ is spanned by $\mathbf{1}_{N_k}$, the
$N_k$-dimensional column vector of all ones. Since the eigenvalues
and eigenvectors of a block diagonal matrix are essentially the
union of those of its blocks (for eigenvectors we need to append
zeros in an appropriate way), we conclude that
$\widetilde{\mathbf{Z}}$ has the largest eigenvalue 1 of
multiplicity $K$ with associated eigenspace spanned by the following
$K$ orthonormal vectors:
\[
%\label{eq:rows_U_tilde}
\frac{1}{\sqrt{N_1}} \begin{pmatrix} \mathbf{1}_{N_1}\\  \mathbf{0}\\ \vdots \\
\mathbf{0} \end{pmatrix}, \frac{1}{\sqrt{N_2}}
\begin{pmatrix} \mathbf{0}\\ \mathbf{1}_{N_2}\\ \vdots \\ \mathbf{0} \end{pmatrix},
\ldots, \frac{1}{\sqrt{N_K}}\begin{pmatrix} \mathbf{0} \\\vdots \\
\mathbf{0} \\ \mathbf{1}_{N_K}
\end{pmatrix} \in \mathbb{R}^N.
\]

We note that the $K$ eigenvectors associated with the eigenvalue 1
can only be determined up to an orthonormal transformation. That is,
%if $\widetilde{\mathbf{U}}$ contains the $K$ eigenvectors as its
%columns, then
\begin{equation}\label{eq:U_tilde}
\widetilde{\mathbf{U}}=\begin{pmatrix}
\frac{1}{\sqrt{N_1}}\mathbf{1}_{N_1} & \mathbf{0} & \ldots & \mathbf{0} \\
\mathbf{0} & \frac{1}{\sqrt{N_2}}\mathbf{1}_{N_2} & \ldots & \mathbf{0}\\
%\hdotsfor{4} \\
\vdots & \vdots & \ddots & \vdots\\
\mathbf{0} & \mathbf{0} & \ldots &
\frac{1}{\sqrt{N_K}}\mathbf{1}_{N_K}
\end{pmatrix} \mathbf{Q} \in \mathbb{R}^{N\times K},
\end{equation}
%\begin{equation}\widetilde{\mathbf{U}}=\diag\{
%\frac{1}{\sqrt{N_1}}\mathbf{1},
%\frac{1}{\sqrt{N_2}}\mathbf{1},\ldots,
%\frac{1}{\sqrt{N_K}}\mathbf{1} \}\cdot \mathbf{Q} \in
%\mathbb{R}^{N\times K},\end{equation}
where $\mathbf{Q}$ is a $K\times K$ orthonormal matrix. %However, the
%subspace spanned by the columns of $\widetilde{\mathbf{U}}$ must be
%unique.
%Clearly, the rows of $\widetilde{\mathbf{U}}$ are $K$ orthogonal
%vectors in $\mathbb{R}^K$.

%When we renormalize each row of $\widetilde{\mathbf{U}}$ to have
%unit length, we obtain that \begin{equation} \label{eq:V_tilde}
%\widetilde{\mathbf{V}}=\begin{pmatrix}
%\mathbf{1}_{N_1} & \mathbf{0} & \ldots & \mathbf{0}\\
%\mathbf{0} & \mathbf{1}_{N_2} & \ldots & \mathbf{0}\\
%\vdots & \vdots & \ddots & \vdots\\
%\mathbf{0} & \mathbf{0} & \ldots & \mathbf{1}_{N_K}
%\end{pmatrix} \mathbf{Q}.\end{equation}

If we write
$\mathbf{Q}=(\mathbf{q}_1,\mathbf{q}_2,\ldots,\mathbf{q}_K)'$, where
$\mathbf{q}_k$ is the $k^{\text{th}}$ column of $\mathbf{Q}'$, then
equation~\eqref{eq:U_tilde} implies that the $K$ clusters are mapped
one-to-one to the $K$ mutually orthogonal vectors
$\frac{1}{\sqrt{N_1}}\cdot\mathbf{q}_1,\ldots,
\frac{1}{\sqrt{N_K}}\cdot\mathbf{q}_K \in \mathbb{R}^K$. %while
%equation~\eqref{eq:V_tilde} implies that the $K$ clusters are mapped
%one-to-one to the $K$ mutually orthonormal vectors
%$\mathbf{q}_1,\ldots, \mathbf{q}_K \in \mathbb{R}^K$.

\subsection{Proof of Lemma~\ref{lem:dist_tv}}
\label{prf:lem_dist_tv} We first note that
$P^K(\mathbf{Z})=\mathbf{U}\mathbf{U}'$ and
$P^K(\widetilde{\mathbf{Z}})=\widetilde{\mathbf{U}}\widetilde{\mathbf{U}}'$,
due to the fact that both $\mathbf{U}$ and $\widetilde{\mathbf{U}}$
are composed of orthonormal columns. Therefore,
\begin{align*}
\normF{P^K(\mathbf{Z})-P^K(\widetilde{\mathbf{Z}})}^2 &=
\normF{\mathbf{U}\mathbf{U}'-\widetilde{\mathbf{U}}\widetilde{\mathbf{U}}'}^2
= \trace\left(\left(\mathbf{U}\mathbf{U}'-\widetilde{\mathbf{U}}\widetilde{\mathbf{U}}'\right)^2\right)\\
&=
\trace\left(\mathbf{U}\mathbf{U}'-\mathbf{U}\mathbf{U}'\widetilde{\mathbf{U}}\widetilde{\mathbf{U}}'
-\widetilde{\mathbf{U}}\widetilde{\mathbf{U}}'\mathbf{U}\mathbf{U}'+\widetilde{\mathbf{U}}\widetilde{\mathbf{U}}'\right).
\end{align*}
Since
$$\trace\left(\mathbf{U}\mathbf{U}'\right)=\trace\left(\mathbf{U}'\mathbf{U}\right)=\trace(\mathbf{I}_K)=K,$$
and similarly,
$$\trace\left(\widetilde{\mathbf{U}}\widetilde{\mathbf{U}}'\right)=K,$$
we have
\begin{align*}
\normF{P^K(\mathbf{Z})-P^K(\widetilde{\mathbf{Z}})}^2
=2K-2\cdot\trace\left(\mathbf{U}\mathbf{U}'\widetilde{\mathbf{U}}\widetilde{\mathbf{U}}'\right).
%= 2K - 2 \|\mathbf{U}'\widetilde{\mathbf{U}}\|_F^2 \nonumber \\
\end{align*}

In the formula of the matrix $\widetilde{\mathbf{U}}$
(equation~\eqref{eq:U_tilde}), there is an arbitrary orthonormal
matrix $\mathbf{Q}$. However, the product
$\widetilde{\mathbf{U}}\widetilde{\mathbf{U}}'$ does not depend on
$\mathbf{Q}$. Hence, we can use a representation of
$\widetilde{\mathbf{U}}$ where $\mathbf{Q}$ is the identity matrix,
and proceed as follows:
\begin{align}
\label{eq:norm_centers_dist}
\normF{P^K(\mathbf{Z})-P^K(\widetilde{\mathbf{Z}})}^2
&= 2K - 2\cdot \normF{\mathbf{U}'\widetilde{\mathbf{U}}}^2  \nonumber \\
&= 2K - 2\cdot \normF{\left[\sum_{i\in \mathrm{I}_1}
\frac{1}{\sqrt{N_1}} \left(\mathbf{u}^{(i)}\right)' \ldots
\sum_{i\in \mathrm{I}_K}\frac{1}{\sqrt{N_K}} \left(\mathbf{u}^{(i)}\right)'\right]}^2  \nonumber\\
%&= 2K - 2\sum_{k=1}^K \left\|\sum_{i\in \mathrm{I}_k}
%\frac{1}{\sqrt{N_k}} \left(\mathbf{u}^{(i)}\right)' \right\|_2^2 \\
&= 2K-2\cdot\sum_{k=1}^K \frac{1}{N_k}\normV{\sum_{i\in
\mathrm{I}_k}\mathbf{u}^{(i)}}^2 \nonumber\\
&= 2K - 2\cdot \sum_{k=1}^K N_k\normV{\mathbf{c}^{(k)}}^2.
\end{align}

%We then use the definition of $\mathbf{c}^{(k)}, 1\leq k\leq K$ to
%rewrite the last equation and obtain that
%\begin{align}
%\|P^K(\mathbf{Z})-P^K(\widetilde{\mathbf{Z}})\|_F^2 &= 2K - 2
%\sum_{k=1}^K N_k\|\mathbf{c}^{(k)}\|_2^2.
%\end{align}

Since the columns of the matrix $\mathbf{U}$ are unit vectors, we
have
\begin{align}
\sum_{i=1}^N \normV{\mathbf{u}^{(i)}}^2 =
\normF{\mathbf{U}}^2=\sum_{k=1}^K \normV{\mathbf{u}_k}^2=K.
\end{align}
Combining the last two equations we get that
\begin{align}
\normF{P^K(\mathbf{Z})-P^K(\widetilde{\mathbf{Z}})}^2 &= 2\cdot
\left( \sum_{i=1}^N \normV{\mathbf{u}^{(i)}}^2 - \sum_{k=1}^K
N_k\cdot
\normV{\mathbf{c}^{(k)}}^2\right) \nonumber \\
%&=2 \left( \sum_{k=1}^K
%\sum_{i\in\mathrm{I}_k}\|\mathbf{u}^{(i)}\|_2^2 - \sum_{k=1}^K N_k
%\|\mathbf{c}^{(k)}\|_2^2\right) \nonumber \\
&=2\cdot \sum_{k=1}^K
\left(\sum_{i\in\mathrm{I}_k}\normV{\mathbf{u}^{(i)}}^2 - N_k\cdot
\normV{\mathbf{c}^{(k)}}^2\right) \nonumber \\
&=2\cdot\sum_{k=1}^K
\sum_{i\in\mathrm{I}_k}\normV{\mathbf{u}^{(i)}-\mathbf{c}^{(k)}}^2.
\end{align}

%Using the definitions in
%Subsection~\ref{subsubsec:measure_goodness_of_clustering}, the above
%equation becomes
%\[\dist^2(E_K(\mathbf{Z}),E_K(\widetilde{\mathbf{Z}})) = 2\cdot
%\tv.\]

\subsection{Proof of Lemma~\ref{lem:separation_leq_tv}}
\label{prf:lem_separation_leq_tv}
Equation~\eqref{eq:norm_centers_tv} is a direct consequence of
combining equation~\eqref{eq:norm_centers_dist} and
Lemma~\ref{lem:dist_tv}.

To show equation~\eqref{ineq:separation_between_centers}, we first
expand the following two products
\begin{align} \mathbf{UU}'&= \left(\langle \mathbf{u}^{(i)},
\mathbf{u}^{(j)}\rangle\right)_{1\leq i,j\leq N}, \\
\widetilde{\mathbf{U}}\widetilde{\mathbf{U}}' &=
\diag\left\{\frac{1}{N_1}\mathbf{1}_{N_1\times N_1}, \ldots,
\frac{1}{N_K}\mathbf{1}_{N_K\times N_K}\right\}. \end{align}
%where
%$\mathbf{1}_{N_k\times N_k}, 1\leq k \leq K$ is the $N_k\times N_k$
%matrix of all ones. Therefore,
Then
\begin{align}\label{eq:pert_result_rows}
&\normF{P^K(\mathbf{Z})-P^K(\widetilde{\mathbf{Z}})}^2
=\normF{\mathbf{U}\mathbf{U}'-\widetilde{\mathbf{U}}\widetilde{\mathbf{U}}'}^2
\nonumber \\
&= \sum_{1\leq k\leq K}\sum_{i,j\in \mathrm{I}_k}
\left(\langle\mathbf{u}^{(i)},\mathbf{u}^{(j)}\rangle-\frac{1}{N_k}\right)^2
+ \sum_{1\leq k\neq \ell\leq K}\sum_{i\in \mathrm{I}_k, j\in I_\ell}
\left(\langle\mathbf{u}^{(i)},\mathbf{u}^{(j)}\rangle\right)^2 \nonumber \\
& \geq \sum_{1\leq k\neq \ell\leq K}\sum_{i\in \mathrm{I}_k, j\in
I_\ell}
\left(\langle\mathbf{u}^{(i)},\mathbf{u}^{(j)}\rangle\right)^2.
\end{align}
We next apply the inequality $\left(\sum_{i=1}^m a_i\right)^2\leq
m\cdot \sum_{i=1}^m a_i^2$ and conclude that
\begin{align*}
\normF{P^K(\mathbf{Z})-P^K(\widetilde{\mathbf{Z}})}^2 &\geq
\sum_{1\leq k\neq \ell\leq K} \frac{1}{N_k N_\ell} \cdot
\left(\sum_{i\in \mathrm{I}_k, j\in I_\ell}
\langle\mathbf{u}^{(i)},\mathbf{u}^{(j)}\rangle\right)^2 \\
%&= \sum_{1\leq k\neq \ell\leq K} \frac{1}{N_k N_\ell} \cdot
%\left\langle\sum_{i\in \mathrm{I}_k}\mathbf{u}^{(i)},\sum_{j\in I_\ell}\mathbf{u}^{(j)}\right\rangle^2 \\
&=\sum_{1\leq k\neq \ell\leq K} N_k N_\ell \cdot
\langle\mathbf{c}^{(k)},\mathbf{c}^{(\ell)}\rangle^2.
\end{align*}
Finally, combining the last equation and Lemma~\ref{lem:dist_tv}
completes the proof.

\subsection{Review of Principal Angles and Proof of Lemma~\ref{lem:dist_principal_angles}}
\label{subsec:review_principal_angles}
\subsubsection*{Review of Principal Angles}
The principal angles $0\leq \theta_1\leq\cdots\leq\theta_K\leq
\pi/2$ between two $K$-dimensional subspaces $S$ and $T$ are defined
recursively as follows (see e.g., \cite{Golub96}):
\begin{align*}
\cos\theta_1 &= \max_{\mathbf{x}\in
S,\normV{\mathbf{x}}=1}\;\max_{\mathbf{y}\in T,\normV{\mathbf{y}}=1}
\mathbf{x}'\mathbf{y} = \mathbf{x}'_1
\mathbf{y}_1,\\
\cos\theta_2 &= \max_{\latop{\mathbf{x}\in
S,\normV{\mathbf{x}}=1}{\mathbf{x}\perp
\mathbf{x}_1}}\;\max_{\latop{\mathbf{y}\in T,\normV{\mathbf{y}}=1}{
\mathbf{y}\perp \mathbf{y}_1}} \mathbf{x}'\mathbf{y} = \mathbf{x}'_2
\mathbf{y}_2,\\
\ldots &  \ldots \\
\cos\theta_K &= \max_{\latop{\mathbf{x}\in
S,\normV{\mathbf{x}}=1}{\mathbf{x}\perp
\{\mathbf{x}_1,\ldots,\mathbf{x}_{K-1}\}}}\;\max_{\latop{\mathbf{y}\in
T,\normV{\mathbf{y}}=1}{\mathbf{y}\perp
\{\mathbf{y}_1,\ldots,\mathbf{y}_{K-1}\}}} \mathbf{x}'\mathbf{y} =
\mathbf{x}'_K \mathbf{y}_K.
\end{align*}

Another formula for the cosines of the principal angles is obtained
as follows. Let $\mathbf{S}$ and $\mathbf{T}$ be two matrices whose
columns define orthonormal bases of $S$ and $T$ respectively. Since
any $\mathbf{x}\in S$ and $\mathbf{y}\in T$ can be represented as
$\mathbf{x}=\mathbf{Su}$ and $\mathbf{y}=\mathbf{Tv}$ respectively,
where $\mathbf{u}$ and $\mathbf{v}$ are unit vectors in
$\mathbb{R}^K$, it follows that
\[
\cos\theta_k = \sigma_k\left(\mathbf{S}'\mathbf{T}\right) \ \text{
for } \ 1\leq k \leq K ,
\]
where $\sigma_k\left(\mathbf{S}'\mathbf{T}\right)$ denotes the
$k^{\text{th}}$ largest singular value of $\mathbf{S}'\mathbf{T}$.
%We can reformulate Theorem~\ref{thm:perturbation_main} in terms of
%principal angles as follows.
%%
%\newtheorem*{theorema}{Theorem~\ref{thm:perturbation_main}$'$}
%\begin{theorema}\label{thm:perturbation_principal_angles}
%Let $0\leq\theta_1\leq \theta_2\leq \cdots \leq \theta_K\leq \pi/2$
%be the $K$ principal angles between the two subspaces spanned by the
%first $K$ eigenvectors of $\mathbf{Z}$ and $\widetilde{\mathbf{Z}}$.
%%Let Assumption~\ref{assmp:N_j} hold as
%%well as
%If either Assumption~\ref{assmp:large_D_ii} or
%Assumption~\ref{assmp:weaker_than_d_ii} holds, then
%\begin{align}
%%\label{ineq:pert_result_principal_angles}
%\sum_{k=1}^{K}\sin^2\theta_k\leq 1/2\cdot
%C(K,d,\varepsilon_1,\varepsilon_2) \cdot
%N^{-(d+2)}\norm{\mathcal{E}}_F^2.
%\end{align}
%\end{theorema}
\subsubsection*{Proof of Lemma~\ref{lem:dist_principal_angles}}
From the proof of Lemma~\ref{lem:dist_tv} we have that
\begin{align}
\normF{P^K(\mathbf{Z})-P^K(\widetilde{\mathbf{Z}})}^2 &= 2K - 2
\normF{\mathbf{U}'\widetilde{\mathbf{U}}}^2
=2K-2\sum_{k=1}^{K}\sigma_k^2\left(\mathbf{U}'\widetilde{\mathbf{U}}\right) \nonumber\\
&= 2K - 2\sum_{k=1}^{K}\cos^2\theta_k
 = 2\sum_{k=1}^{K}\sin^2\theta_k. \nonumber
\label{eq:diff_proj_Z}
\end{align}
%\end{proof}

\subsection{Proof of Theorem~\ref{thm:perturbation_main}}
\label{prf:thm_perturbation_main}
The proof is based on a perturbation result by Zwald and
Blanchard~\cite[Theorem 3]{Zwald06}. In fact, we only need a special
case of it which is formulated below.
\begin{theorem}[Matrix version of Theorem~3 in Zwald and Blanchard~\cite{Zwald06}]
\label{thm:zwald} Let $\mathbf{S}$ be a symmetric positive square
matrix with nonzero eigenvalues
$\lambda_1\geq\cdots\geq\lambda_K>\lambda_{K+1}\geq\cdots\geq 0$,
where $K>0$ is an integer. Define
$\delta_K=\lambda_K-\lambda_{K+1}>0$, which denotes the
$K^{\text{th}}$ eigengap of $\mathbf{S}$. Let $\mathbf{B}$ be
another symmetric matrix such that $\normF{\mathbf{B}}<\delta_K/4$
and $\mathbf{S}+\mathbf{B}$ is still a positive matrix. Then
\begin{equation}\label{eq:operator_peturbation}
\normF{P^K(\mathbf{S}+\mathbf{B})-P^K(\mathbf{S})}\leq
2\normF{\mathbf{B}}/{\delta_K}.
\end{equation}
\end{theorem}

In order to apply the above theorem to the quantity
$\normF{P^K(\mathbf{Z})-P^K(\widetilde{\mathbf{Z}})}$, we need a
lower bound on $\widetilde{\delta}_K$, the $K^{\text{th}}$ eigengap
of $\widetilde{\mathbf{Z}}$, and an upper bound on the Frobenius
norm of the difference $\mathbf{B}:=
\mathbf{Z}-\widetilde{\mathbf{Z}}$. While the former bound is
immediate, we find the latter bound somewhat challenging.

First, equation~\eqref{evals:Z_tilde_j}, together with
$N_1=\min_{1\leq k\leq K} N_k$, implies that:
\be
\widetilde{\delta}_K=1-\frac{d+1}{(N_1-1)(N_1-d-1)}.
\label{eq:eigengap}
\ee
Since $N_1\geq 2(d+1)+1$ by equation~\eqref{eq:comparable_sizes}, we then
obtain that
\begin{equation}\label{ineq:delta_K}
\widetilde{\delta}_K\geq \frac{2d+3}{2d+4}\geq \frac{3}{4}.
\end{equation}

Next, we estimate the Frobenius norm of the perturbation
$\mathbf{B}$ as follows. Using the definitions of the matrices
$\mathbf{Z}$ and $\mathbf{W}$, we rewrite $\mathbf{B}$ in the
following way:
\begin{align*}
%\label{eq:pert_B}
\mathbf{B} %&= {\mathbf{D}}^{-1/2}{\mathbf{W}}{\mathbf{D}}^{-1/2} -
%\widetilde{\mathbf{D}}^{-1/2}\widetilde{\mathbf{W}}\widetilde{\mathbf{D}}^{-1/2}
&={\mathbf{D}}^{-1/2}{\mathbf{A}}{\mathbf{A}}'{\mathbf{D}}^{-1/2} -
\widetilde{\mathbf{D}}^{-1/2}\widetilde{\mathbf{A}}\widetilde{\mathbf{A}}'\widetilde{\mathbf{D}}^{-1/2}.
\end{align*}
Regrouping terms gives that
\begin{align*}
\mathbf{B} %&= \|\widetilde{\mathbf{Z}}-\mathbf{Z}\|_F
   &=\left({\mathbf{D}}^{-1/2}{\mathbf{A}}-\widetilde{\mathbf{D}}^{-1/2}\widetilde{\mathbf{A}}\right)
     \left({\mathbf{D}}^{-1/2}{\mathbf{A}}-\widetilde{\mathbf{D}}^{-1/2}\widetilde{\mathbf{A}}\right)'\\
&\qquad +
\left({\mathbf{D}}^{-1/2}{\mathbf{A}}-\widetilde{\mathbf{D}}^{-1/2}\widetilde{\mathbf{A}}\right)\widetilde{\mathbf{A}}'\widetilde{\mathbf{D}}^{-1/2}
+
\widetilde{\mathbf{D}}^{-1/2}\widetilde{\mathbf{A}}\left({\mathbf{D}}^{-1/2}{\mathbf{A}}-\widetilde{\mathbf{D}}^{-1/2}\widetilde{\mathbf{A}}\right)'.
\end{align*}
We thus get an initial upper bound on its Frobenius norm:
\begin{align}
\label{eq:first_B_bound}
\normF{\mathbf{B}} %&= \|\widetilde{\mathbf{Z}}-\mathbf{Z}\|_F
   \leq
        \normF{{\mathbf{D}}^{-1/2}{\mathbf{A}}-\widetilde{\mathbf{D}}^{-1/2}\widetilde{\mathbf{A}}}^2
         +
        2\normF{\widetilde{\mathbf{D}}^{-1/2}\widetilde{\mathbf{A}}}\normF{{\mathbf{D}}^{-1/2}{\mathbf{A}}-\widetilde{\mathbf{D}}^{-1/2}
\widetilde{\mathbf{A}}}.
\end{align}

By using equations \eqref{mat:Z_tilde} and \eqref{mat:Z_tilde_j}, we
get that
\begin{align*}
\normF{{\widetilde{\mathbf{D}}}^{-1/2}{\widetilde{\mathbf{A}}}}^2
=\trace\Big({{\widetilde{\mathbf{D}}}^{-1/2}{\widetilde{\mathbf{W}}}{\widetilde{\mathbf{D}}}^{-1/2}}\Big)
=\trace\big(\widetilde{\mathbf{Z}}\big)
=\sum_{k=1}^K\frac{N_k}{N_k-d-1}.
\end{align*}
Equation~\eqref{eq:comparable_sizes} implies that
\[
\frac{N_k}{N_k-d-1}< 2, \quad 1\leq k\leq K.
\]
Consequently, we have
\[
\normF{{\widetilde{\mathbf{D}}}^{-1/2}{\widetilde{\mathbf{A}}}}^2 <
2 K,
\]
and thus equation~\eqref{eq:first_B_bound} becomes
\begin{equation}
\normF{\mathbf{B}} <
                 \normF{{\mathbf{D}}^{-1/2}{\mathbf{A}}-\widetilde{\mathbf{D}}^{-1/2}\widetilde{\mathbf{A}}}^2
                 +
                 2\sqrt{2K}\cdot\normF{{\mathbf{D}}^{-1/2}{\mathbf{A}}-\widetilde{\mathbf{D}}^{-1/2}\widetilde{\mathbf{A}}}.
\label{ineq:est_norm_of_B}
\end{equation}
Therefore, in order to control $\normF{\mathbf{B}}$, we only need to
bound
$\normF{{\mathbf{D}}^{-1/2}{\mathbf{A}}-\widetilde{\mathbf{D}}^{-1/2}\widetilde{\mathbf{A}}}$.

Let $$\mathbf{E}:=\mathbf{A}-\widetilde{\mathbf{A}}.$$ Replacing
${\mathbf{A}}$ with $\widetilde{\mathbf{A}}+\mathbf{E}$ yields that
\begin{align}
\normF{{\mathbf{D}}^{-1/2}{\mathbf{A}}-\widetilde{\mathbf{D}}^{-1/2}\widetilde{\mathbf{A}}}
&=
\normF{\left({\mathbf{D}}^{-1/2}-\widetilde{\mathbf{D}}^{-1/2}\right)\widetilde{\mathbf{A}}
+ \mathbf{D}^{-1/2}{\mathbf{E}}} \nonumber \\
&\leq
\normF{\left({\mathbf{D}}^{-1/2}-\widetilde{\mathbf{D}}^{-1/2}\right)\widetilde{\mathbf{A}}}+
\normF{{\mathbf{D}}^{-1/2}{\mathbf{E}}}.
\label{ineq:diff_D_neg_sqrt_A_first}
\end{align}

The second term on the right hand side of
equation~\eqref{ineq:diff_D_neg_sqrt_A_first} is bounded as follows
\begin{align}
\normF{\mathbf{D}^{-1/2}{\mathbf{E}}} &\leq
\normV{\mathbf{D}^{-1/2}} \cdot \normF{{\mathbf{E}}} \nonumber \leq
\normV{(\varepsilon_2\widetilde{\mathbf{D}})^{-1/2}}\cdot\normF{\mathbf{E}} \nonumber \\
&=
\left(\varepsilon_2\widetilde{d}_1\right)^{-1/2}\cdot\normF{\mathbf{E}},
\label{ineq:normF_D_tilde_neg_sqrt_E}
\end{align}
in which the second inequality follows from
Assumption~\ref{assmp:large_D_ii}
($\mathbf{D}\geq\varepsilon_2\widetilde{\mathbf{D}}>0$), and the
last equality is due to our convention: $N_1=\min_{1\leq k\leq
K}N_k$ (which implies that $\widetilde{d}_1=\min_{1\leq k\leq
K}\widetilde{d}_k$).

Bounding the first term of the right hand side of
equation~\eqref{ineq:diff_D_neg_sqrt_A_first} requires more work. We
estimate it as follows:
\begin{align}
\normF{\left({\mathbf{D}}^{-1/2}-\widetilde{\mathbf{D}}^{-1/2}\right)\cdot\widetilde{\mathbf{A}}}
&=
\normF{\widetilde{\mathbf{D}}^{-1/2}{\mathbf{D}}^{-1/2}\left({\mathbf{D}}^{1/2}+\widetilde{\mathbf{D}}^{1/2}\right)^{-1}
\left({\mathbf{D}}-\widetilde{\mathbf{D}}\right) \cdot \widetilde{\mathbf{A}}}\nonumber\\
&\leq
\normF{\widetilde{\mathbf{D}}^{-1/2}\left(\varepsilon_2\widetilde{\mathbf{D}}\right)^{-1/2}
\left(\widetilde{\mathbf{D}}^{1/2}\right)^{-1} \left({\mathbf{D}}-
\widetilde{\mathbf{D}}\right) \cdot
\widetilde{\mathbf{A}}}\nonumber\\
&=\varepsilon_2^{-1/2}\normF{\widetilde{\mathbf{D}}^{-3/2}
\left({\mathbf{D}}- \widetilde{\mathbf{D}}\right)\cdot
\widetilde{\mathbf{A}}}. \label{ineq:1}
\end{align}

We proceed by using the index sets
$\mathrm{I}_1,\ldots,\mathrm{I}_K$ (see equation~\eqref{eq:I_k}) to
expand the last equation:
\begin{align}
\normF{\left({\mathbf{D}}^{-1/2}-\widetilde{\mathbf{D}}^{-1/2}\right)\cdot\widetilde{\mathbf{A}}}
&\leq \varepsilon_2^{-1/2} \sqrt{\sum_{1\leq k \leq K}\sum_{i\in
\mathrm{I}_k} \left(D_{ii}- \widetilde{d}_k\right)^2
\widetilde{d}_k^{-3} \cdot
\normV{\widetilde{\mathbf{A}}(i,:)}^2}\nonumber\\
&= \varepsilon_2^{-1/2} \sqrt{\sum_{1\leq k \leq K}\sum_{i\in
\mathrm{I}_k}
\frac{\left(D_{ii}-\widetilde{d}_k\right)^2}{(N_k-d-1)\cdot\widetilde{d}_k^2}}\nonumber\\
&\leq \varepsilon_2^{-1/2}\widetilde{d}_1^{-1}(N_1-d-1)^{-1/2}\cdot
\normF{\mathbf{D}-\widetilde{\mathbf{D}}}. \label{ineq:2}
\end{align}
%
%Using that $\mathbf{D}=\diag\{\mathbf{W}\cdot\mathbf{1}\}$,
%$\widetilde{\mathbf{D}}=\diag\{\widetilde{\mathbf{W}}\cdot\mathbf{1}\}$,
Using the definitions of $\mathbf{D}$ and $\widetilde{\mathbf{D}}$,
we obtain that
\begin{align}
\normF{\mathbf{D}-\widetilde{\mathbf{D}}} &=
\normV{\left({\mathbf{W}}-\widetilde{\mathbf{W}}\right)\cdot\mathbf{1}_N}
\leq \normF{{\mathbf{W}}-\widetilde{\mathbf{W}}}\cdot \normV{\mathbf{1}_N}\nonumber\\
&=
{N}^{1/2}\cdot \normF{\widetilde{\mathbf{A}}\mathbf{E}'+\mathbf{E}\widetilde{\mathbf{A}}'+\mathbf{E}\mathbf{E}'}\nonumber \\
&\leq {N}^{1/2}\cdot
\left(2\normF{\widetilde{\mathbf{A}}}\normF{\mathbf{E}}+\normF{\mathbf{E}}^2\right).
\label{ineq:3}
\end{align}
Combining equations~\eqref{ineq:2} and~\eqref{ineq:3} and applying
$N_1-d-1>\frac{N_1}{2}\geq \frac{\varepsilon_1 N}{2K}$ (following
equation~\eqref{eq:comparable_sizes})
gives that
\begin{align}
\normF{\left({\mathbf{D}}^{-1/2}-\widetilde{\mathbf{D}}^{-1/2}\right)\widetilde{\mathbf{A}}}
&\leq
\left(\frac{2K}{\varepsilon_1\varepsilon_2}\right)^{1/2}\widetilde{d}_1^{-1}
\left(2\normF{\widetilde{\mathbf{A}}}\normF{\mathbf{E}}+\normF{\mathbf{E}}^2\right).
\label{ineq:diff_D_neg_sqrt_A_mid2}
\end{align}

By substituting equations~\eqref{ineq:normF_D_tilde_neg_sqrt_E} and
\eqref{ineq:diff_D_neg_sqrt_A_mid2} into
equation~\eqref{ineq:diff_D_neg_sqrt_A_first}, we arrive at
\begin{align}
\normF{{\mathbf{D}}^{-1/2}{\mathbf{A}}-\widetilde{\mathbf{D}}^{-1/2}\widetilde{\mathbf{A}}}
&\leq \left(\frac{2K}{\varepsilon_1\varepsilon_2}\right)^{1/2}
\widetilde{d}_1^{-1}
\left(2\normF{\widetilde{\mathbf{A}}}\normF{\mathbf{E}}+\normF{\mathbf{E}}^2\right)
\nonumber \\
& \qquad + \varepsilon_2^{-1/2}\widetilde{d}_1^{-1/2}
\normF{{\mathbf{E}}}. \label{ineq:diff_D_neg_sqrt_A_mid}
\end{align}
In order to complete the above estimate for
$\normF{{\mathbf{D}}^{-1/2}{\mathbf{A}}-\widetilde{\mathbf{D}}^{-1/2}\widetilde{\mathbf{A}}}$,
we need to estimate $\normF{\widetilde{\mathbf{A}}}$  from above
\begin{gather*}
\normF{\widetilde{\mathbf{A}}} < \sqrt{N^{d+2}} = N^{(d+2)/2},
%\norm{\mathbf{E}}_F\leq \sqrt{N^{d+2}} = N^{(d+2)/2} .
\end{gather*}
and $\widetilde{d}_1$ from below
\begin{align*}
\widetilde{d}_1 = (N_1-d-1) \cdot \perm{N_1-1}{d+1} \geq
(N_1/2)^{d+2} \geq \left(\frac{\varepsilon_1 N}{2K}\right)^{d+2}.
\end{align*}
We also note that all the elements of the matrix $\mathbf{E}$ are
between -1 and 1, and thus $$\normF{\mathbf{E}}\leq N^{(d+2)/2}.$$
We then continue from equation~\eqref{ineq:diff_D_neg_sqrt_A_mid},
together with the last three estimates, and get that
\begin{align}
&\ \normF{{\mathbf{D}}^{-1/2}{\mathbf{A}}-\widetilde{\mathbf{D}}^{-1/2}\widetilde{\mathbf{A}}} \nonumber \\
&\leq
\left(\frac{2K}{\varepsilon_1\varepsilon_2}\right)^{1/2}\left(\frac{\varepsilon_1
N}{2K}\right)^{-(d+2)} 3N^{(d+2)/2}\normF{\mathbf{E}}
+ \varepsilon_2^{-1/2} \left(\frac{\varepsilon_1 N}{2K}\right)^{-(d+2)/2}\normF{\mathbf{E}} \nonumber \\
& \leq 4\varepsilon_2^{-1/2}
\left(\frac{2K}{\varepsilon_1}\right)^{d+5/2}N^{-(d+2)/2}\normF{\mathbf{E}}.
\label{ineq:diff_D_neg_sqrt_A_final}
\end{align}

Finally, it follows from equations~\eqref{ineq:est_norm_of_B}
and~\eqref{ineq:diff_D_neg_sqrt_A_final} that
\begin{align}\label{ineq:est_norm_of_B_final}
\normF{\mathbf{B}} &\leq C_0(K,d,\varepsilon_1,\varepsilon_2) \cdot
N^{-(d+2)/2}\normF{\mathbf{E}},
\end{align}
where
\begin{equation}\label{eq:constant_C0}
C_0(K,d,\varepsilon_1,\varepsilon_2) := 16 \varepsilon_2^{-1}
\left(\frac{2K}{\varepsilon_1}\right)^{2d+5}+ 2\sqrt{2K}\cdot
4\varepsilon_2^{-1/2} \left(\frac{2K}{\varepsilon_1}\right)^{d+5/2}.
\end{equation}
By combining Theorem~\ref{thm:zwald} with
equations~\eqref{ineq:delta_K} and \eqref{ineq:est_norm_of_B_final},
we obtain that when $$C_0(K,d,\varepsilon_1,\varepsilon_2)\cdot
N^{-(d+2)/2}\normF{\mathbf{E}}< 3/16,$$ then
\begin{align}\label{ineq:projector_norm_E}
\normF{P^K(\mathbf{Z})-P^K(\widetilde{\mathbf{Z}})}&\leq 8/3\cdot
C_0(K,d,\varepsilon_1,\varepsilon_2)\cdot
N^{-(d+2)/2}\normF{\mathbf{E}}.
\end{align}
Letting
\begin{equation}\label{eq:constant_C1}
C_1(K,d,\varepsilon_1,\varepsilon_2):=32/9\cdot
C^2_0(K,d,\varepsilon_1,\varepsilon_2),
\end{equation}
and noting \[\normF{\mathbf{E}}\equiv \normF{\mathcal{E}},\] we
complete the proof by combining Lemma~\ref{lem:dist_tv} and
equations~\eqref{ineq:projector_norm_E} and \eqref{eq:constant_C1}.

%\subsection{Proof of Proposition~\ref{prop:perfect_tensor_unnorm}}
%\label{prf:prop_perfect_tensor_unnorm} The proof follows that of
%Proposition~\ref{prop:perfect_tensor} in
%Appendix~\ref{prf:prop_perfect_tensor}. Specifically, we start with
%equations~\eqref{mat:W_tilde} and~\eqref{mat:W_tilde_j}, and obtain
%by straightforward calculation that the eigenvalues of the matrix
%$\widetilde{\mathbf{W}}$.

\subsection{Proof of Theorem~\ref{thm:perturbation_main_unnorm}}
\label{prf:thm_perturbation_main_unnorm}
The proof proceeds in parallel to that of
Theorem~\ref{thm:perturbation_main}. That is, we bound from below
the $K^{\text{th}}$ eigengap $\widetilde{\delta}_K$ of
$\widetilde{\mathbf{W}}$, estimate from above the Frobenius norm of
the perturbation $\mathbf{B}:=\mathbf{W}-\widetilde{\mathbf{W}}$,
and then conclude the theorem by combining these two bounds with
Theorem~\ref{thm:zwald}.

Straightforward calculation shows that the matrix
$\widetilde{\mathbf{W}}$ (see formula in
Equation~\eqref{mat:W_tilde_j}) has the following eigenvalues:
\[ \widetilde{d}_K\geq \dotsb \geq \widetilde{d}_2
\geq\widetilde{d}_1 \text{ and } \nu_K\geq \dotsb \geq \nu_2\geq
\nu_1,\] where $\widetilde{d}_k, 1\leq k\leq K$, are defined in
equation~\eqref{eq:d_j}, and $$\nu_k := (d+1)\cdot\perm{N_k-2}{d},
\quad k=1,\ldots,K.$$ Using equation~\eqref{eq:comparable_sizes} we
obtain that
\be \label{eq:bound_N_K} N_K = N - \sum_{k=1}^{K-1} N_k \leq
N-(K-1)\cdot \frac{\varepsilon_1 N}{K} =
\left(1-\frac{K-1}{K}\varepsilon_1\right)\cdot N. \ee
The above equation together with equations~\eqref{eq:comparable_sizes}
and \eqref{ineq:N_unnorm} implies that
\begin{align}
\widetilde{\delta}_K &= \widetilde{d}_1-\nu_K \nonumber\\
&\geq \left(\frac{N_1}{2}\right)^{d+2}- \ \ (d+1)\cdot {N_K}^{d}\nonumber\\
&\geq \left(\frac{\varepsilon_1 N}{2K}\right)^{d+2}- (d+1)\cdot
{\left(1-\frac{K-1}{K}\varepsilon_1\right)}^{d}N^d \nonumber \\
&\geq \frac{1}{2}\left(\frac{\varepsilon_1N}{2K}\right)^{d+2}.
\label{eq:unnorm_delta}
\end{align}

We follow by bounding the magnitude of the perturbation
$\mathbf{B}=\widetilde{\mathbf{W}}-{\mathbf{W}}$:
\begin{align} \label{eq:unnorm_B}
\normF{\mathbf{B}}
&=\normF{\mathbf{A}\mathbf{E}'+\mathbf{E}\widetilde{\mathbf{A}}'}
\leq\normF{\mathbf{A}}\normF{\mathbf{E}}+\normF{\mathbf{E}}\normF{\widetilde{\mathbf{A}}}
\leq 2 N^{(d+2)/2}\normF{\mathbf{E}}.
%                 &=2\sqrt{\sum_{j=1}^3 N_j(N_j-1)\ldots(N_j-d-1)}\cdot\|\mathbf{E}\|_F+\|\mathbf{E}\|^2_F
\end{align}
Therefore, by combining equations~\eqref{eq:unnorm_delta}
and~\eqref{eq:unnorm_B} with Theorem~\ref{thm:zwald} we conclude
that when \[N^{-(d+2)/2}\normF{\mathbf{E}} \leq
\frac{1}{16}\left(\frac{\varepsilon_1}{2K}\right)^{d+2},\] we have
\begin{equation*}
\normF{P^K(\mathbf{W})-P^K(\widetilde{\mathbf{W}})}\leq
8\left(\frac{2K}{\varepsilon_1}\right)^{d+2}
N^{-(d+2)/2}\normF{\mathbf{E}}.
\end{equation*}
Theorem~\ref{thm:perturbation_main_unnorm} is then a direct
consequence of combining the above equation and
Lemma~\ref{lem:dist_tv}.

\subsection{Proof of Lemma~\ref{lem:id_error}}
\label{prf:lem_id_error} In the $\mathbf{T}$ space the centers of
the underlying clusters are
\begin{align}
\mathbf{c}_\mathbf{T}^{(k)} &:= \sqrt{N_k}\cdot
\mathbf{c}^{(k)},\quad 1\leq k\leq K.
\end{align}
Applying Lemma~\ref{lem:separation_leq_tv} with $K=2$ gives that
\begin{align*}
\normV{\mathbf{c}_\mathbf{T}^{(1)}-\mathbf{c}_\mathbf{T}^{(2)}}^2 &=
N_1\cdot
\normV{\mathbf{c}^{(1)}}^2+N_2\cdot\normV{\mathbf{c}^{(2)}}^2 -
2\sqrt{N_1N_2}\cdot \langle \mathbf{c}^{(1)},\mathbf{c}^{(2)}\rangle\\
& \geq 2-\tv-2\sqrt{\tv}.
\end{align*}

When \[\tv<\left(\sqrt{3}-1\right)^2,\] we can let
\[\tau:=\sqrt{2-\tv-2\sqrt{\tv}}.
\] Then the clustering identification error of TSCC in the
$\mathbf{T}$ space is bounded as follows:
\begin{equation*}
e_\mathrm{id}(\mathbf{T}) \leq \frac{1}{N}\cdot\sum_{k=1}^2
\#\left\{i\in\mathrm{I}_k \mid
\normV{\mathbf{t}^{(i)}-\mathbf{c}_\mathbf{T}^{(k)}}\geq \tau/2\right\}.
\end{equation*}

For each $k=1,2$, we apply Chebyshev's inequality and obtain that
\begin{align*}
%\sum_{i\in\mathrm{I}_k}
%\norm{\mathbf{t}^{(i)}-\mathbf{c}_\mathbf{T}^{(k)}}_2^2 &\geq
%\sum_{\norm{\mathbf{t}^{(i)}-\mathbf{c}_\mathbf{T}^{(k)}}_2\geq
%\tau/2} \left(\frac{\tau}{2}\right)^2 \\ &= \frac{\tau^2}{4} \cdot
\#\left\{i\in\mathrm{I}_k
\mid\normV{\mathbf{t}^{(i)}-\mathbf{c}_\mathbf{T}^{(k)}}\geq
\tau/2\right\} \leq \frac{4}{\tau^2}\sum_{i\in\mathrm{I}_k}
\normV{\mathbf{t}^{(i)}-\mathbf{c}_\mathbf{T}^{(k)}}^2.
\end{align*}
Thus,
\begin{align*}
e_\mathrm{id}(\mathbf{T})&\leq \frac{1}{N}\cdot \sum_{k=1}^2
\frac{4}{\tau^2} \sum_{i\in\mathrm{I}_k}
\normV{\mathbf{t}^{(i)}-\mathbf{c}_\mathbf{T}^{(k)}}^2 \\
&\leq \frac{4}{\tau^2} \sum_{k=1}^2 \frac{N_k}{N}
\sum_{i\in\mathrm{I}_k}
\normV{\mathbf{u}^{(i)}-\mathbf{c}^{(k)}}^2 \\
&\leq \frac{4}{\tau^2}\cdot \tv.
\end{align*}

In the $\mathbf{U}$ space, we also apply
Lemma~\ref{lem:separation_leq_tv} with $K=2$, together with the
assumptions $N_2 \geq N_1 \geq \varepsilon_1\cdot N/2$, and obtain
that
\begin{align*}
\normV{\mathbf{c}^{(1)}-\mathbf{c}^{(2)}}^2 &=
\normV{\mathbf{c}^{(1)}}^2+\normV{\mathbf{c}^{(2)}}^2 - 2 \cdot
\langle
\mathbf{c}^{(1)},\mathbf{c}^{(2)}\rangle \\
&\geq \frac{1}{N_2}\cdot \left(N_1\ \normV{\mathbf{c}^{(1)}}^2+ N_2\
\normV{\mathbf{c}^{(2)}}^2\right)
-\frac{2}{\sqrt{N_1N_2}}\cdot\sqrt{N_1N_2}\ \langle
\mathbf{c}^{(1)},\mathbf{c}^{(2)}\rangle \\ &\geq \frac{1}{N_2}\cdot
(2-\tv) -\frac{2}{N_1}\sqrt{\tv}\\
& \geq \frac{1}{N}\cdot \left(2-\tv-4/\varepsilon_1\cdot
\sqrt{\tv}\right).
\end{align*}

When
\[\tv<\left(\sqrt{2+\frac{4}{\varepsilon_1^2}}-\frac{2}{\varepsilon_1}\right)^2,\]
we can %let \[ \kappa:= \sqrt{2-\tv
%-\frac{4}{\varepsilon_1}\cdot\sqrt{\tv}},
%\] and
apply similar steps as above to obtain that
\begin{align*}
e_\mathrm{id}(\mathbf{U})
&\leq
\frac{4\tv}{2-\tv-4/\varepsilon_1\cdot \sqrt{\tv}} .
\end{align*}

\subsection{Proofs of Main Statements of
Subsection~\ref{subsubsec:existence_assmp1}}
\label{prf:lem_large_degrees}
\subsubsection{Proof of Lemma~\ref{lem:large_degrees1}}
\label{prf:lem_large_degrees1}
For any $1\leq k\leq K$ and $i\in \mathrm{I}_k$, we have
\begin{align}\label{ineq:degrees_sum_of_weights}
D_{ii}&\geq \sum_{j\in \mathrm{I}_k} W_{ij} \geq \sum_{j\in
\mathrm{I}_k}\sum_{i_2,\dotsc,i_{d+2}\in \mathrm{I}_k}
\mathcal{A}(i,i_2,\dotsc,i_{d+2})\mathcal{A}(j,i_2,\dotsc,i_{d+2})\nonumber \\
&= \sum_{j\in \mathrm{I}_k}\sum_{\latop{i_2,\dotsc,i_{d+2}\in
\mathrm{I}_k \setminus \{i,j\}}{\mathrm{and\ are\ distinct}}}
e^{-\frac{c_{\mathrm{p}}\left(\mathbf{x}_i,\mathbf{x}_{i_2},\dotsc,\mathbf{x}_{i_{d+2}}\right)+
c_{\mathrm{p}}\left(\mathbf{x}_j,\mathbf{x}_{i_2},\dotsc,\mathbf{x}_{i_{d+2}}\right)}{\sigma}}.
\end{align}

When the given data is noiseless, the polar curvature of any
distinct $d+2$ points in $\widetilde{\mathrm{C}}_k$ is zero. Hence,
\begin{align*}
D_{ii} &\geq
\sum_{j\in\mathrm{I}_k}\sum_{\latop{i_2,\dotsc,i_{d+2}\in
\mathrm{I}_k \setminus \{i,j\}}{\mathrm{and\ are\ distinct}}} 1 =
\widetilde{d}_k,
\end{align*}
where $\widetilde{d}_k$ (replicated $N_k$ times), $1\leq k\leq K$,
are the diagonal elements of $\widetilde{\mathbf{D}}$ (see
equation~\eqref{eq:d_j}). We have thus proved that $\mathbf{D}\geq
\widetilde{\mathbf{D}}$.

\subsubsection{Proof of Equation~\eqref{eq:naive_eps_2}}
\label{prf:lem_large_degrees2}
We note the following obvious bound on the polar curvature of any
sampled points $\mathbf{x}_{i_1},\dotsc,\mathbf{x}_{i_{d+2}}$ from
the ball $B$ according to Problem~\ref{prob:hlm}:
$$c_{\mathrm{p}}(\mathbf{x}_{i_1},\mathbf{x}_{i_2},\dotsc,\mathbf{x}_{i_{d+2}})\leq
\diam(B)\cdot \sqrt{d+2}.$$
Combining this bound with
equation~\eqref{ineq:degrees_sum_of_weights} we obtain that
\begin{align*}
D_{ii} &\geq \sum_{j\in
\mathrm{I}_k}\sum_{\latop{i_2,\dotsc,i_{d+2}\in \mathrm{I}_k \setminus
\{i,j\}}{\mathrm{and\ are\ distinct}}} e^{-\frac{\sqrt{d+2}\diam(B)}{\sigma}}
e^{-\frac{\sqrt{d+2}\diam(B)}{\sigma}} = e^{-\frac{2 \sqrt{d+2}
\diam(B)}{\sigma}}\cdot \widetilde{d}_k.
\end{align*}

\subsubsection{Proof
 of Lemma~\ref{lem:large_degrees3}}
\label{prf:lem_large_degrees3}
We take the expectation of each side of
equation~\eqref{ineq:degrees_sum_of_weights} with respect to the
measure $\mu_\mathrm{p}$ (defined in equation~\eqref{eq:mu_p}), and
proceed using Jensen's inequality (twice) as follows:
\begin{align*}
E_{\mu_\mathrm{p}}(D_{ii})
%E_{\mu_\mathrm{p}}\left(\sum_{j\in \mathrm{I}_k} W_{ij}\right) &\geq
%\sum_{j\in \mathrm{I}_k}\sum_{\latop{i_2,\dotsc,i_{d+2}\in
%\mathrm{I}_k \setminus \{i,j\}}{\mathrm{and\ are\ distinct}}} E
%e^{-\frac{c_{\mathrm{p}}(\mathbf{x}_i,\mathbf{x}_{i_2},\dotsc,\mathbf{x}_{i_{d+2}})+
%c_{\mathrm{p}}(\mathbf{x}_j,\mathbf{x}_{i_2},\dotsc,\mathbf{x}_{i_{d+2}})}{\sigma}}\\
&\geq \sum_{j\in \mathrm{I}_k}\sum_{\latop{i_2,\dotsc,i_{d+2}\in
\mathrm{I}_k \setminus \{i,j\}}{\mathrm{and\ are\ distinct}}}
e^{-\frac{2}{\sigma} \cdot E_{\mu_k^{d+2}}\,c_{\mathrm{p}}\left(\mathfrak{X}_i,\mathfrak{X}_{i_2},\dotsc,\mathfrak{X}_{i_{d+2}}\right)}\\
&\geq \sum_{j\in \mathrm{I}_k}\sum_{\latop{i_2,\dotsc,i_{d+2}\in
\mathrm{I}_k \setminus \{i,j\}}{\mathrm{and\ are\ distinct}}}
e^{-\frac{2}{\sigma} \cdot \sqrt{E_{\mu_k^{d+2}}\,c^2_{\mathrm{p}}\left(\mathfrak{X}_i,\mathfrak{X}_{i_2},\dotsc,\mathfrak{X}_{i_{d+2}}\right)}}\\
&=e^{-\frac{2}{\sigma}\cdot c_{\mathrm{p}}(\mu_k)}\cdot
\widetilde{d}_k,
\end{align*}
where in the last step we have used
equation~\eqref{def:polar_curv_mu}. Letting
$$\varepsilon_2 := \min_{1\leq k\leq K} e^{-\frac{2}{\sigma} \cdot
c_{\mathrm{p}}(\mu_k)} = e^{-\frac{2}{\sigma}\cdot \max_{1\leq k\leq
K} c_{\mathrm{p}}(\mu_k)},$$ we have that
\[E_{\mu_\mathrm{p}}(D_{ii})
%\geq E_{\mu_\mathrm{p}}\left(\sum_{j\in \mathrm{I}_k} W_{ij}\right)
\geq \varepsilon_2\cdot \widetilde{d}_k, \quad i\in \mathrm{I}_k,
1\leq k\leq K.\]
Equivalently,
\[E_{\mu_\textrm{p}}(\mathbf{D})\geq \varepsilon_2\cdot \widetilde{\mathbf{D}}.\]

\subsection{Proof of Theorem~\ref{thm:good_with_high_prob}}
\label{prf:thm_good_with_high_prob}
We first bound the expectation of the perturbation
$\normF{\mathcal{E}_\textrm{p}}^2$, where
$\mathcal{E}_\textrm{p}=\mathcal{A}_\textrm{p}-\widetilde{\mathcal{A}}$,
and then apply McDiarmid's inequality~\cite{McDiarmid89} to obtain a
probabilistic estimate for $\normF{\mathcal{E}_\textrm{p}}^2$.
Finally, we conclude the proof by combining the probabilistic
estimate together with Theorem~\ref{thm:perturbation_main}.

Using the definitions of the sets $\mathrm{I}_1,\ldots,\mathrm{I}_K$
and the tensors $\mathcal{A}_\textrm{p}$ and
$\widetilde{\mathcal{A}}$, we express
$\normF{\mathcal{E}_\textrm{p}}^2$ as a function of the random
variables $\mathfrak{X}_1, \ldots, \mathfrak{X}_N$:
\begin{align*}
\normF{\mathcal{E}_\textrm{p}}^2 &=
\sum_{k=1}^K\sum_{\mathrm{I}_k^{d+2}}
\left(1-e^{\frac{-c_\mathrm{p}(\mathfrak{X}_{i_1},\dotsc,\mathfrak{X}_{i_{d+2}})}{\sigma}}\right)^2
+ \sum_{\left(\bigcup_{k=1}^K\mathrm{I}_{k}^{d+2}\right)^c}
\left(e^{\frac{-c_\mathrm{p}(\mathfrak{X}_{i_1},\dotsc,\mathfrak{X}_{i_{d+2}})}{\sigma}}\right)^2.
\end{align*}
By applying the inequality: $1-e^{-|x|}\leq |x|$, we obtain that
\begin{align*}
\normF{\mathcal{E}_\textrm{p}}^2 &\leq
\sum_{k=1}^K\sum_{\mathrm{I}_k^{d+2}}
\frac{c^2_\mathrm{p}(\mathfrak{X}_{i_1},\ldots,\mathfrak{X}_{i_{d+2}})}{\sigma^2}
+ \sum_{\left(\bigcup_{k=1}^K\mathrm{I}_{k}^{d+2}\right)^c}
e^{\frac{-c_\mathrm{p}(\mathfrak{X}_{i_1},\ldots,\mathfrak{X}_{i_{d+2}})}{\sigma/2}}.
%\leq \\
%& \sum_{k=1}^K \sum_{(\mathfrak{X}_{i_1},\dotsc
%,\mathfrak{X}_{i_{d+2}})\in {S_k}} \frac{M_k^4}{2 \, \sigma^4} +
%\sum_{(\mathfrak{X}_{i_1},\dotsc,\mathfrak{X}_{i_{d+2}})\in {S^c}}
%e^{\frac{-c^2_\mathrm{p}(\mathfrak{X}_{i_1},\ldots
%,\mathfrak{X}_{i_{d+2}})}{\sigma^2}}\,.
\end{align*}
%
%Applying Proposition~\ref{prop:f_g},
%We observe that every single term of the right-hand side of the last
%equation is a function of exactly $d+2$ random variables.
%Moreover, every term of the first sum contains $d+2$ i.i.d.~random
%variables, all supported on $\supp(\mu_k)$, for some $1 \leq k
%\leq K$.
We then take the expectation of $\normF{\mathcal{E}_\textrm{p}}^2$
(with respect to $\mu_\mathrm{p}$) using
equations~\eqref{def:polar_curv_mu} and~\eqref{def_incidence} and
have that
\begin{align}
\label{ineq:Expectaion_norm_E_squared}
E_{\mu_\mathrm{p}}(\normF{\mathcal{E}_\textrm{p}}^2) &\leq
\frac{1}{\sigma^2}\sum_{k=1}^K N_k^{d+2} c_\mathrm{p}^2(\mu_k) %\int_{{\mathrm{S}_k}}
%c^2_\mathrm{p}(\mathbf{y}_{1},\dotsc,\mathbf{y}_{d+2})\, \di\mu_k (\mathbf{y}_{1})\ldots\di\mu_k (\mathbf{y}_{d+2}) \\
+ N^{d+2} C_\mathrm{in}(\mu_1,\ldots,\mu_K; \sigma/2)
%\int_{{S^c}}e^{\frac{-c_\mathrm{p}(\mathbf{y}_{1},\dotsc,\mathbf{y}_{d+2})}{\sigma/2}}
%\di\mu_\mathrm{s}(\mathbf{y}_{1})\ldots\di\mu_\mathrm{s}(\mathbf{y}_{d+2}).
\nonumber \\
&= N^{d+2} \cdot \left(\frac{1}{\sigma^2} \sum_{k=1}^K
\left(\frac{N_k}{N}\right)^{d+2} c_\mathrm{p}^2(\mu_k) +
C_\mathrm{in}(\mu_1,\ldots,\mu_K; \sigma/2)\right) \nonumber \\
&\leq \alpha \cdot N^{d+2},
\end{align}
in which
\[
\alpha := \frac{1}{\sigma^2}\cdot \sum_{k=1}^K c_\mathrm{p}^2(\mu_k)
+ C_\mathrm{in}(\mu_1,\ldots,\mu_K; \sigma/2).
\]
%The constant $K^{d+2}$ in front of the second integral comes from
%the correct substitution of $\di\mu_\mathrm{s}$, defined in
%equation~\eqref{eq:def_mu_\mathrm{s}}, into all relevant $\di\mu_k$,
%$k=1,\ldots, K$.
%Applying McDiarmid's inequality to
%Noting that both $|S|$ and $|S^c|$ are no more than
%$N^{d+2}$, we simplify the last equation as follows.
%Consequently,
%\begin{align}
%\label{ineq:Expectaion_norm_E_squared}
%E_{\mu_p}(\norm{\mathbf{E}}^2_F) &\leq
%\frac{N^{d+2}}{\sigma^2}\sum_{k=1}^{K} c_\mathrm{p}^2(\mu_k) +
%N^{d+2} C_{\mathrm{incd}}(\mu_\mathrm{s},\sigma/2) = \alpha N^{d+2}.
%\end{align}
%Therefore, by applying Theorem~\ref{thm:perturbation_main} we
%conclude that
%\begin{align*}
%&E_{\mu_\mathrm{p}}(\tv
%\mid\mathrm{Assumption~\ref{assmp:large_D_ii} \ holds})\\ &\leq
%C(K,d,\varepsilon_1,\varepsilon_2) N^{-(d+2)}
%E_{\mu_\mathrm{p}}(\norm{\mathcal{E}}_F^2)\\ &\leq \alpha \cdot
%C(K,d,\varepsilon_1,\varepsilon_2).\end{align*}
%
%We have thus shown that if $\sigma$ can be chosen so that $\alpha$
%is sufficiently small, then the TSCC algorithm will perform well
%{\it on average}. We next show that TSCC can perform well {\it with
%high probability}.

We next note that
%Define
%\[f(\mathbf{x}_1,\ldots,\mathbf{x}_N)=\norm{\mathbf{E}}^2_F,\]
%then
for each fixed $1 \leq i\leq  N$,
\[\sup_{\mathfrak{X}_1,\ldots,\mathfrak{X}_N,\widehat{\mathfrak{X}}_i}
|\normF{\mathcal{E}_\mathrm{p}}^2(\mathfrak{X}_1,\ldots,\mathfrak{X}_i,
\ldots,\mathfrak{X}_N)-
\normF{\mathcal{E}_\mathrm{p}}^2(\mathfrak{X}_1,\ldots,\widehat{\mathfrak{X}}_i,
\ldots, \mathfrak{X}_N)| \leq (d+2)\cdot N^{d+1}.\]
Indeed, the number of additive terms in
$\normF{\mathcal{E}_\mathrm{p}}^2(\mathfrak{X}_1,\ldots,\mathfrak{X}_N)$
that contain $\mathfrak{X}_i$ is $(d+2)\cdot\perm{N-1}{d+1}$, and
each of them is between 0 and 1.

The above property implies that $\normF{\mathcal{E}_\mathrm{p}}^2$
satisfies McDiarmid's inequality~\cite{McDiarmid89}, that is,
\begin{align*}
\mu_\mathrm{p}\left(\normF{\mathcal{E}_\mathrm{p}}^2-E_{\mu_\mathrm{p}}(\normF{\mathcal{E}_\mathrm{p}}^2)\geq
\alpha N^{d+2}\right)
%& \equiv
%\mu_p(\{(\mathbf{x}_1,\ldots,\mathbf{x}_N) \mid
%\norm{\mathbf{E}}^2_F-E_{\mu_p}(\norm{\mathbf{E}}^2_F)\geq \alpha
%N^{d+2}\})\\
& \leq e^{-2N\alpha^2/(d+2)^2}.
\end{align*}
Combining the last equation with
equation~\eqref{ineq:Expectaion_norm_E_squared} yields that
\begin{align*}
\mu_\mathrm{p}\left(\normF{\mathcal{E}}^2\geq 2\alpha N^{d+2}\right) & \leq
e^{-2N\alpha^2/(d+2)^2},
\end{align*}
or equivalently,
\begin{align*}
\mu_\mathrm{p}\left(N^{-(d+2)}\normF{\mathcal{E}_\mathrm{p}}^2 <
2\alpha\right) & \geq 1- e^{-2N\alpha^2/(d+2)^2}.
\end{align*}
%Consequently,
%\begin{align*} %\label{eq:est_e_curv2}
%E_\mu(\norm{\mathbf{E}}_F)&\leq\sqrt{E_\mu(\norm{\mathbf{E}}2_F)}\nonumber\\
%&\leq N^{\frac{d+2}{2}}\left(\frac{1}{\sigma2}\left(
%C2(\mu_1)+\dotsb+C2(\mu_K)\right) +
%C_{\mathrm{incd}}(\mu,\frac{\sigma}{2})\right)^{1/2}.
%\end{align*}
%
Consequently, combining Theorem~\ref{thm:perturbation_main} and the
last equation gives that, if \[2\alpha \leq \frac{1}{8C_1},\textrm{
where } C_1=C_1(K,d,\varepsilon_1,\varepsilon_2) \textrm{ is defined
in equation~\eqref{eq:constant_C1}},\] then
\begin{align*}
&\mu_\mathrm{p}\left(\tv < 2\alpha \cdot C_1 \mid
\mathrm{Assumption~\ref{assmp:large_D_ii}\ holds} \right) \\
&\geq \mu_\mathrm{p}\left(\tv < 2\alpha \cdot C_1 \mid
\mathrm{Assumption~\ref{assmp:large_D_ii}\ holds,\ and}\
N^{-(d+2)}\normF{\mathcal{E}_\mathrm{p}}^2 < 2\alpha \right)
\\ &\qquad \cdot \mu_\mathrm{p}\left(N^{-(d+2)}\normF{\mathcal{E}_\mathrm{p}}^2
< 2\alpha \mid \mathrm{Assumption~\ref{assmp:large_D_ii}\ holds}\right)\\
& = 1 \cdot \mu_\mathrm{p}\left(N^{-(d+2)}\normF{\mathcal{E}_\mathrm{p}}^2< 2\alpha\right) \\
&\geq 1-e^{-2N\alpha^2/(d+2)^2}.\end{align*}
%\end{proof}

\subsection{Proof of Equation~\eqref{incd:lines_tscc} in Example~\ref{ex:lines_tscc}}
\label{prf:ex_lines_tscc} For any three points
$\mathbf{p}_1(x_1,0),\mathbf{p}_2(x_2,0)\in \text{L}1$, and
$\mathbf{q}(0,y)\in \text{L}2$, their polar curvature is
bounded below by
\begin{align*}
c_\mathrm{p}(\mathbf{p}_1,\mathbf{p}_2,\mathbf{q}) &=
\diam\{\mathbf{p}_1,\mathbf{p}_2,\mathbf{q}\}\cdot
\sqrt{\sin^2\angle\mathbf{p}_1\mathbf{p}_2\mathbf{q}+\sin^2\angle\mathbf{p}_2\mathbf{p}_1\mathbf{q}+\sin^2\angle\mathbf{p}_1\mathbf{q}\mathbf{p}_2}\\
&\geq \max\left(\sqrt{x_1^2+y^2},\sqrt{x_2^2+y^2}\right)
\cdot\sqrt{\frac{y^2}{x_1^2+y^2}+\frac{y^2}{x_2^2+y^2}} \\
&\geq \sqrt{y^2+y^2} = \sqrt{2}\cdot y.
\end{align*}
Thus, by using the symmetry of the lines, we obtain that
\begin{align*}
C_\mathrm{in}(\mu_1,\mu_2;\sigma) &=
\int_{\text{L1}}\int_{\text{L1}}\int_\text{L2} e^{-\frac{c_\mathrm{p}(\mathbf{p}_1,\mathbf{p}_2,\mathbf{q})}{\sigma}}\,\di\mu_1(\mathbf{p}_1)\di\mu_1(\mathbf{p}_2)\di\mu_2(\mathbf{q}) \\
&\leq \int_0^L e^{-\frac{\sqrt{2}\,y}{\sigma}}\, \frac{\di y}{L} =
\frac{\sigma}{\sqrt{2}L}\left(1-e^{-\sqrt{2}L/\sigma}\right).
\end{align*}

\subsection{Proof of Equation~\eqref{incd:lines} in Example~\ref{ex:lines}}
\label{prf:ex_lines} For any two points $\mathbf{p}(x,0)\in
\text{L}1,\mathbf{q}(r\cos\theta,r\sin\theta)\in \text{L}2$, the
polar curvature of $\mathbf{p,q}$ and the origin $\mathbf{o}$ is
bounded below by
\begin{align*}
c_\mathrm{p}(\mathbf{o,p,q}) &= \diam\{\mathbf{o,p,q}\}\cdot
\sqrt{\sin^2\theta+\sin^2\angle\mathbf{opq}+\sin^2\angle\mathbf{oqp}}\\&\geq
\max(x,r)\cdot\sin\theta.
\end{align*}
Thus, the incidence constant is bounded above by
\begin{align*}
C_\mathrm{in,L}(\mu_1,\mu_2;\sigma) &=
\int_{\text{L1}}\int_\text{L2} e^{-\frac{c_\mathrm{p}(\mathbf{o,p,q})}{\sigma}}\,\di\mu_1(\mathbf{p})\di\mu_2(\mathbf{q}) \\
&\leq \int_0^L\int_0^L
e^{-\frac{\max(x,r)\cdot \sin\theta}{\sigma}}\, \frac{\di x}{L}\frac{\di r}{L}  \\
&=2 \iint_{0\leq x\leq r\leq L} e^{-\frac{r\sin\theta}{\sigma}} \, \frac{\di x}{L}\frac{\di r}{L}  \\
&=\frac{2}{L}\int_0^L r\cdot e^{-\frac{r\sin\theta}{\sigma}} \,\frac{\di r}{L}  \\
 &=2\left(\frac{\sigma}{L\sin\theta}\right)^2 \cdot
\left(1-e^{-\frac{L\sin\theta}{\sigma}}\left(1+\frac{L\sin\theta}{\sigma}\right)\right).
\end{align*}

\subsection{Proof of Equation~\eqref{incd:rectangles} in Example~\ref{ex:rectangles}}
\label{prf:ex_rectangles} For any $\mathbf{p}(x,y_2)\in \text{R1},
\mathbf{q}(x_1,y)\in \text{R2}$, we define
$\widetilde{\mathbf{p}}(x,\epsilon)\in \text{R}1,
\widetilde{\mathbf{q}}(\epsilon,y)\in \text{R}2$. The polar
curvature of $\mathbf{p,q}$ and the origin $\mathbf{o}$ is bounded
below by
\begin{align*}
c_\mathrm{p}(\mathbf{o,p,q}) &\geq \max(\norm{\mathbf{op}},
\norm{\mathbf{oq}}) \cdot \sin\angle\mathbf{poq} \geq \max(x,y)
\cdot
\sin\angle\widetilde{\mathbf{p}}\mathbf{o}\widetilde{\mathbf{q}}
\\ &=
\frac{\max(x,y)\cdot(xy-\epsilon^2)}{\sqrt{(x^2+\epsilon^2)(y^2+\epsilon^2)}}.
\end{align*}
Thus, the incidence constant is
\begin{align*}
C_\mathrm{in,L}(\mu_1,\mu_2;\sigma) &=
\int_{\text{R1}}\int_\text{R2}
e^{-\frac{c_\mathrm{p}(\mathbf{o,p,q})}{\sigma}}\,
\frac{\di x}{L}\frac{\di y_2}{\epsilon}\frac{\di x_1}{\epsilon}\frac{\di y}{L}\\
&\leq
\frac{1}{L^2}\cdot\int_{\epsilon}^{L+\epsilon}\int_{\epsilon}^{L+\epsilon}
e^{-\frac{\max(x,y)\cdot(xy-\epsilon^2)}{\sigma\cdot\sqrt{(x^2+\epsilon^2)(y^2+\epsilon^2)}}}\,
\di x\di y.
\end{align*}
Changing variables $x:=x/\epsilon,y:=y/\epsilon$ and setting $\omega
:= L/\epsilon$ gives that
\begin{align*}
C_\mathrm{in}(\mu_1,\mu_2;\sigma) &\leq
\frac{1}{\omega^2}\cdot\int_{1}^{1+\omega}\int_{1}^{1+\omega}
e^{-\frac{\max(x,y)
\cdot(xy-1)}{\sigma\cdot\sqrt{(x^2+1)(y^2+1)}}}\, \di x \di y.
\end{align*}

We observe that the integrand is bounded between 0 and 1,
symmetric about $x$ and $y$, and decreasing in each of its
arguments. We thus obtain that
\begin{align*}
C_\mathrm{in,L}(\mu_1,\mu_2;\sigma) &\leq
\frac{1}{\omega^2}\cdot\left(\int_{1}^{1+\sqrt[4]{\sigma}}\int_{1}^{1+\sqrt[4]{\sigma}}
+ 2\int_{1}^{1+\sqrt[4]{\sigma}}\int_{1+\sqrt[4]{\sigma}}^{1+\omega}
+
\int_{1+\sqrt[4]{\sigma}}^{1+\omega}\int_{1+\sqrt[4]{\sigma}}^{1+\omega}
\right) \\
& \qquad \qquad
e^{-\frac{\max(x,y)\cdot(xy-1)}{\sigma\cdot\sqrt{(x^2+1)(y^2+1)}}}
\,\di x\di y\\
&\leq \frac{1}{\omega^2}\cdot \left(\left(\sqrt[4]{\sigma}\right)^2
+ 2\cdot \sqrt[4]{\sigma}\cdot (\omega-\sqrt[4]{\sigma})\cdot
e^{-\frac{\left(1+\sqrt[4]{\sigma}\right)\cdot\left(1\cdot\left(1+\sqrt[4]{\sigma}\right)-1\right)}{\sigma\cdot\sqrt{2\cdot\left(1+\left(1+\sqrt[4]{\sigma}\right)^2\right)}}}
\right)\\ &\qquad +
\frac{1}{\omega^2}\cdot\left(\omega-\sqrt[4]{\sigma}\right)^2 \cdot
e^{-\frac{\left(1+\sqrt[4]{\sigma}\right)\cdot\left(\left(1+\sqrt[4]{\sigma}\right)^2-1\right)}{\sigma\cdot\left(1+\left(1+\sqrt[4]{\sigma}\right)^2\right)}}\\
&\leq \frac{\sqrt{\sigma}}{\omega^2} +
\frac{2\sqrt[4]{\sigma}}{\omega}\cdot
e^{-1/\left(2\sigma^{3/4}\right)} + e^{-1/\sigma^{3/4}}.
\end{align*}

\subsection{Proof of Equation~\eqref{incd:disks} in Example~\ref{ex:disks}}
\label{prf:ex_disks} Let
$\mathbf{p}(0,\rho\cos\varphi,\rho\sin\varphi)\in \text{D1}$, and
$\mathbf{q}_1(0,r_1\cos\theta_1,r_1\sin\theta_1),\mathbf{q}_2(0,r_2\cos\theta_2,r_2\sin\theta_2)\in
\text{D2}$. Then the polar curvature of these three points and the
origin $\mathbf{o}$ has the following lower bound:
\begin{align*}
c_\mathrm{p}(\mathbf{o,p},\mathbf{q}_1,\mathbf{q}_2) \geq
|\mathbf{op}| \cdot
\mathrm{psin}_\mathbf{o}(\mathbf{p},\mathbf{q}_1,\mathbf{q}_2)=\rho
\cdot \sin\varphi\, \sin\abs{\theta_1-\theta_2}.
\end{align*}

Due to the symmetry of the two disks, we have that
\begin{align*}
C_\mathrm{in,L}(\mu_1,\mu_2; \sigma) &=
\int_\mathrm{D1}\int_\mathrm{D2}\int_\mathrm{D2}
e^{-c_\mathrm{p}(\mathbf{o},\mathbf{p},\mathbf{q}_1,\mathbf{q}_2)/\sigma}\,\di\mu_1(\mathbf{p})\di\mu_2(\mathbf{q}_1)\di\mu_2(\mathbf{q}_2)\\
%&\leq \int_\mathrm{D1}\int_\mathrm{D2}\int_\mathrm{D2}
%e^{-\frac{\rho\sin\varphi\cdot\sin\abs{\theta_1-\theta_2}}{\sigma}}
%\,\di\mu_1(\mathbf{p})\di\mu_2(\mathbf{q}_1)\di\mu_2(\mathbf{q}_2)\\
&\leq \int_0^1\int_0^\pi\int_{-\pi/2}^{\pi/2}\int_{-\pi/2}^{\pi/2}
e^{-\frac{\rho\sin\varphi\cdot\sin\abs{\theta_1-\theta_2}}{\sigma}}
\,\frac{\rho\di\rho \di\varphi}{\pi/2}\frac{\di\theta_1}{\pi}\frac{\di\theta_2}{\pi}\\
&=\frac{4}{\pi^3} \cdot\int_0^1\int_0^\pi\iint_{-\frac{\pi}{2}\leq
\theta_2\leq \theta_1\leq \frac{\pi}{2}}
e^{\frac{-\rho\sin\varphi\cdot\sin(\theta_1-\theta_2)}{\sigma}}
\,\rho\di\rho\di\varphi\di\theta_1 \di\theta_2.
\end{align*}

Changing variables $\theta:=\theta_1-\theta_2,\theta_2:=\theta_2$
and exchanging the corresponding double integral, we obtain that
\begin{align*}
C_\mathrm{in,L}(\mu_1,\mu_2; \sigma) &\leq \frac{4}{\pi^3} \cdot
\int_0^1 \int_0^\pi \int_0^\pi e^{-\frac{\rho\sin\varphi\cdot
\sin\theta}{\sigma}} \, \rho\di\rho \di\varphi \,
(\pi-\theta)\di\theta
\\
&\leq \frac{4}{\pi^2} \cdot \int_0^1 \int_0^\pi \int_0^\pi
e^{-\frac{\rho\sin\varphi\cdot \sin\theta}{\sigma}} \, \rho \di\rho \di\varphi \di\theta\\
&= \frac{16}{\pi^2} \cdot\int_0^1 \int_0^{\pi/2} \int_0^{\pi/2}
e^{-\frac{\rho\sin\varphi\cdot \sin\theta}{\sigma}}\, \rho\di\rho
\di\varphi \di\theta.
\end{align*}

We observe that the integrand is bounded between 0 and 1,
symmetric about $\varphi$ and $\theta$, and decreasing in each of
them. Thus,
\begin{align*}
C_\mathrm{in,L}(\mu_1,\mu_2; \sigma) &\leq
\frac{16}{\pi^2}\cdot\int_0^1 \left(\int_0^{\sqrt[4]{\sigma}}
\int_0^{\sqrt[4]{\sigma}} +
2\int_0^{\sqrt[4]{\sigma}}\int_{\sqrt[4]{\sigma}}^{\frac{\pi}{2}} +
\int_{\sqrt[4]{\sigma}}^{\frac{\pi}{2}}\int_{\sqrt[4]{\sigma}}^{\frac{\pi}{2}}
\right) \\ &\qquad \qquad e^{-\frac{\rho\sin\varphi\cdot \sin\theta}{\sigma}} \,\rho \di\rho \di\varphi \di\theta \\
&\leq \frac{16}{\pi^2}\cdot \left(
\left(\sqrt[4]{\sigma}\right)^2 + 2 \cdot \sqrt[4]{\sigma} \cdot
\left(\frac{\pi}{2}-\sqrt[4]{\sigma}\right)\right) \cdot \int_0^1 \rho\di\rho
\\& \qquad + \frac{16}{\pi^2}\cdot \left(\frac{\pi}{2}-\sqrt[4]{\sigma}\right)^2 \cdot \int_0^1
e^{-\frac{\rho\cdot\left(\sin\sqrt[4]{\sigma}\right)^2}{\sigma}} \,\rho \di\rho\\
%&\leq \frac{8}{\pi^2}\sqrt{\sigma} + \frac{8}{\pi}\sqrt[4]{\sigma} +
%4\int_0^1
%e^{-\frac{\rho\cdot\left(\sin\sqrt[4]{\sigma}\right)^2}{\sigma}} \,\rho\di\rho\\
&\leq \frac{8\sqrt{\sigma}}{\pi^2} + \frac{8\sqrt[4]{\sigma}}{\pi} +
\frac{4\sigma^2}{(\sin\sqrt[4]\sigma)^4}.
\end{align*}

%\bibliographystyle{plain}
%\bibliography{myrefs}

\end{document}